\documentclass{article}

\usepackage{amsmath,amsfonts,bm}

\def\eqref#1{equation~\ref{#1}}

\def\1{\bm{1}}

\DeclareMathAlphabet{\mathsfit}{\encodingdefault}{\sfdefault}{m}{sl}
\SetMathAlphabet{\mathsfit}{bold}{\encodingdefault}{\sfdefault}{bx}{n}

\newcommand{\R}{\mathbb{R}}

\DeclareMathOperator*{\argmax}{arg\,max}
\DeclareMathOperator*{\argmin}{arg\,min}

\usepackage{microtype}
\usepackage{graphicx}
\usepackage{booktabs} 

\usepackage{hyperref}

\usepackage[accepted]{icml2024}

\usepackage{amsmath}
\usepackage{amssymb}
\usepackage{mathtools}
\usepackage{amsthm}
\usepackage{bbm}
\usepackage{stackengine}

\usepackage[capitalize,noabbrev]{cleveref}

\graphicspath{ {./images/} }
\usepackage{caption}
\usepackage{subcaption}
\usepackage{wrapfig}
\usepackage{enumitem}
\usepackage{float}
\usepackage{multirow}
\usepackage{stfloats}
\usepackage{array}
\usepackage[export]{adjustbox}

\newtheorem{theorem}{Theorem}[section]
\newtheorem{corollary}[theorem]{Corollary}

\newcommand{\bx}{\mathbf{x}}
\newcommand{\by}{\mathbf{y}}

\newcommand{\mcS}{\mathcal{S}}
\newcommand{\mcA}{\mathcal{A}}
\newcommand{\mcG}{\mathcal{G}}
\newcommand{\mcP}{\mathcal{P}}
\newcommand{\mcD}{\mathcal{D}}

\usepackage[textsize=tiny]{todonotes}

\icmltitlerunning{Learning to Reach Goals via Diffusion}

\begin{document}

\twocolumn[
\icmltitle{Learning to Reach Goals via Diffusion}



\icmlsetsymbol{equal}{*}

\begin{icmlauthorlist}
\icmlauthor{Vineet Jain}{yyy,comp}
\icmlauthor{Siamak Ravanbakhsh}{yyy,comp}
\end{icmlauthorlist}

\icmlaffiliation{yyy}{Department of Computer Science, McGill University, Montr\'{e}al, Canada}
\icmlaffiliation{comp}{Mila - Quebec Artificial Institute Institute, Montr\'{e}al, Canada}


\icmlcorrespondingauthor{Vineet Jain}{jain.vineet@mila.quebec}

\icmlkeywords{Machine Learning, ICML}

\vskip 0.3in
]



\printAffiliationsAndNotice{}  

\begin{abstract}
We present a novel perspective on goal-conditioned reinforcement learning by framing it within the context of denoising diffusion models.  Analogous to the diffusion process, where Gaussian noise is used to create random trajectories that walk away from the data manifold, we construct trajectories that move away from potential goal states. We then learn a goal-conditioned policy to reverse these deviations, analogous to the score function. 
This approach, which we call Merlin\footnote{In Arthurian legends, Merlin the Wizard is said to have lived backward in time, which gave him the ability to predict the future.}, can reach specified goals from arbitrary initial states without learning a separate value function.
In contrast to recent works utilizing diffusion models in offline RL, Merlin stands out as the first method to perform diffusion in the state space, requiring only one ``denoising" iteration per environment step.
We experimentally validate our approach in various offline goal-reaching tasks, demonstrating substantial performance enhancements compared to state-of-the-art methods while improving computational efficiency over other diffusion-based RL methods by an order of magnitude. Our results suggest that this perspective on diffusion for RL is a simple and scalable approach for sequential decision making\footnote{Code for Merlin is available at \href{https://github.com/vineetjain96/merlin}{\texttt{https://github.com/}} \href{https://github.com/vineetjain96/merlin}{\texttt{vineetjain96/merlin}}.}.

\end{abstract}

\section{Introduction}
Reinforcement learning (RL) is a powerful paradigm for agents to learn behaviors supervised by only a reward signal. However, the agent usually excels in accomplishing a specific task, and this process requires constructing a reward function. Ideally, a generalist agent should acquire a repertoire of skills that can be applied across a range of tasks. Goal-conditioned RL (GCRL) \citep{kaelbling1993learning, schaul2015universal, chane2021goal} is a paradigm to learn general policies that can reach arbitrary target states within an environment without requiring extensive retraining. The reward function is sparse and binary---that is a  reward of one for reaching the desired goal and zero reward otherwise--- eliminating the need for hand-engineered reward functions. The sparse reward function necessitates intensive exploration, which can be infeasible and often unsafe in practical settings. Concurrently, offline RL has gained attention in recent years for learning policies from large amounts of pre-collected datasets  without any environmental interaction \citep{levine2020offline, prudencio2023survey}. In recent years, the size of offline RL datasets has steadily increased, allowing agent skills to scale with data  \citep{guss2019minerl, mathieu2021starcraft}. Combining offline and goal-conditioned RL can potentially enhance both generalization and scalability in practical scenarios.

However, offline GCRL introduces new challenges. Many existing methods rely on learning a value function \citep{yang2021rethinking, ma2022offline}, which estimates the expected discounted return for a given state-action pair. During training, policies often generate actions that are not present in the offline dataset, leading to inaccuracies in the value function estimation. These inaccuracies, compounded over time, can cause policies to diverge \citep{levine2020offline}. The value estimation problem is further exacerbated by a sparse binary reward signal common in goal-conditioned settings.  Previous attempts to address this problem involve constraints on policies or conservative value function updates \citep{kumar2019stabilizing, kumar2020conservative}, which compromise policy performance \citep{levine2020offline} and make generalization challenging.

\begin{figure}[t]
  \centering
  \includegraphics[trim={15.5cm 1cm 13cm 2cm},clip,width=0.8\linewidth]{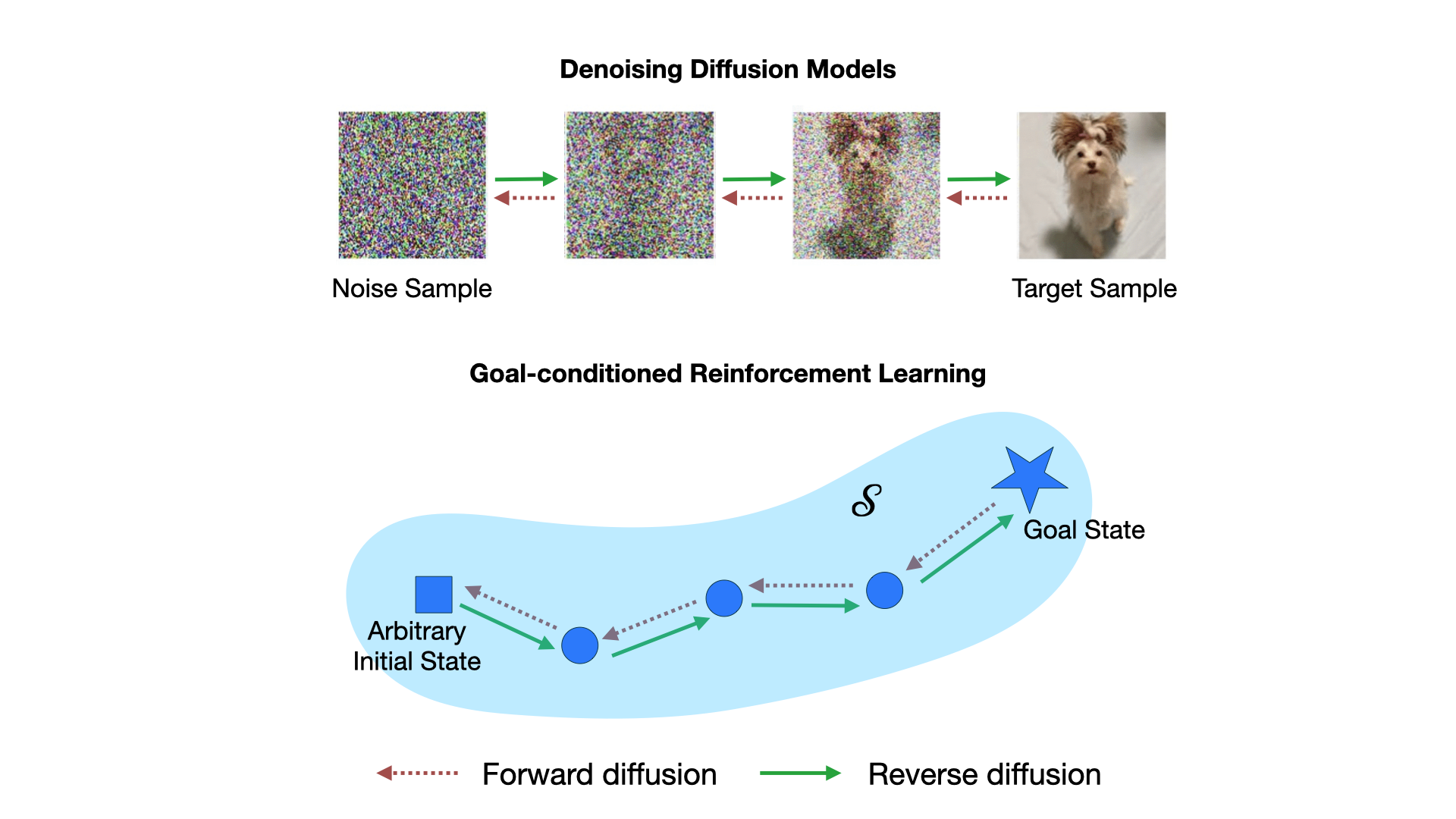}
  \caption{\small{Reverse diffusion policy.}}
  \vspace{-1.4em}
  \label{fig:diff_diagram}
\end{figure}

To tackle the GCRL problem of reaching specified goal states from arbitrary initial states, this paper draws inspiration from diffusion models - a powerful class of generative models that can map random noise in high-dimensional spaces to target data samples through iterative denoising \citep{sohl2015deep, ho2020denoising}. Building upon the idea of introducing controlled noise to destroy the structure of the target data distribution, we employ a similar strategy for GCRL by constructing trajectories that move away from desired goals during the learning process. A goal-conditioned policy is then trained to reverse (or effectively ``denoise") these trajectories; see \Cref{fig:diff_diagram}. By navigating away from desired goals and subsequently correcting these deviations, the policy learns to reach any predefined goal state from arbitrary initial states. This backward view of diffusion gives us control over the goal distribution, so that all the trajectories that are used to train the policy end up in some desired goal state. In contrast, the traditional forward view of RL uses some exploration policy, which may not end up finding meaningful goals during rollouts. Notably, our approach, which we call Merlin, does not learn a value function, circumventing the value estimation issues discussed above. 

Recent works have leveraged diffusion models for offline RL by denoising Gaussian noise iteratively to generate trajectory segments \citep{janner2022planning, ajay2022conditional} or sample actions \citep{wang2022diffusion, reuss2023goal}. These methods can learn expressive policies but are computationally expensive due to the iterative denoising process required for each environment step. To our knowledge, Merlin is the first method that performs diffusion in the state space, requiring only one ``denoising" iteration per environment step. This distinction makes Merlin conceptually simpler and 10-15$\times$ more efficient 
than prior diffusion-based methods. Constructing the forward diffusion process to align with the underlying Markov Decision Process (MDP) is not immediately evident. This paper explores and experiments with three potential methods for establishing such a process. 

Our contributions can be summarized as follows: First, we develop a novel goal-conditioned reinforcement learning approach inspired by diffusion models without learning a value function. Second, we theoretically justify viewing the reverse trajectories in RL as a forward diffusion process and prove that learning a score-like policy maximizes the likelihood of reaching the specified goal states. Third, we propose three possible choices for constructing the forward diffusion process - I) reverse play of the offline trajectories in the dataset; II) learning a reverse dynamics model, and; III) a novel latent-space trajectory stitching technique grounded in nearest-neighbor search, which enables the generation of state-goal pairs \textit{across} trajectories. In \Cref{sec:exp}, we demonstrate the effectiveness of our approach in various offline goal-reaching tasks with significant performance enhancements compared to state-of-the-art methods, while improving computational efficiency by an order of magnitude compared to other diffusion-based RL methods.

\section{Preliminaries}
\subsection{Diffusion Probabilistic Models}
\label{sec:ddpm}

Diffusion probabilistic models \citep{sohl2015deep,ho2020denoising} are generative models that can be used to model a data distribution. These latent variable models are characterized by a probability distribution that evolves over time, following a forward diffusion process. The forward diffusion process is generally fixed to add Gaussian noise at each timestep according to a variance schedule $\beta_1,\ldots,\beta_T$. Let $\bx_0 \sim q(\bx_0)$ denote the data and $\bx_1,\ldots,\bx_T$ denote the corresponding latent variables. The approximate posterior $q(\bx_{1:T} \mid \bx_0)$ is given by,
\begin{align}
    q(\bx_{1:T}| \bx_0) :=  \prod_{t=1}^T q(\bx_{t}| \bx_{t-1}),\\
    q(\bx_{t} | \bx_{t-1}) := \mathcal{N}(\bx_t; \sqrt{1-\beta_t}\bx_{t-1},\beta_t \bf{I})
\end{align}
Diffusion probabilistic models then learn a denoising function that reverses the forward diffusion process. Starting at $p(\bx_T) = \mathcal{N}(\bx_T;0,\bf{I})$, the joint distribution of the reverse process is given by, 
\begin{align}
    p_\theta(\bx_{0:T}) := p(\bx_T)\prod_{t=1}^T p_\theta(\bx_{t-1} | \bx_{t}),\\ p_\theta(\bx_{t-1} | \bx_{t}) := \mathcal{N}(\bx_{t-1}; \boldsymbol{\mu}_\theta(\bx_t,t), \boldsymbol{\Sigma}_\theta(\bx_t,t))
\end{align}
where $\boldsymbol{\mu}_\theta$ and $\boldsymbol{\Sigma}_\theta$ can be neural networks.
The reverse process can produce samples matching the data distribution after a finite number of transition steps. Note that traditionally, $t=0$ corresponds to the data and higher timesteps correspond to noisy latent variables. In our discussion, we reverse this notation to be consistent with RL notation -- that is we use the maximum timestep $T$ for data (goal in GCRL), and decrease the timestep during forward diffusion.

\subsection{Goal-conditioned Reinforcement Learning}
\label{sec:gcrl}

The RL problem can be described using a Markov Decision Process (MDP), denoted by a tuple $(\mcS, \mcA, \mcP, r, \mu, \gamma)$, where $\mcS$ and $\mcA$ are the state and action spaces; $\mcP$ describes the transition probability as $\mcS \times \mcA \times \mcS \rightarrow [0, 1]$, $r: \mcS \times \mcA \rightarrow \mathbb{R}$ is the reward function, $\mu(s)$ is the initial state distribution, and $\gamma \in (0, 1]$ is the discount factor.

Goal-conditioned RL additionally considers a goal space $\mcG \coloneqq \{\phi(s)  \mid s \in \mcS\}$, where $\phi: \mcS \rightarrow \mcG$ is a known state-to-goal mapping \citep{andrychowicz2017hindsight}. For example, in the FetchPush task, a robotic arm is tasked with pushing a block to a goal position specified by $(x,y,z)$ coordinates of the block, but the state represents the positions and velocities of the various effectors and components of the robotic arm as well as the block. The reward function now depends on the goal, $r: \mcS  \times \mcA \times \mcG \rightarrow \R$. Generally, the reward function is sparse and binary, defined as $r(s, a, g) = \mathbb{I}[\|\phi(s) - g\|_2^2 \leq \delta]$, where $\delta$ is some threshold distance.

A goal-conditioned policy is denoted by $\pi : \mcS \times \mcG \rightarrow \mcA$, and given a distribution over desired goals $p(g)$, an optimal policy $\pi^*$ aims to maximize the expected return,
\begin{align*}
    J(\pi) = \mathbb{E}_{\substack{g\sim p(g),s_0\sim\mu(s_0)\\a_t\sim\pi(\cdot|s_t,g),s_{t+1}\sim \mcP(\cdot|s_t,a_t)}} \left[ \sum_{t=0}^\infty \gamma^t r(s_t, a_t, g) \right],
\end{align*}
For offline RL problems, the agent cannot interact with the environment during training and only has access to a static dataset $D = \{(s_t, a_t, g, r_t)\}$, which can be collected by some unknown policy.

\section{Reaching Goals via Diffusion}
\label{sec:diff_gcrl}
Consider the generative modeling problem of generating samples from some distribution $p_{\textrm{data}}(\bx)$ given a set of samples $\{\bx_i\}_{i=1}^N, \bx_i \in \R^d$. Diffusion modeling entails constructing a Markov chain which iteratively adds Gaussian noise to these samples. The perturbed data effectively covers the space surrounding the data manifold. The learned reverse diffusion process can then map any point drawn from a Gaussian distribution in $\R^d$ to the data manifold.

Now consider a goal-augmented MDP $(\mcS, \mcA, \mcG, \mcP, r, \mu,\gamma)$ with goals $g \in \mcG$ sampled from some unknown goal distribution $g \sim p(g)$. Goal-conditioned RL aims to learn a policy that can learn an optimal path from any state $s \in \mcS$ to the desired goal $g$. This can be viewed as similar to learning to map random noise in $\R^d$ to some target data manifold, except that the underlying space is now restricted to the state space of the MDP.

\begin{figure*}[!t]
  \centering
  \includegraphics[width=0.8\textwidth]{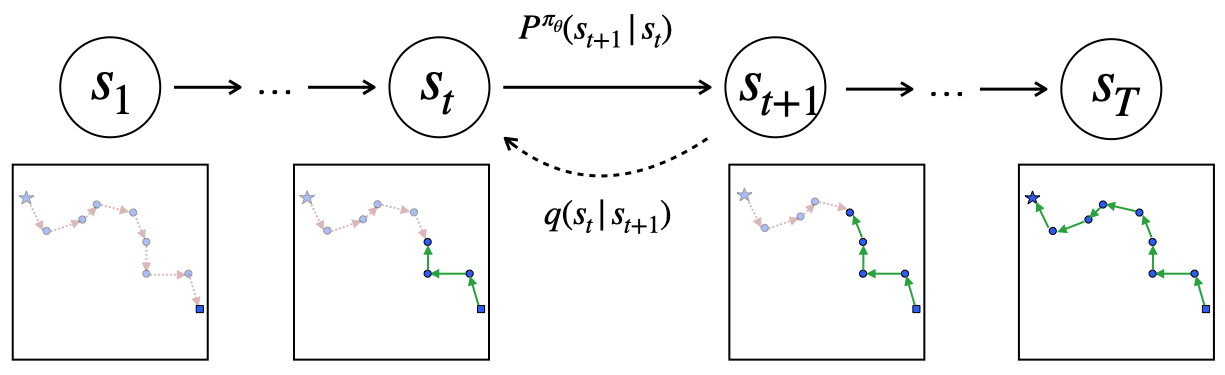}
  \caption{\small{Forward and reverse diffusion process for GCRL using 2D navigation as an example. Star represents the goal state, the red dotted arrows denote the forward process transitions $q(s_t|s_{t+1})$, and the green arrows denote the reverse process transitions $\mcP^{\pi_\theta}(s_{t+1}|s_t)$.}}
  \vspace{-1em}
  \label{fig:rl_diff}
\end{figure*}

\subsection{Forward and Reverse Process}
\label{sec:processes}
For a fixed policy, the state transitions in an MDP are given by $\mcP(s_{t+1}|s_t, a_t)$, where $a_t \sim \pi(\cdot | s_t)$. 
This is a Markov chain, and we denote the Markov transition kernel by $\mcP^\pi(s_{t+1}|s_t)$. 
Moreover, if we could assume the existence of a unique stationary distribution, running this Markov chain backwards corresponds to another Markov chain $\mcP^\pi(s_{t}|s_{t+1})$; where the relation between the two is given by the detailed balance condition \citep[e.g.,][]{chung1969reverse}. We consider this reverse Markov chain to be a forward diffusion process over the state space of the MDP. 

The question is now if we can assume a stationary distribution in different settings. Fortunately, the answer is yes! While assuming a stationary distribution for the policy is common with infinite horizons, \citet{bojun2020steady} prove the existence and uniqueness of the steady state in the episodic setting. The assumptions for their result are that (1) all policies have finite average episode length and (2) all terminal states have action-independent transitions. The latter assumption can be interpreted as resurrecting the agent according to some initial state distribution once it reaches some goal state. 

The upshot is that we can safely consider the reverse process in RL as a forward diffusion process.
The following table summarizes the analogy between diffusion models and goal-conditioned RL:
\begin{table}[h!]
\centering
  \vspace{-0.5em}
  \begin{tabular}{c | c}
  diffusion model & GCRL (Merlin)  \\ \hline
  target dataset & goal states \\
  score function & policy \\
  forward process & $q(s_t | s_{t+1})$ \\
  reverse process & $\mcP^\pi(s_{t+1}|s_t)$
  \end{tabular}
  \vspace{-1.2em}
\end{table}

Reaching goals using diffusion requires constructing a forward diffusion process and learning the corresponding reverse process. In the context of RL, the forward diffusion process involves moving backward from the goal states. It is not immediately clear how to construct this process, which we denote by $q(s_t | s_{t+1})$, and we discuss several possibilities in \Cref{sec:gen_traj}. 
Note that, as mentioned in \Cref{sec:ddpm}, we set $s_1$ as the initial state and $s_T$ as the final state for consistency with RL notation -- that is the time index is reversed in comparison to the convention for diffusion models.

\subsection{Optimization Objective}

Consider the offline setting where we have a fixed dataset of trajectories $\mathcal{D}$ generated by some unknown behavior policy $\pi_\beta$, and a trajectory $\tau \sim \mcD$, where $\tau = \{s_1, a_1,\ldots,s_T\}$. We can view this trajectory in reverse -- starting from the final state $s_T$, we apply the forward diffusion transformation $q(s_{t} | s_{t+1})$ to each state $s_{t+1}$ to obtain state $s_t$. We train a policy denoted by $\pi_\theta$ to reverse this diffusion. The corresponding reverse diffusion process is given by $\mcP(s_{t+1} | s_t, \pi_\theta(\cdot | s_t, g))$, where $g = \phi(s_T)$ is the goal. Our objective is to maximize the log-likelihood of the goal states under the reverse diffusion process. The following theorem shows that this objective can be achieved by behavior cloning, which is analogous to the process used for approximating the score in denoising diffusion models.
\begin{theorem}
\label{thm}
Consider a dataset $\mathcal{D}(g)$ collected by an unknown behavior policy $\pi_\beta$, consisting of trajectories ending in states $S_T \coloneqq \{s_T \mid g=\phi(s_T)\}$ with $q(s_T|g)$ denoting the distribution of final states corresponding to $g$. Then, behavior cloning given by $\theta^* = \argmax_\theta \mathbb{E}_{(s, a) \sim \mathcal{D}(g)} \left[ \log \pi_\theta(a | s)\right]$ is equivalent to maximizing a lower bound on the log-likelihood of the final states under the reverse diffusion process $L = \mathbb{E}_{s_T \sim q(s_T|g)} \left[\log p_\theta(s_T)\right]$.
\end{theorem}

\begin{figure*}[!b]
  \centering
  \begin{subfigure}[b]{0.265\linewidth}
      \centering
      \includegraphics[trim={4em 2.5em 0 0.5em},clip,width=\linewidth]{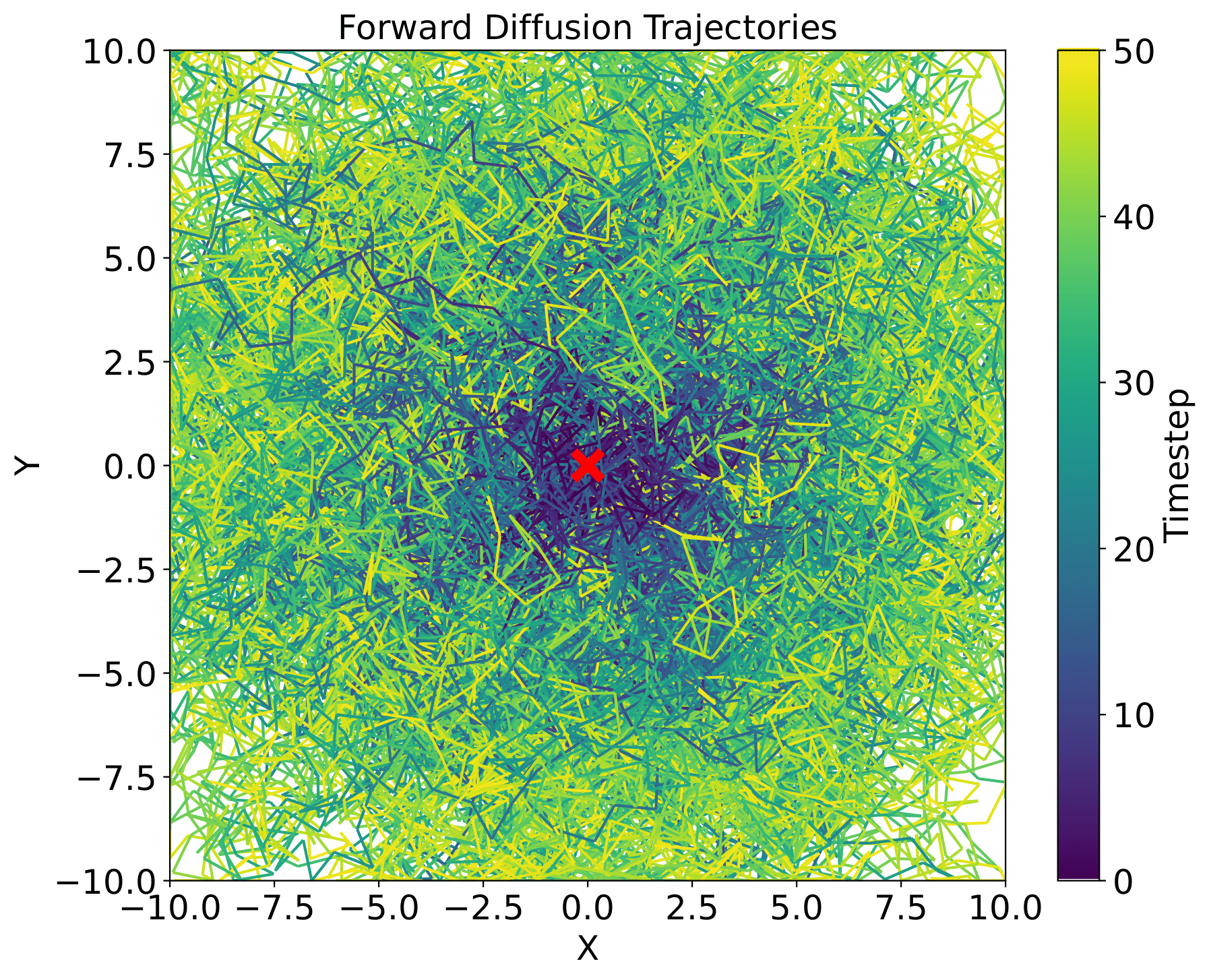}
      \caption{\small{}}
      \label{fig:diff_toy_traj}
  \end{subfigure}
  \hspace{0.5em}
  \begin{subfigure}[b]{0.22\linewidth}
      \centering
      \includegraphics[trim={4em 2.5em 0 0.5em},clip,width=\linewidth]{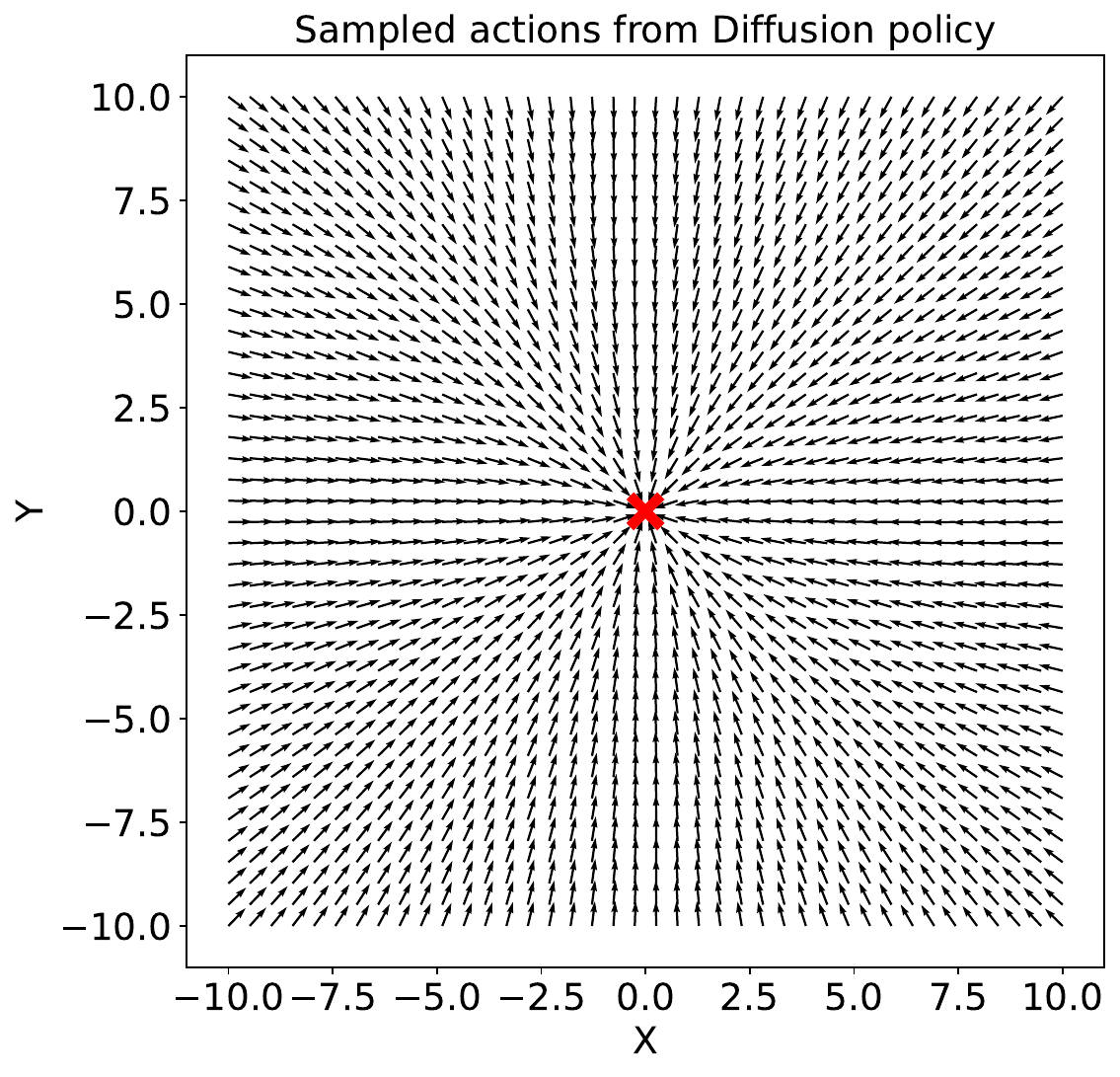}
      \caption{\small{}}
      \label{fig:diff_toy_main}
  \end{subfigure}
  \hspace{0.5em}
  \begin{subfigure}[b]{0.22\linewidth}
      \centering
      \includegraphics[trim={4em 2.5em 0 0.5em},clip,width=\linewidth]{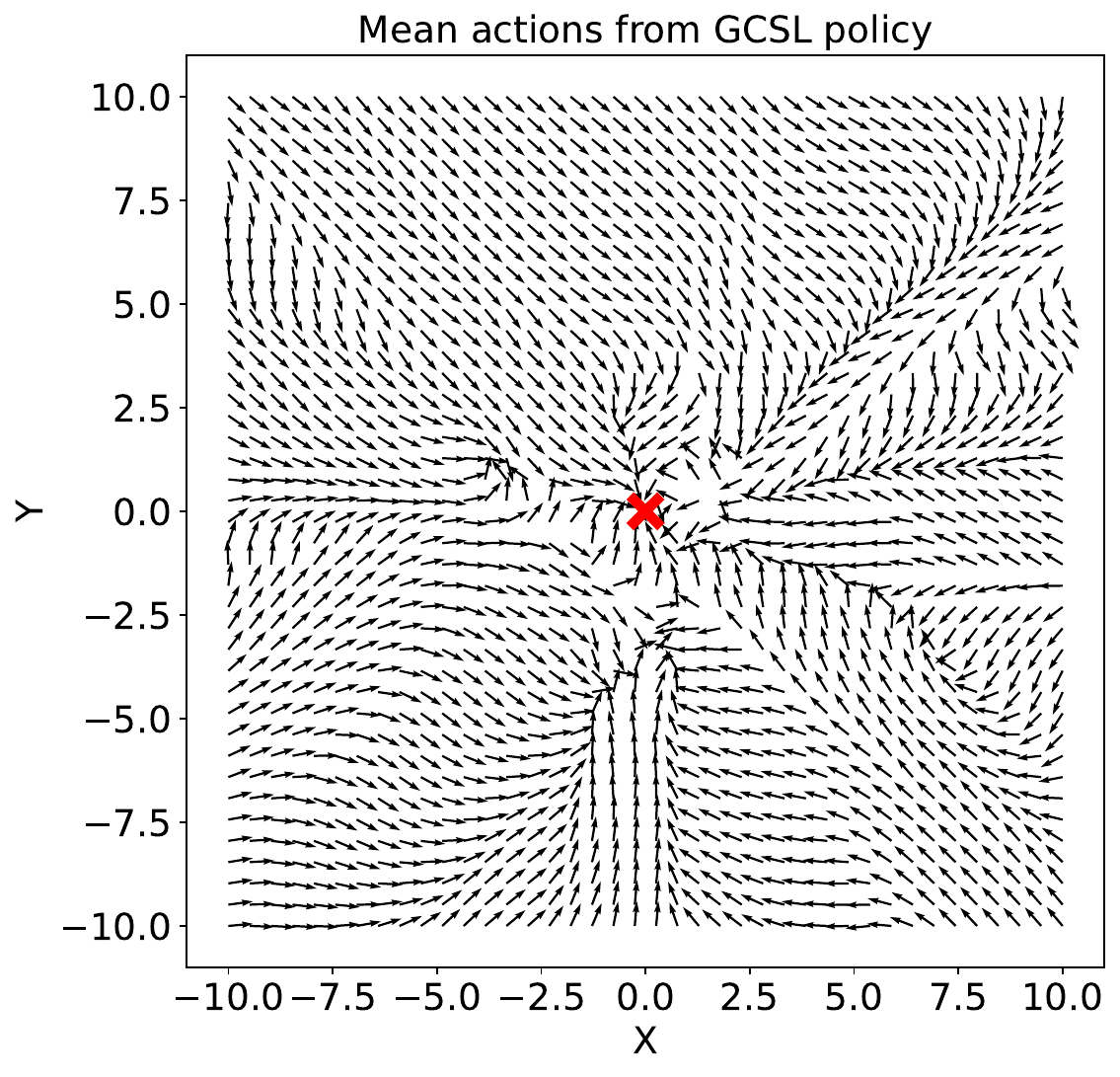}
      \caption{\small{}}
      \label{fig:diff_toy_gcsl}
  \end{subfigure}
  \vspace{-0.5em}
  \caption{\small{(a) Visualization of trajectories starting from the goal \textbf{\textcolor{red}{X}} generated during the forward process, (b) Predicted actions from policy trained via diffusion, (c) Predicted actions from policy trained using GCSL.}}
  \vspace{-1em}
  \label{fig:diff_toy}
\end{figure*}

The proof is provided in \Cref{sec:app_proof}.
Generalizing the above statement to a distribution over goals is given by the following corollary. Suppose we sample different datasets  $\mathcal{D}(g)$  for different goals $g \sim  p(g)$, where $\mathcal{D}(g)$ is produced from dataset $\mathcal{D}$ using hindsight relabeling -- that is we are considering all partial trajectories that pass through specified goal states. 
Additionally, we condition the policy on the goal $g$. Repeated application of the theorem above for $g \sim p(g)$ results in the following corollary. 

\begin{corollary}
\label{rmk}
Given a dataset $\mcD$ and target goal distribution $p(g)$, behavior cloning using a goal-conditioned policy $\theta^* = \argmax_\theta \mathbb{E}_{g\sim p(g), (s, a) \sim \mathcal{D}(g)} \left[ \log \pi_\theta(a | s, g)\right]$ maximizes a lower bound on the log-likelihood of the goal states $L = \mathbb{E}_{g \sim p(g), s_T \sim q(s_T \mid g)} \left[\log p_\theta(s_T)\right]$.
\end{corollary}

Similar to denoising diffusion models, we additionally condition the policy on the time horizon $h$ separating the current state and the goal state. In our experiments, the policy is parameterized as a diagonal Gaussian distribution,
\begin{align*}
    \pi_\theta(\cdot | s_t, g, h) = \mathcal{N}(\cdot | \mu_\theta(s_t, g, h), \sigma^2_\theta(s_t, g, h) \textbf{I})
\end{align*}
Note that most prior works that employ behavior cloning do not learn the variance term and minimize the mean squared error between observed and predicted actions. As shown in \Cref{sec:toy}, predicting the variance allows the policy to incorporate uncertainty in their action predictions. This is important in learning from a trajectory far from the goal state. 
Following \Cref{thm}, the policy is trained using behavior cloning to maximize the log probability of actions given by these transitions,
\begin{align}
\label{eq:bc}
    \theta^* = \argmax_\theta \mathbb{E}_{(s_t,a_t,g,h) \sim \mathcal{D}_\textrm{new}} \left[\log \pi_\theta(a_t | s_t, g, h)\right]
\end{align}


\subsection{An Illustrative Example}
\label{sec:toy}
We illustrate this concept using a simple 2D navigation environment consisting of an agent tasked with reaching a target goal state. The states are the agent's $(x,y)$ coordinates, and actions represent the displacement in the $x$ and $y$ directions, normalized to be unit length. For these experiments, the goal state during training is fixed to $g=(0,0)$, and the initial agent state is sampled uniformly at random. The forward diffusion process is constructed by taking random actions starting from the goal state. For this simple environment, the forward process transitions $q(s_t|s_{t+1})$ are simply displacement vectors based on the random action $a_t$.

\Cref{fig:diff_toy_traj} visualizes the trajectories representing the forward diffusion process, obtained by taking random actions starting from the goal for $T=50$ steps. We set $s_T = g$ and the random state reached at the end of the diffusion is $s_1$. The policy is parameterized as a diagonal Gaussian distribution and is trained to reverse these trajectories by conditioning on the final goal or some future state in the trajectory. For any state $s_t$ in the trajectory, we maximize the likelihood of the observed action $a_t$, conditioned on any future (goal) state $g'$, given the time horizon $h$ separating the two states in the observed trajectory.  More formally, the policy parameters are trained by optimizing $\theta^* = \arg\max_\theta \mathbb{E}_{(s_t, a_t, g') \sim \tau} \log \pi_\theta(a_t|s_t, g', h)$ where $g' = \phi(s_i)$ and $h = i-t$ for $t < i \leq T$. 

\Cref{fig:diff_toy_main} visualizes the actions sampled from the trained policy for different states when conditioned on the goal $g=(0,0)$ using an input time horizon of one. For comparison, \Cref{fig:diff_toy_gcsl} visualizes the trained policy using GCSL \citep{ghosh2020learning}, which uses hindsight relabeling to train policies on data collected by the agent itself. Both methods were trained for $100k$ policy updates. The policy learned via diffusion learns the optimal path, which takes the shortest time to reach the goal. We extend this example to more complex settings, including multiple goals and a walled environment, in \Cref{app:add_toy}. 
Note that the trained policy is analogous to the score function for diffusion models, and the action is analogous to the predicted noise, which serves to ``denoise” the states towards the goal. 

\begin{figure}[h]
  \centering
  \begin{subfigure}[b]{0.242\linewidth}
      \centering
      \includegraphics[trim={4em 2.5em 0 2.5em},clip,width=\textwidth]{images/diff_100000_t_1_0.pdf}
      \caption{\texttt{h=1}}
      \label{fig:diff_toy_t1}
  \end{subfigure}
  \begin{subfigure}[b]{0.242\linewidth}
      \centering
      \includegraphics[trim={4em 2.5em 0 2.5em},clip,width=\textwidth]{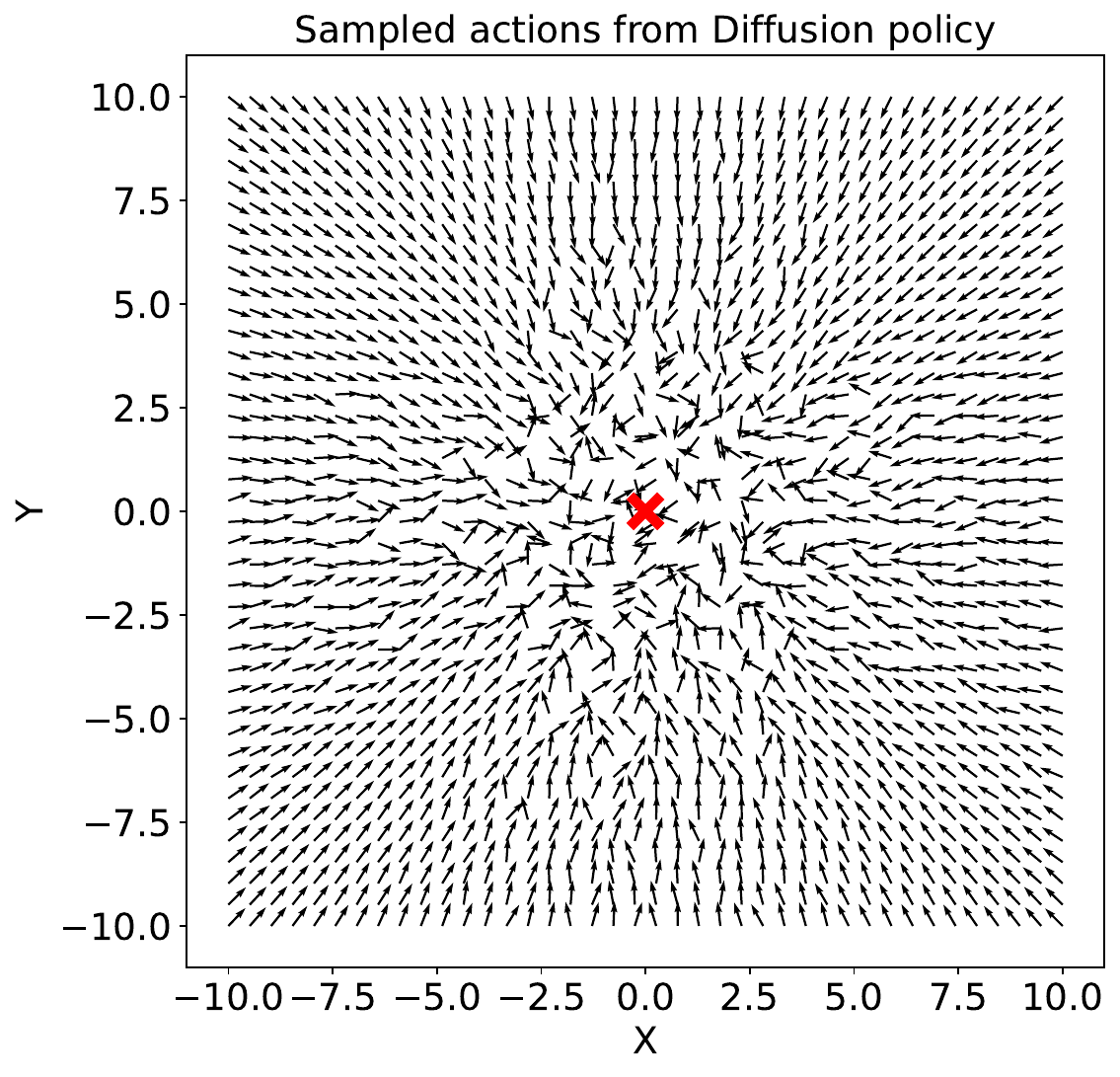}
      \caption{\texttt{h=5}}
      \label{fig:diff_toy_t5}
  \end{subfigure}
  \begin{subfigure}[b]{0.242\linewidth}
      \centering
      \includegraphics[trim={4em 2.5em 0 2.5em},clip,width=\textwidth]{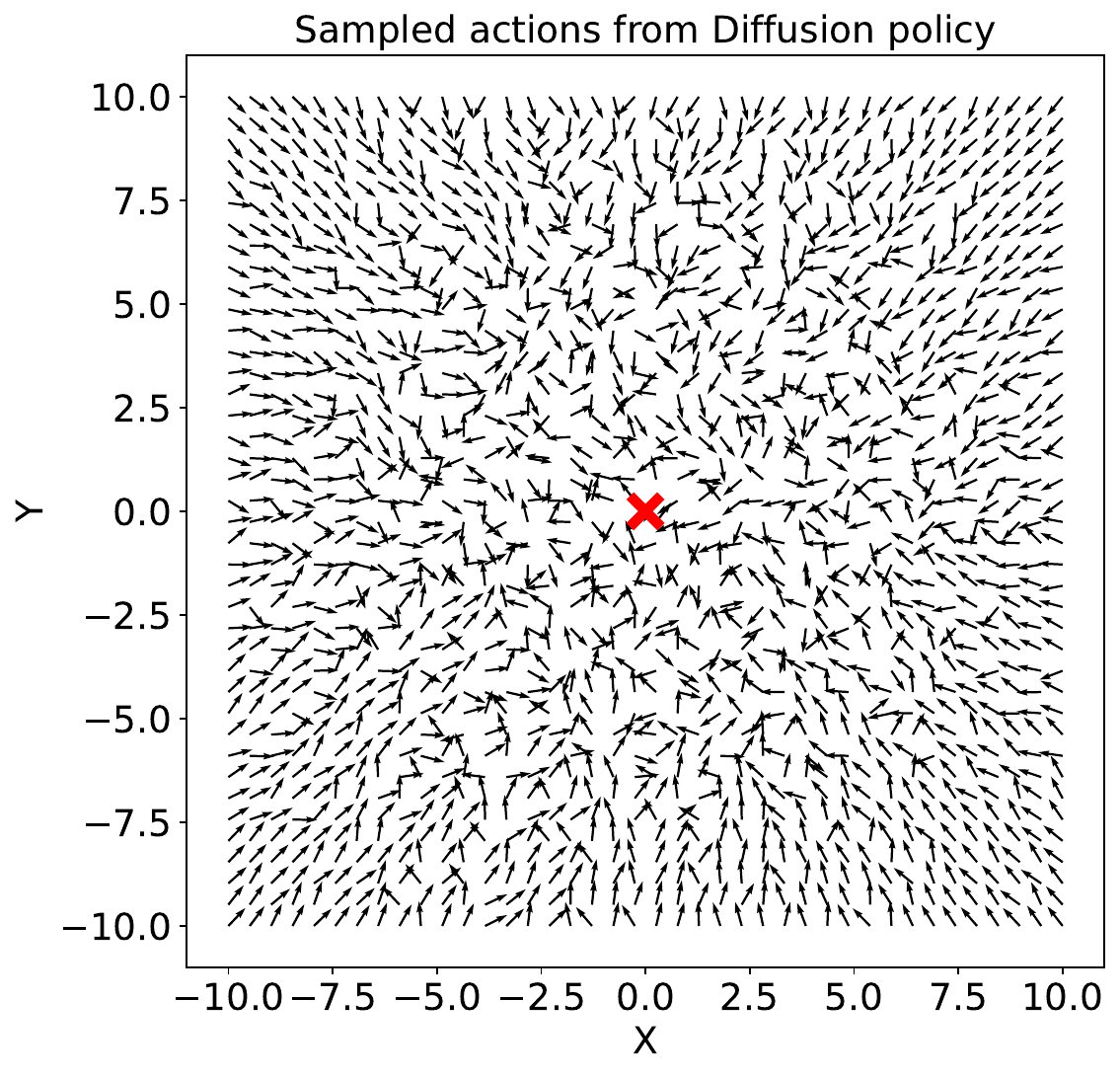}
      \caption{\texttt{h=10}}
      \label{fig:diff_toy_t10}
  \end{subfigure}
  \begin{subfigure}[b]{0.242\linewidth}
      \centering
      \includegraphics[trim={4em 2.5em 0 2.5em},clip,width=\textwidth]{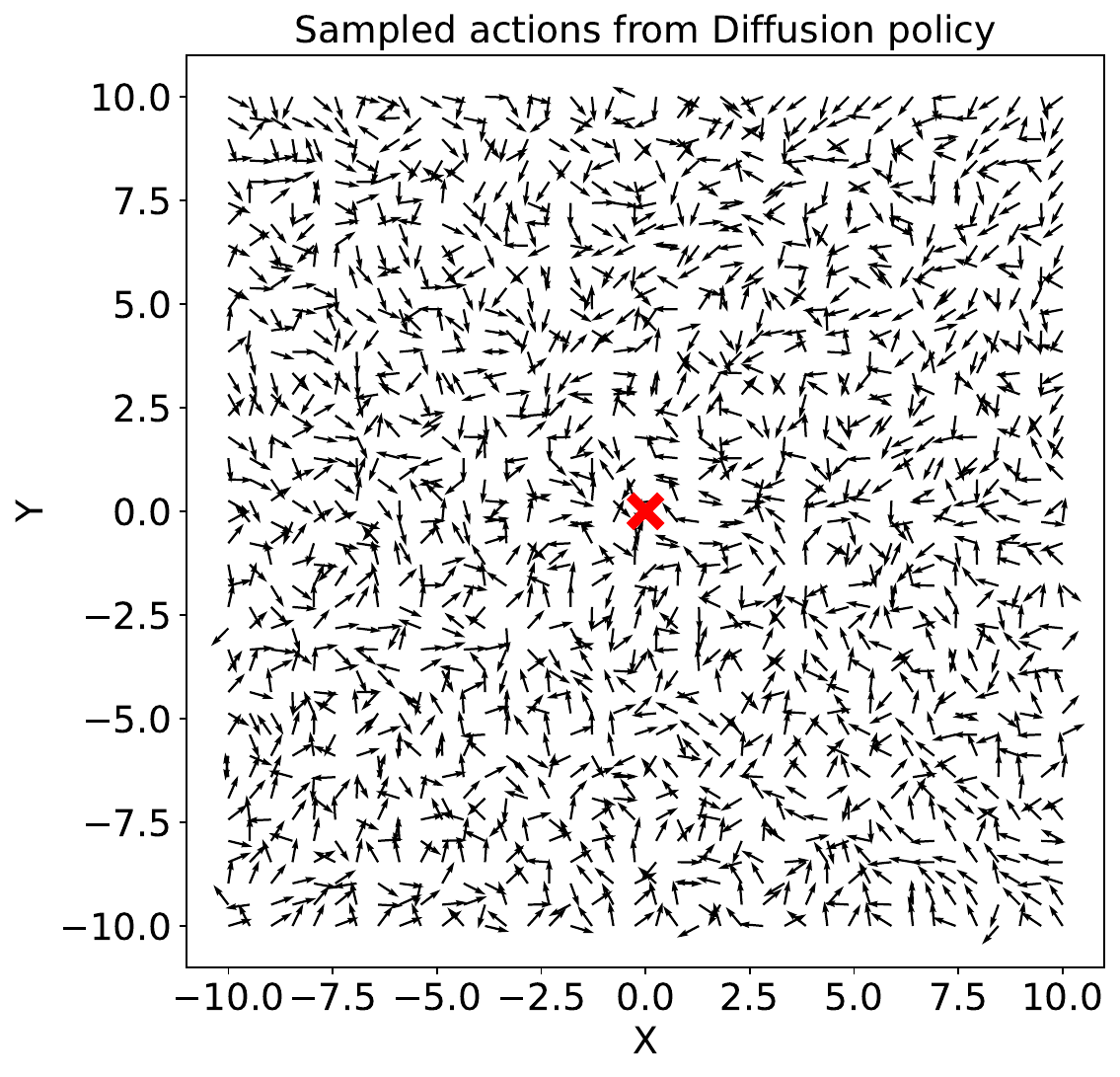}
      \caption{\texttt{h=20}}
      \label{fig:diff_toy_t20}
  \end{subfigure}
  \caption{\small{Actions sampled from the trained policy, showcasing the effect of time horizon during evaluation.}}
  \vspace{-1em}
  \label{fig:diff_toy_horizon}
\end{figure}

For policy evaluation, the choice of time horizon $h$ to be used is not immediately obvious. We investigate the effect of changing the time horizon, shown in \Cref{fig:diff_toy_horizon}. For $h=1$, the policy always takes the most direct path to the goal regardless of the input state. For larger values of the time horizon, the policy has a high variance close to the goal and a low variance for the optimal action further away. In \Cref{sec:ablation}, we perform ablations to further investigate its effect on performance.

\begin{figure}[h]
  \centering
  \begin{subfigure}[b]{0.242\linewidth}
      \centering
      \includegraphics[trim={4em 2.5em 0 1em},clip,width=\linewidth]{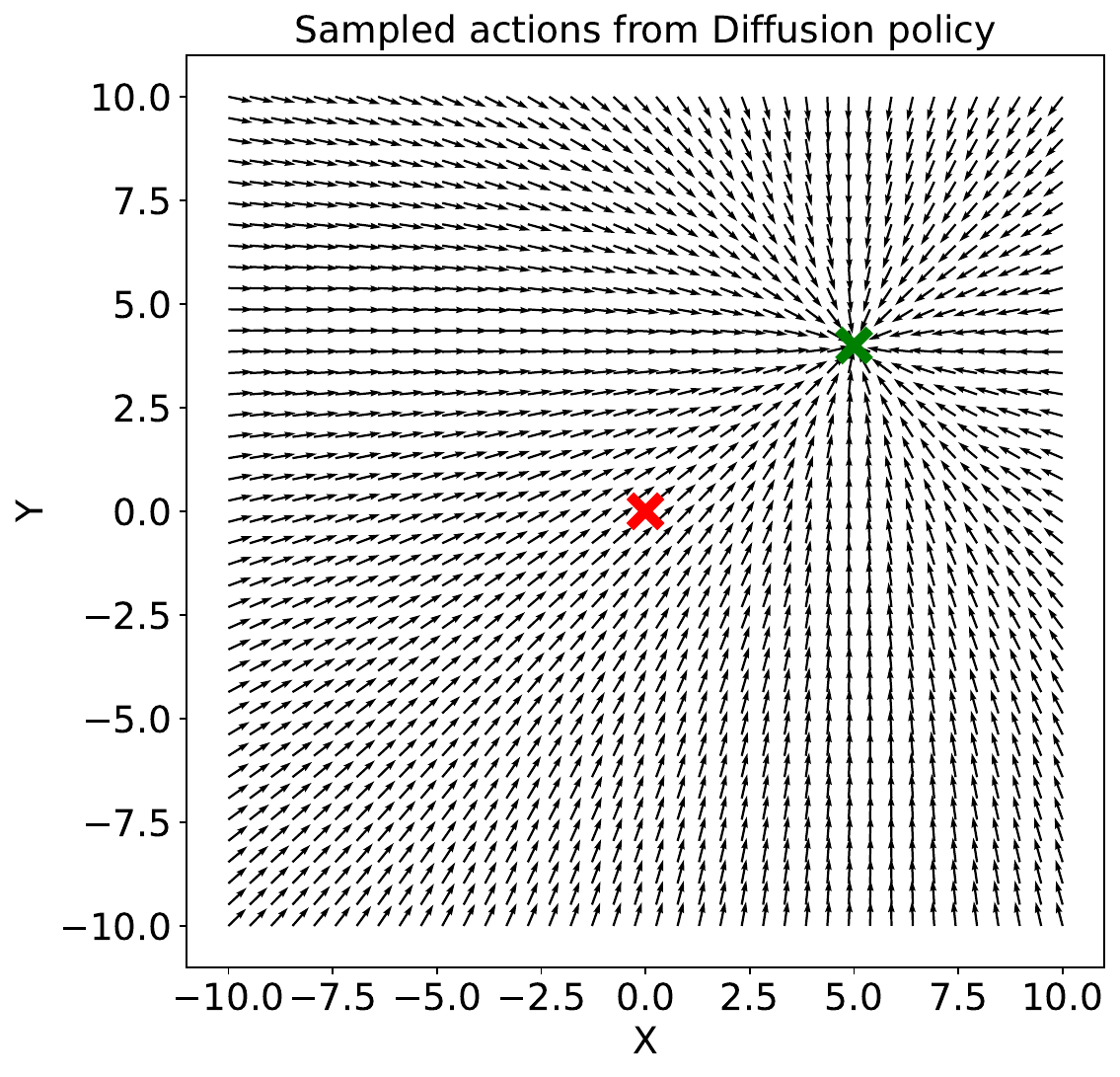}
  \end{subfigure}
  \begin{subfigure}[b]{0.242\linewidth}
      \centering
      \includegraphics[trim={4em 2.5em 0 1em},clip,width=\linewidth]{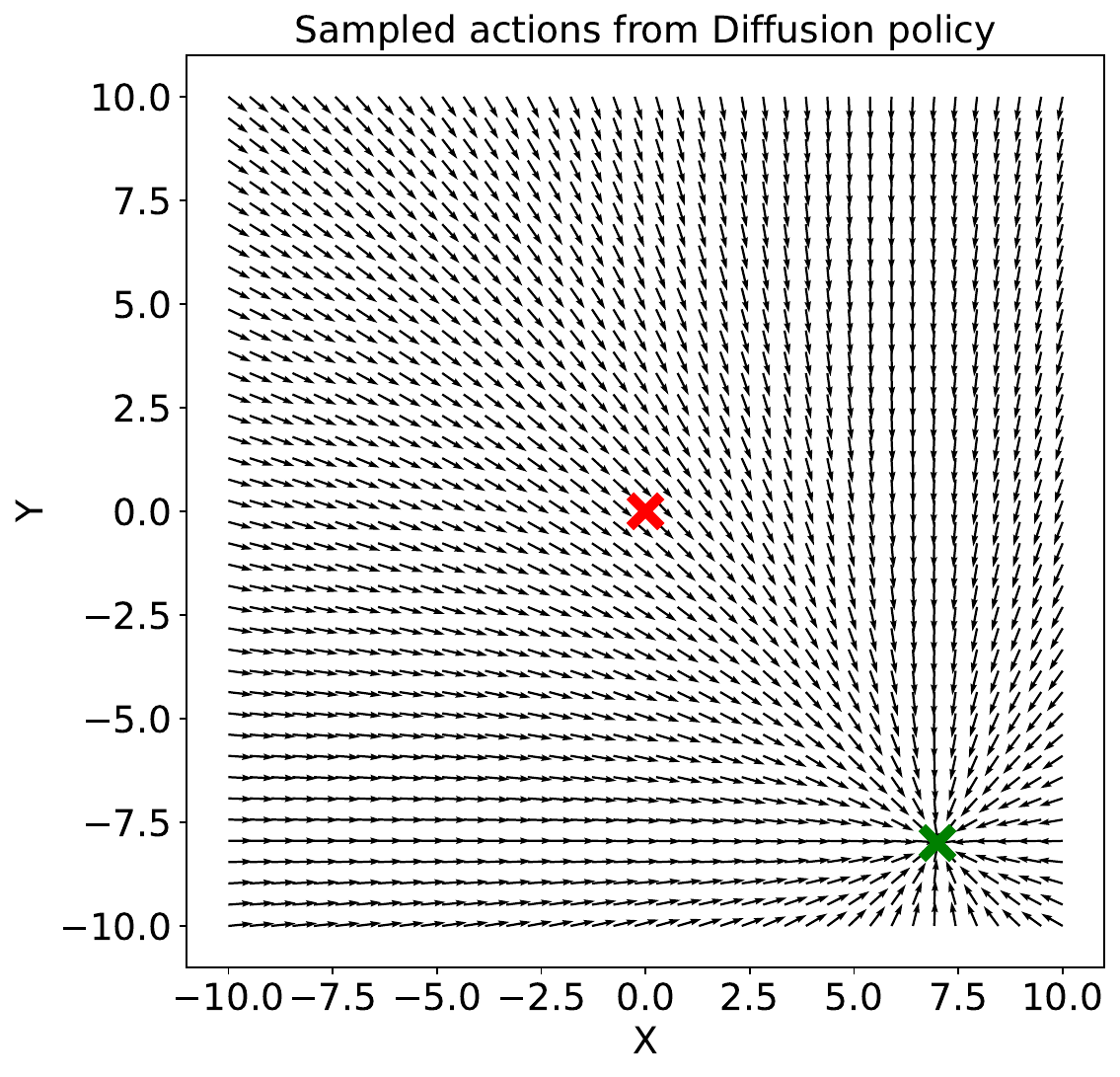}
  \end{subfigure}
  \begin{subfigure}[b]{0.242\linewidth}
      \centering
      \includegraphics[trim={4em 2.5em 0 1em},clip,width=\linewidth]{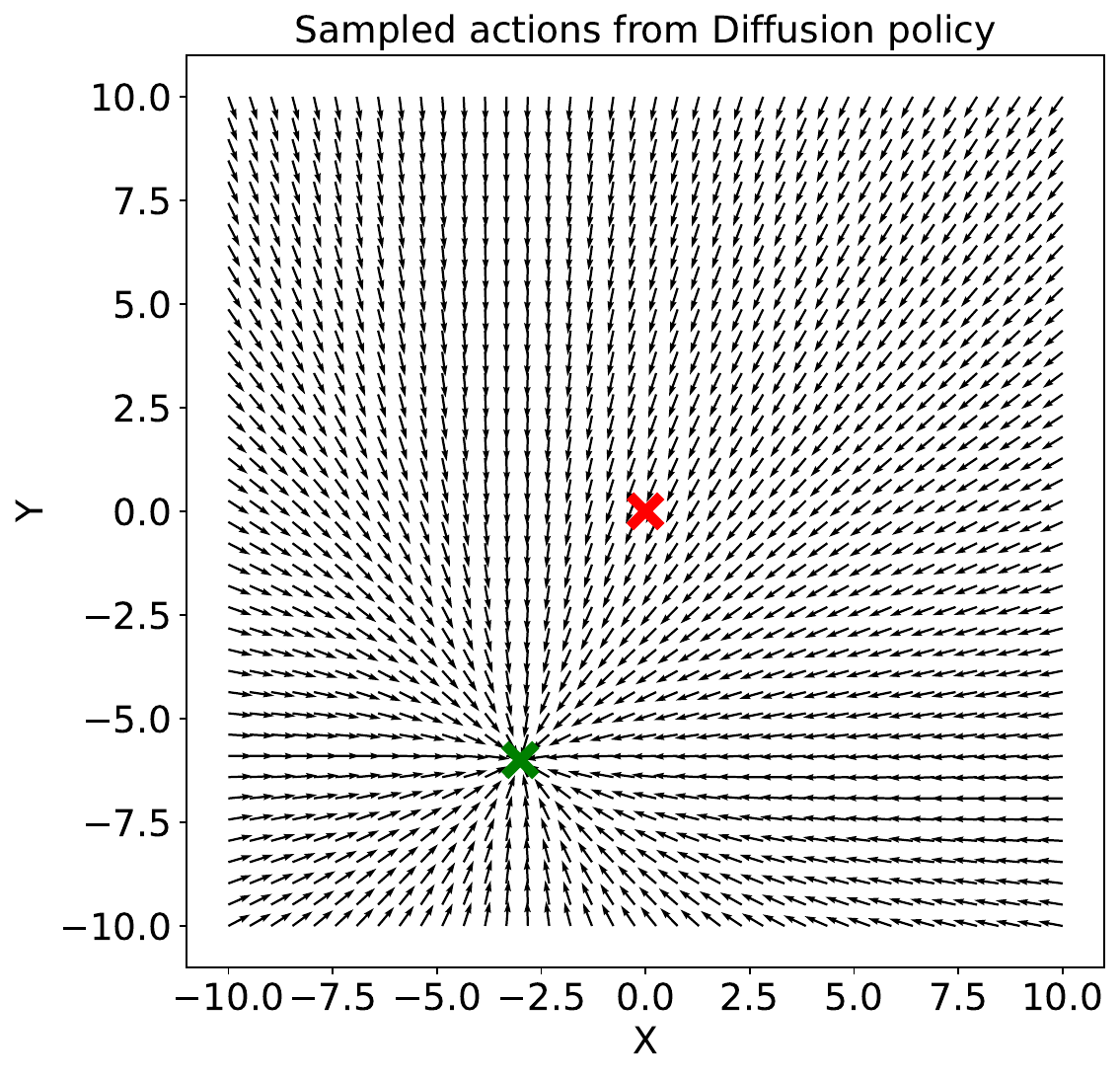}
  \end{subfigure}
  \begin{subfigure}[b]{0.242\linewidth}
      \centering
      \includegraphics[trim={4em 2.5em 0 1em},clip,width=\linewidth]{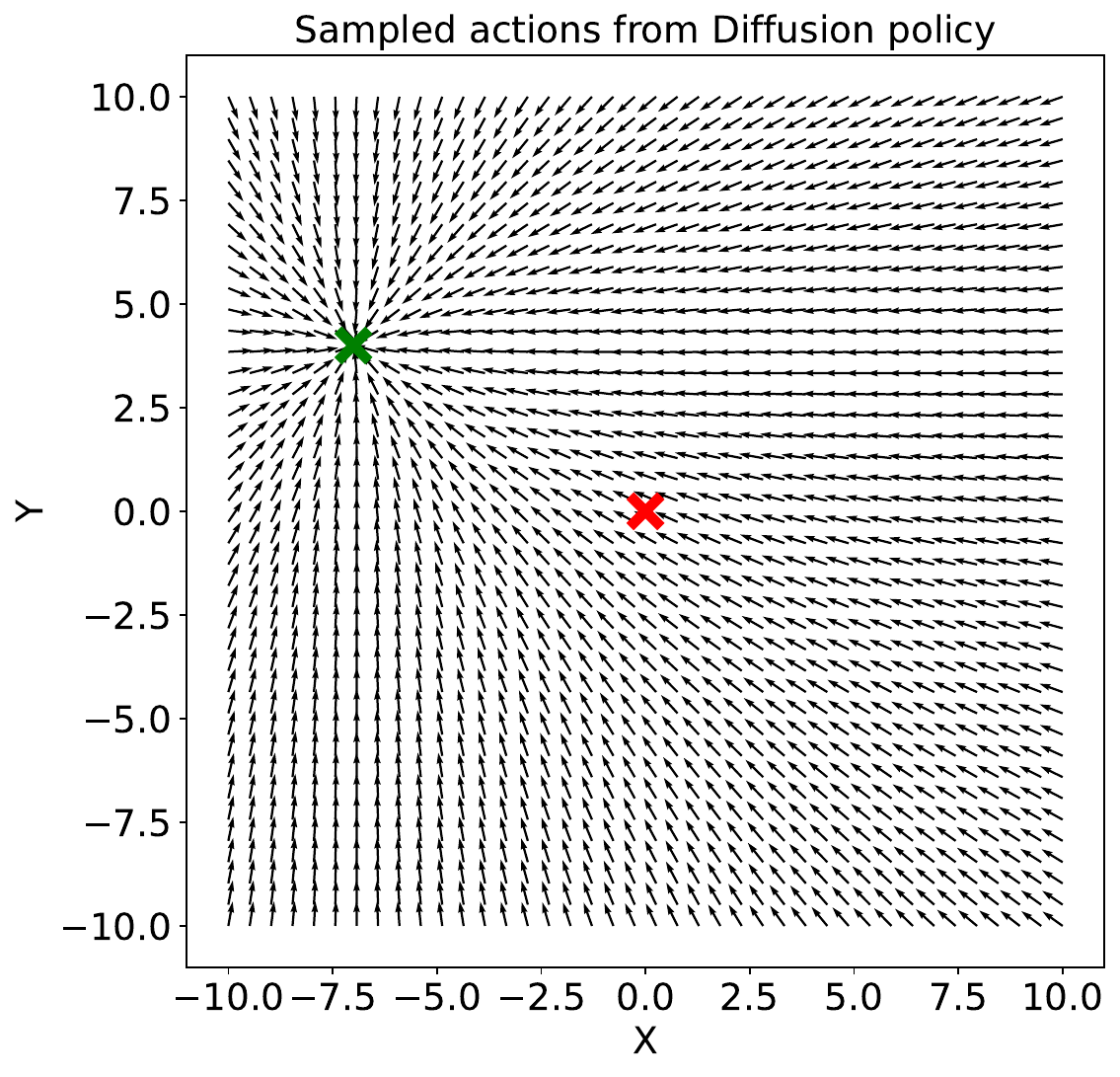}
  \end{subfigure}
  \newline
  \begin{subfigure}[b]{0.242\linewidth}
      \centering
      \includegraphics[trim={4em 2.5em 0 1em},clip,width=\linewidth]{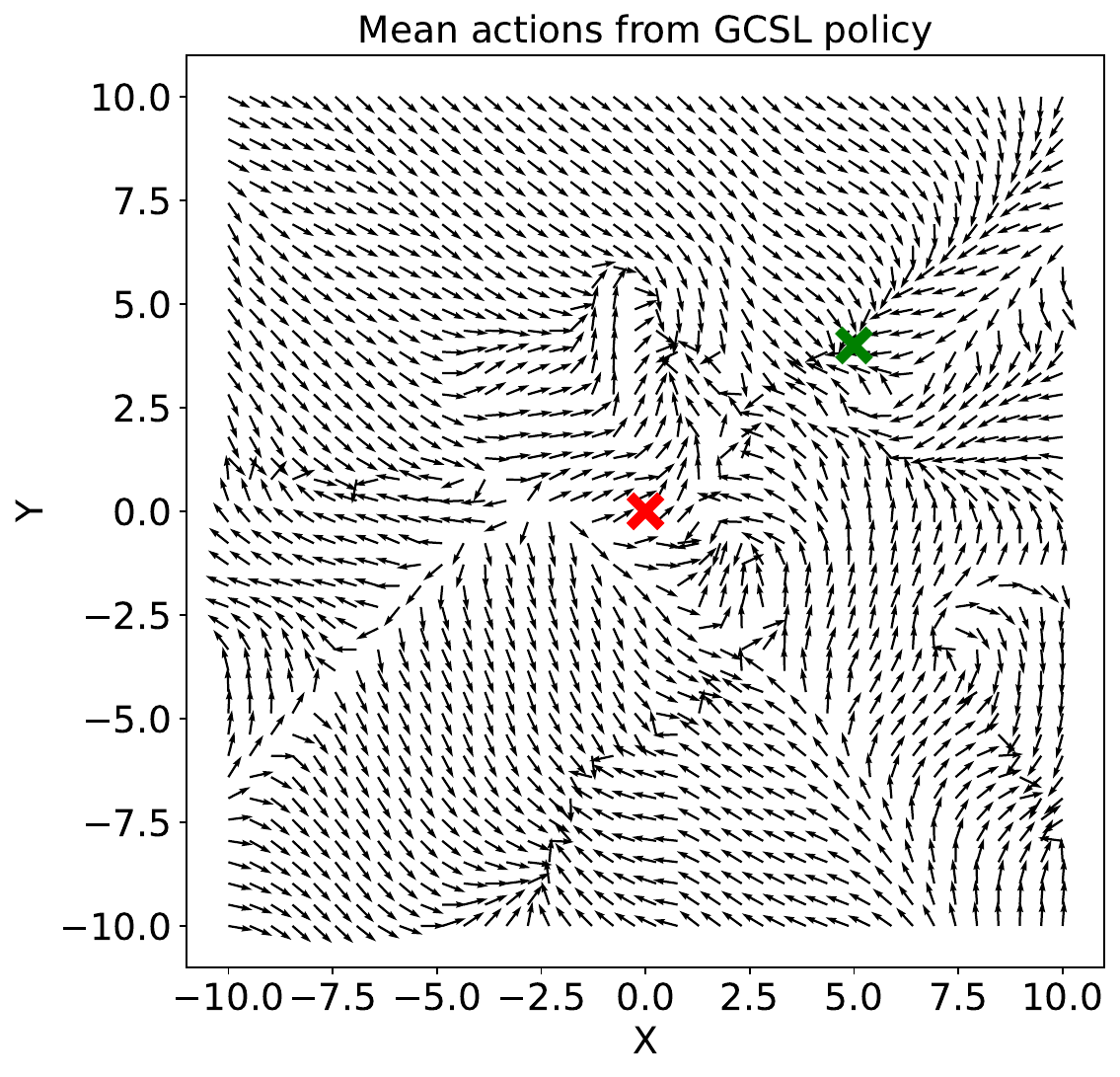}
  \end{subfigure}
  \begin{subfigure}[b]{0.242\linewidth}
      \centering
      \includegraphics[trim={4em 2.5em 0 1em},clip,width=\linewidth]{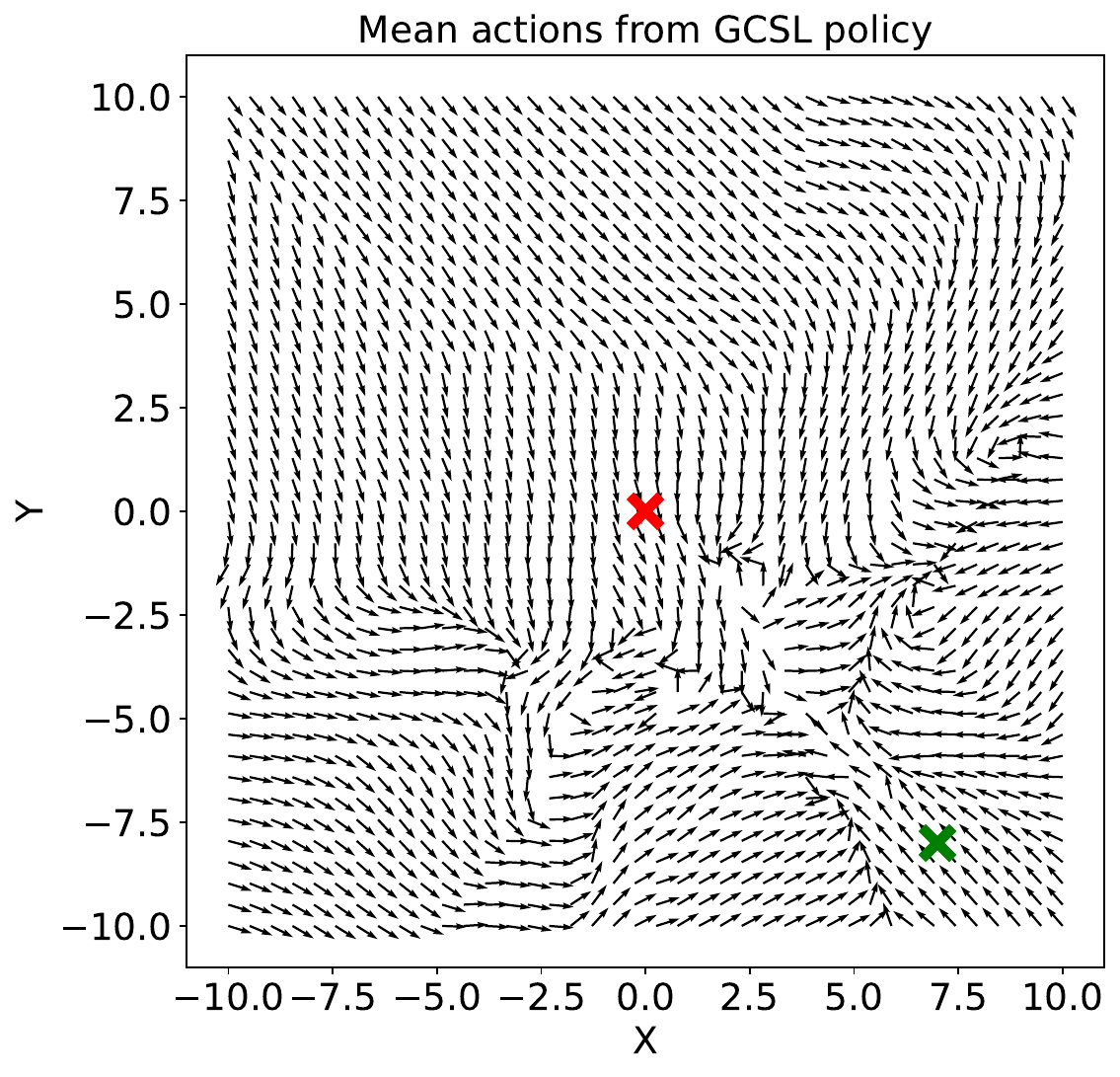}
  \end{subfigure}
  \begin{subfigure}[b]{0.242\linewidth}
      \centering
      \includegraphics[trim={4em 2.5em 0 1em},clip,width=\linewidth]{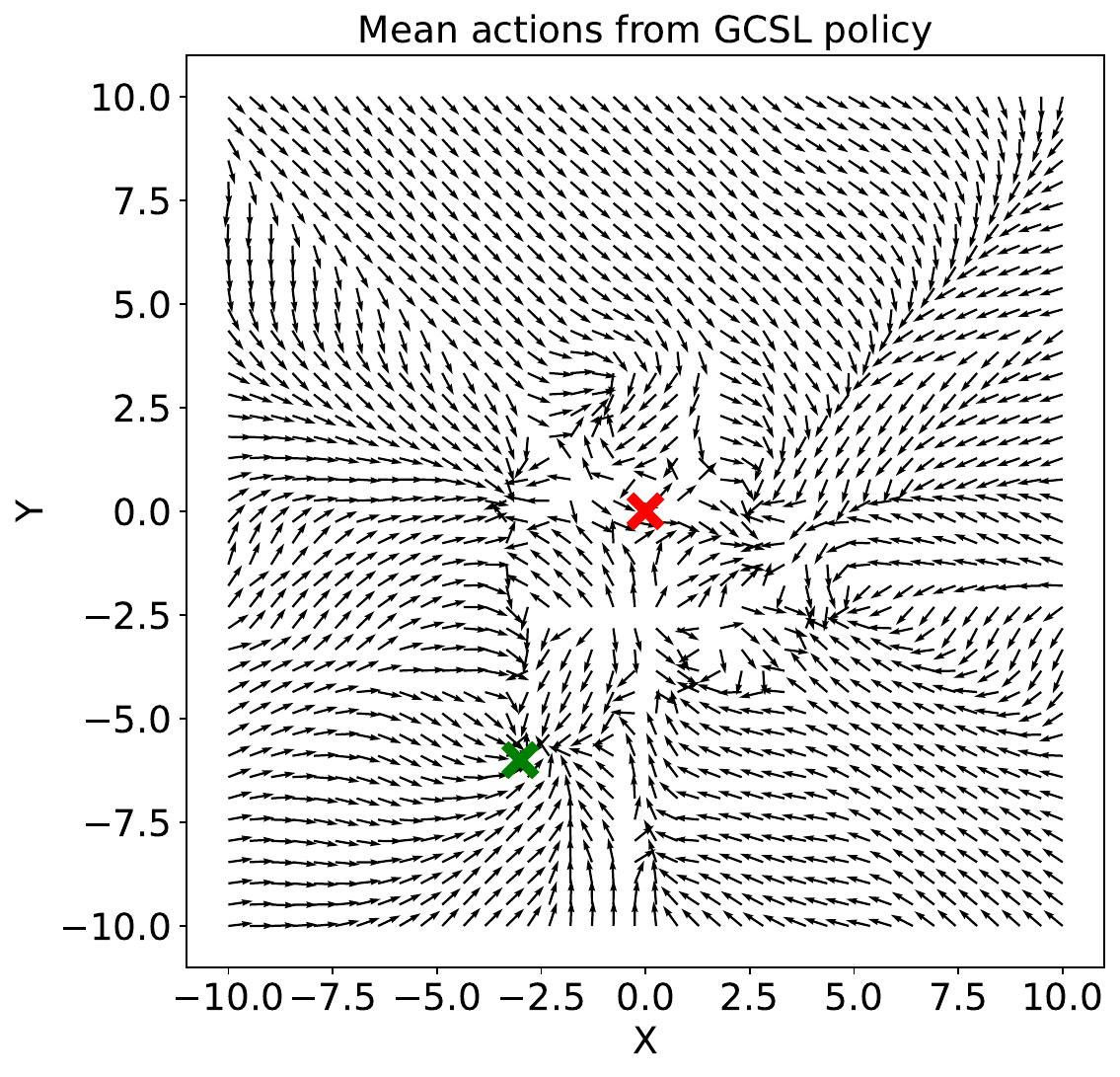}
  \end{subfigure}
  \begin{subfigure}[b]{0.242\linewidth}
      \centering
      \includegraphics[trim={4em 2.5em 0 1em},clip,width=\linewidth]{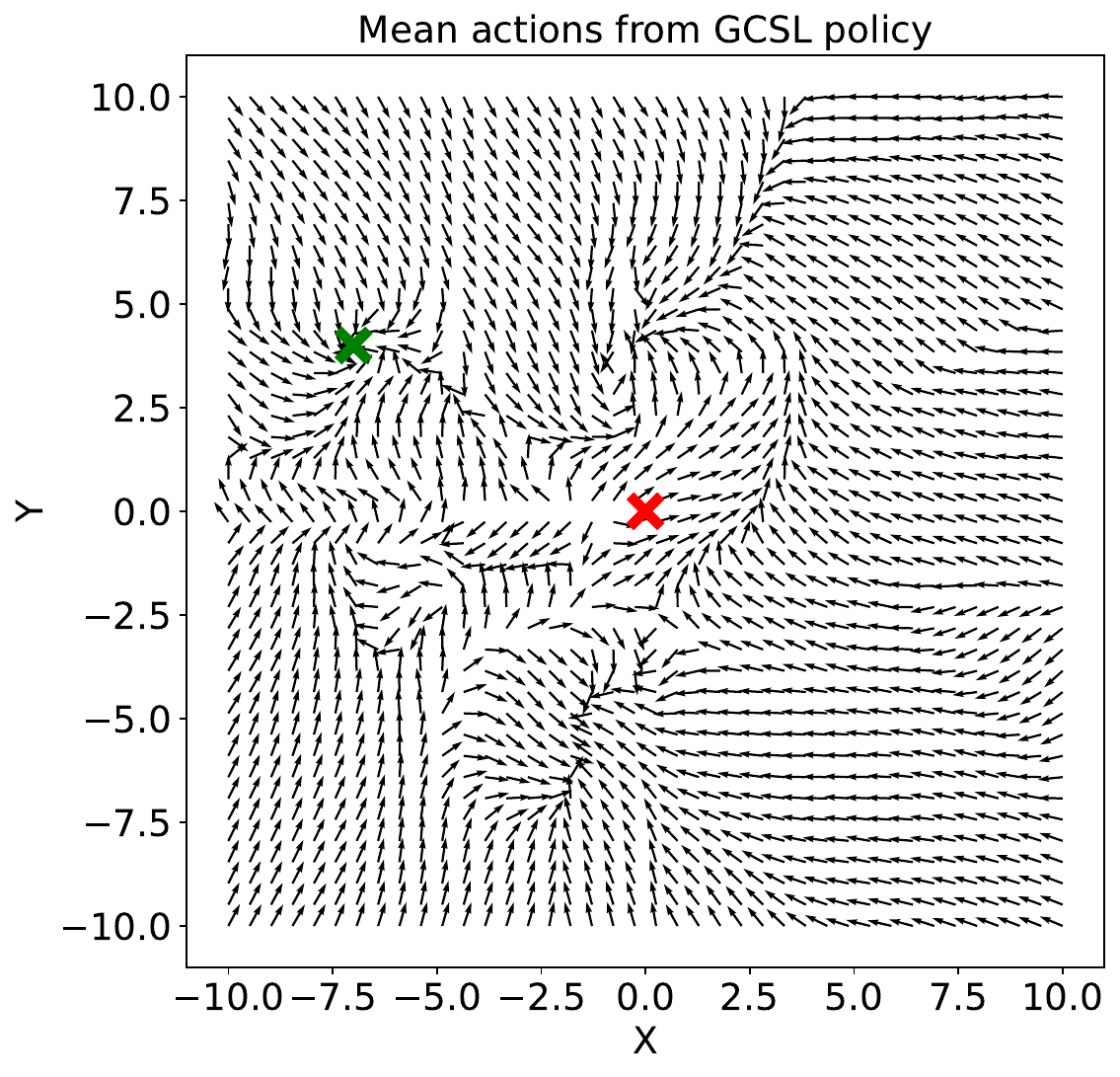}
  \end{subfigure}
  \caption{\small{Evaluating the trained policy on out-of-distribution goals. Red \textbf{\textcolor{red}{X}} denotes the goal during training, and green \textbf{\textcolor{teal}{X}} denotes the goal used for evaluation. \textbf{Top:} Merlin; \textbf{Bottom:} GCSL.}}
  \vspace{-1em}
  \label{fig:diff_toy_ood}
\end{figure}

We then test the generalization capabilities of our approach by evaluating the policy on out-of-distribution goals. During training, the goal is fixed to be at the center but during evaluation, we condition the policy on random goals. \Cref{fig:diff_toy_ood} shows that the policy can effectively generalize to different goals due to hindsight relabeling. Note that GCSL also uses hindsight relabeling, but unlike Merlin, it often takes sub-optimal paths to reach the goal.

\section{Generating Reverse Trajectories}
\label{sec:gen_traj}
In this section, we discuss the application of Merlin to offline datasets. A straightforward application of Merlin to offline data involves I) loading the dataset to a replay buffer and sampling trajectories in reverse. To create more varied diffusion trajectories beyond the dataset, we propose two potential ways to construct the forward process: II) using a reverse model to simulate diffusion trajectories and; III) a novel non-parametric method that stitches trajectories to create diverse diffusion paths.

\subsection{Reverse Parametric Model}
\label{sec:model}

One method to simulate the forward (diffusion) process $q(s_t | s_{t+1})$ is to train a reverse (RL) model that takes states as inputs and produces candidate previous actions and previous states. We can break this into two parts: (a) the \emph{reverse policy} that given $s_{t+1}$ produces the previous action $a_t$; (b) the \emph{reverse dynamics model} produces the previous state $s_{t}$ given $s_{t+1}$ and $a_t$.

\paragraph{Reverse policy} To generate diverse candidate actions for reverse rollouts, we follow the procedure described in \citet{wang2021offline}. The reverse policy is parameterized as a conditional variational autoencoder (CVAE), consisting of an action encoder $E_\omega$ that outputs a latent vector $z$, and an action decoder $D_\xi$ which reconstructs the action given latent vector $z$. The reverse policy is trained by maximizing the variational lower bound, 
\begin{align}
\label{eq:rev_policy}
    \mathcal{L}(\omega, \xi) &= \mathbb{E}_{(s_t,a_t,s_{t+1}) \sim \mathcal{D}} \left[ D_{KL}\left(E_\omega(s_{t+1},a_t) || \mathcal{N}(0, \mathbf{I})\right)\right]\nonumber\\
    &\;+ \mathbb{E}_{z \sim E_\omega(s_{t+1},a_t)} \left[ \left(a_t - D_\xi(s_{t+1},z)\right)^2\right],
\end{align}

\paragraph{Reverse Dynamics Model}
We parameterize the dynamics model, denoted as $f_\psi$, using a Gated Recurrent Unit (GRU) \citep{cho2014learning} and optimize the model parameters by minimizing the mean squared error,
\begin{align}
\label{eq:rev_model}
    \mathcal{L}(\psi) = \mathbb{E}_{\tau \sim \mathcal{D}} \|s_t - f_\psi(a_t,s_{t+1},\ldots,s_T)\|_2^2,
\end{align}
Using such a dynamics model moves beyond the Markovian assumption, as mentioned in \Cref{sec:processes}. In our implementation, the reverse dynamics model $f_\psi$ predicts the state difference $s_\Delta = s' - s$ instead of the absolute state $s$. For image inputs, we use a Convolutional Neural Network (CNN) to produce latent representations of the images, and predict the state difference in the latent space. Full details of the model architecture and hyperparameters are provided in \Cref{app:implementation}. 


\subsection{Reverse Non-Parametric Model}
\label{sec:traj}

\begin{figure}[h]
    \centering
    \includegraphics[trim={0em 10em 0em 10em},clip,width=0.9\linewidth]{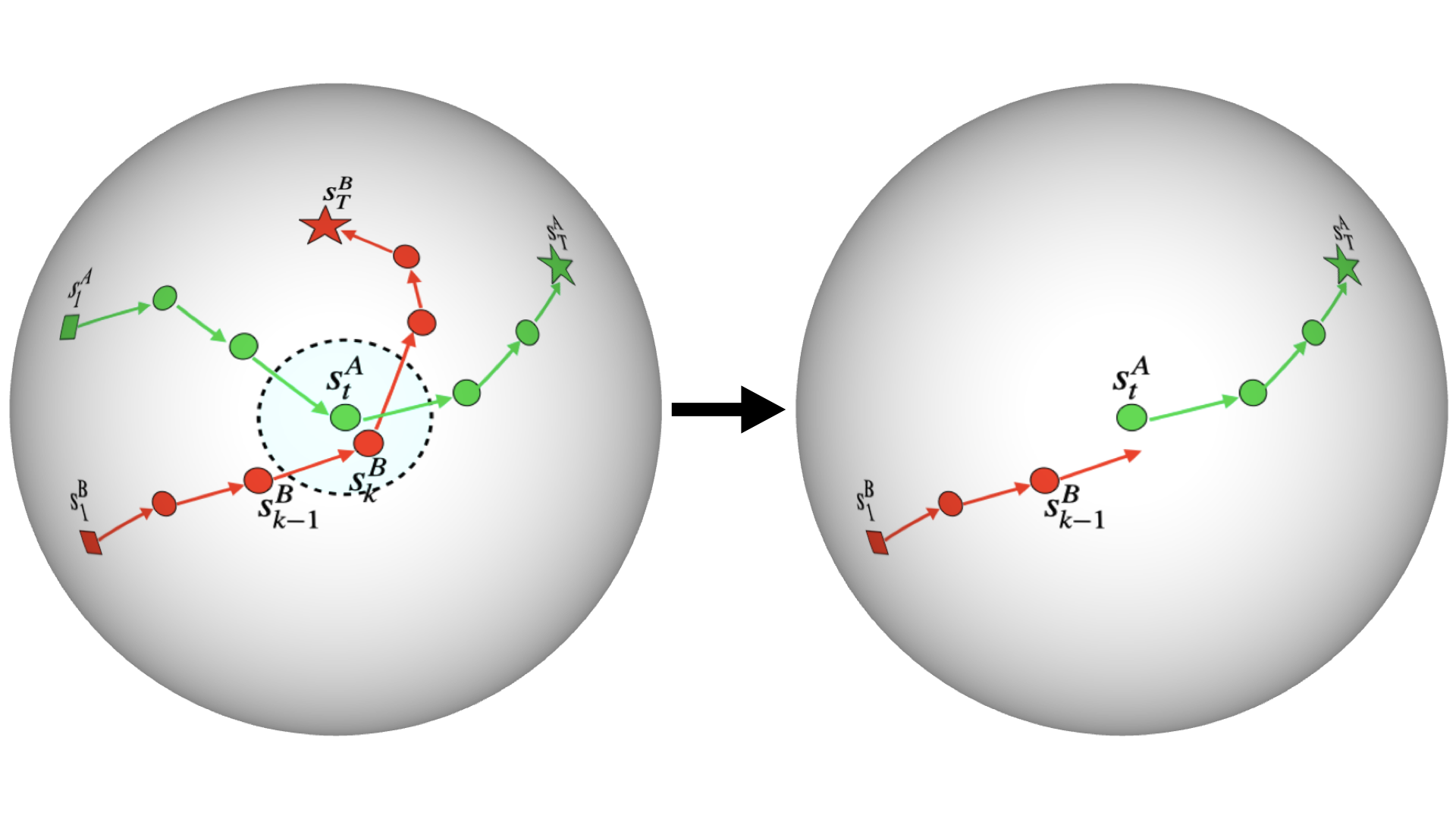}
    \captionof{figure}{\small{Trajectory stitching, where nearest-neighbors are computed in learned latent space such that connected states are nearby.}}
    \label{fig:traj_stitch}
    \vspace{-1em}
\end{figure}

As an alternative to the model-based approach, we introduce a trajectory stitching operation to generate state-goal pairs \emph{across} trajectories. The basic idea behind this operation is that if two states from different trajectories are very close to each other, then the sub-trajectory leading up to one of these states is also a reasonably good path to reach the other state; see \cref{fig:traj_stitch}. One caveat with this method is that nearby states may not necessarily be connected (think of nearby positions on two sides of a barrier.) To address this issue, we learn representations based on state connectivity using contrastive learning \citep{oord2018representation}. The positive pairs are consecutive state pairs and negative pairs are randomly sampled state pairs from the dataset. The loss function to train the contrastive encoder $h_\phi$ is given by,
\begin{align}
    \mathcal{L}(\phi) = \mathbb{E}_{(s,s') \sim \mathcal{D}}\left[-\log\frac{\exp(h_\phi(s)^\top h_\phi(s'))}{\sum_{s'_\textrm{neg} \in \mathcal{D}} \exp(h_\phi(s)^\top h_\phi(s'_\textrm{neg}))}\right]
\end{align}
We then perform the trajectory stitching in the latent space, since nearby states in this space are more likely to be connected. 

Next, we have to choose a metric and corresponding threshold, so as to identify which states can be stitched across trajectories. Since nearest-neighbors are computed based on contrastive representations, we choose cosine similarity as the metric and a threshold of $0.9999$. 
We map all states in the dataset to the latent space and construct a ball tree for all the state representations to allow quick nearest-neighbor search. For a $d$-dimensional latent space with $N$ samples, the ball-tree can be efficiently queried with the time complexity $O(d\log N)$. We sample random goal states from the dataset and iteratively add the previous state-action to the new trajectories. At each step, we query the ball tree for the nearest neighbor and if the cosine similarity is greater than the threshold, we switch to the trajectory corresponding to the neighbor state, otherwise, we stick to the same trajectory. \Cref{app:merlinnp} provides implementation details as well as a discussion on the time complexity of tree search and choice of threshold.

\subsection{Theoretical Implications}
Our theoretical justification given in \Cref{thm} relates likelihood maximization under the reverse diffusion process to behavior cloning for a given dataset. Using the methods above to generate additional forward diffusion trajectories results in a different augmented dataset. Our theorem still applies in the sense that for the augmented dataset, likelihood maximization is still related to behavior cloning. The only difference is that the augmented dataset has potentially more diverse and informative diffusion paths to train the policy.

\begin{algorithm}[h!]
    \centering
    \begin{algorithmic}
    \label{algo:merlin}
        \STATE \textbf{Input:} Dataset $\mathcal{D}$, hindsight ratio $p$, number of training steps $N$, \textcolor{purple}{number of new trajectories to collect $M$}.
        \STATE \textbf{Output: } Policy $\pi_\theta$
        \STATE \textcolor{red}{Train $f_\psi$, $E_\omega$, $D_\xi$ on $\mathcal{D}$ by minimizing Equation (6,7)}
        \STATE \textcolor{blue}{Train $h_\phi$ on $\mathcal{D}$ by minimizing Equation (8)}
        \STATE \textcolor{blue}{Construct ball tree $T$ for all states encoded using $h_\phi$}
        \STATE \texttt{\# Simulate forward diffusion process}
        \STATE \textcolor{red}{Collect $M$ trajectories in $\mathcal{D}_\textrm{new}$ using $f_\psi$, $E_\omega$, $D_\xi$.}
        \STATE \textcolor{blue}{Collect $M$ trajectories in $\mathcal{D}_\textrm{new}$ using tree $T$}
        \STATE \textcolor{purple}{$\mathcal{D} \gets \mathcal{D}_\textrm{new} \cup \mcD$}
        \STATE \texttt{\# Train policy via reverse diffusion}
        \FOR{$n\gets 1$ {\bfseries to} $N$}
        \STATE Sample batch $(s,a,g)$ from $\mathcal{D}$
        \STATE Relabel fraction $p$ of batch 
        \STATE Update policy $\pi_\theta$ as per \Cref{eq:bc}
        \ENDFOR
        \STATE \textbf{Return:} $\pi_\theta$
    \end{algorithmic}
    \caption{Merlin algorithm. \textcolor{red}{Red} and \textcolor{blue}{blue} statements apply only for Merlin-P and Merlin-NP, respectively. \textcolor{purple}{Purple} statements apply to both.}
\end{algorithm}

\section{Experiments}
\label{sec:exp}

\begin{figure*}[t!]
\begin{minipage}{0.7\textwidth}
\captionof{table}{\small{Discounted returns for state-space input, averaged over 10 seeds.}}
\vspace{-1em}
\begin{center}
\begin{adjustbox}{width=\textwidth}
\begin{tabular}{|ll|rrr|rrrr|rrr|}
\hline \\[-2.35ex]
\multicolumn{2}{c|}{\multirow{2}{*}{\textbf{Task Name}}} & \multicolumn{3}{c|}{\textbf{Ours}} & \multicolumn{4}{c|}{\textbf{Offline GCRL}} & \multicolumn{3}{c|}{\textbf{Diffusion-based}}\\
\rule{0pt}{2ex}
& & \multicolumn{1}{c}{\textbf{Merlin}} & \multicolumn{1}{c}{\textbf{Merlin-P}} & \multicolumn{1}{c|}{\textbf{Merlin-NP}} & \multicolumn{1}{c}{\textbf{GoFAR}} & \multicolumn{1}{c}{\textbf{WGCSL}} & \multicolumn{1}{c}{\textbf{GCSL}} & \multicolumn{1}{c|}{\textbf{AM}} & \multicolumn{1}{c}{\textbf{DD}} & \multicolumn{1}{c}{\textbf{g-DQL}} & \multicolumn{1}{c|}{\textbf{BESO}} \\
\hline \rule{0pt}{3ex}
\multirow{10}{*}{\rotatebox[origin=c]{90}{\textbf{Expert}}}
& PointReach        & 29.26\scriptsize{$\pm$0.04} & \textbf{29.34}\scriptsize{$\pm$0.15} & \textbf{29.34}\scriptsize{$\pm$0.05} & 27.18\scriptsize{$\pm$0.65} & 25.91\scriptsize{$\pm$0.87} & 22.85\scriptsize{$\pm$1.26} & 26.14\scriptsize{$\pm$1.11} & 15.03\scriptsize{$\pm$0.88} & 28.65\scriptsize{$\pm$0.44} & 29.10\scriptsize{$\pm$0.28} \\
& PointRooms        & 25.38\scriptsize{$\pm$0.37} & 25.25\scriptsize{$\pm$0.27} & \textbf{25.63}\scriptsize{$\pm$0.32} & 20.40\scriptsize{$\pm$1.00}  & 19.90\scriptsize{$\pm$0.99} & 18.28\scriptsize{$\pm$2.29}  & 23.24\scriptsize{$\pm$1.58} & 10.84\scriptsize{$\pm$2.67} & \textbf{27.53}\scriptsize{$\pm$0.57} & 24.13\scriptsize{$\pm$0.46} \\
& Reacher           & 22.75\scriptsize{$\pm$0.59} & \textbf{24.25}\scriptsize{$\pm$0.47} & \textbf{24.97}\scriptsize{$\pm$0.54}  & 22.51\scriptsize{$\pm$0.82} & 23.35\scriptsize{$\pm$0.64} & 20.05\scriptsize{$\pm$1.37} & 22.36\scriptsize{$\pm$1.03} & 14.39\scriptsize{$\pm$1.08} & 22.54\scriptsize{$\pm$1.42} & 22.78\scriptsize{$\pm$1.02} \\
& SawyerReach       & 26.89\scriptsize{$\pm$0.07} & \textbf{26.92}\scriptsize{$\pm$0.09} & \textbf{27.35}\scriptsize{$\pm$0.06} & 22.82\scriptsize{$\pm$1.15}& 22.07\scriptsize{$\pm$1.46} & 19.20\scriptsize{$\pm$1.79} & 23.56\scriptsize{$\pm$0.33}& 13.39\scriptsize{$\pm$0.75} & 24.17\scriptsize{$\pm$0.01}  & 26.44\scriptsize{$\pm$0.31} \\
& SawyerDoor        &\textbf{26.18}\scriptsize{$\pm$2.19} & 25.85\scriptsize{$\pm$0.97} & 26.15\scriptsize{$\pm$2.08} & 23.62\scriptsize{$\pm$0.35} & 23.92\scriptsize{$\pm$1.10} & 20.12\scriptsize{$\pm$1.33} & \textbf{26.39}\scriptsize{$\pm$0.42} & 12.85\scriptsize{$\pm$0.77} & 24.81\scriptsize{$\pm$0.38} & 23.14\scriptsize{$\pm$0.56} \\
& FetchReach        & 30.29\scriptsize{$\pm$0.03} & \textbf{30.34}\scriptsize{$\pm$0.02} & \textbf{30.42}\scriptsize{$\pm$0.04} & 29.21\scriptsize{$\pm$0.26} & 28.17\scriptsize{$\pm$0.38} & 23.68\scriptsize{$\pm$1.07} & 29.08\scriptsize{$\pm$0.12} & 11.55\scriptsize{$\pm$0.68} & 28.71\scriptsize{$\pm$0.15} & 29.18\scriptsize{$\pm$0.25} \\
& FetchPush         & 19.91\scriptsize{$\pm$1.20} & 22.13\scriptsize{$\pm$1.41} & 21.58\scriptsize{$\pm$1.63} & \textbf{22.41}\scriptsize{$\pm$1.69} & \textbf{22.22}\scriptsize{$\pm$1.51} & 17.58\scriptsize{$\pm$1.47} & 19.86\scriptsize{$\pm$3.16}& 9.49\scriptsize{$\pm$2.85} & 17.82\scriptsize{$\pm$0.55} & 14.52\scriptsize{$\pm$0.95} \\
& FetchPick         & 19.66\scriptsize{$\pm$0.78} & \textbf{21.78}\scriptsize{$\pm$1.01} & \textbf{20.41}\scriptsize{$\pm$0.92} & 19.79\scriptsize{$\pm$1.12} & 18.32\scriptsize{$\pm$1.56} & 12.95\scriptsize{$\pm$1.90} & 17.04\scriptsize{$\pm$3.81} & 8.76\scriptsize{$\pm$0.64} & 14.45\scriptsize{$\pm$0.61} & 18.56\scriptsize{$\pm$0.82} \\
& FetchSlide        & 4.19\scriptsize{$\pm$1.89} & 4.98\scriptsize{$\pm$1.46} & \textbf{5.19}\scriptsize{$\pm$2.02}  & 3.34\scriptsize{$\pm$1.01} & \textbf{5.17}\scriptsize{$\pm$3.17} & 1.67\scriptsize{$\pm$1.41} & 3.31\scriptsize{$\pm$1.46} & 1.21\scriptsize{$\pm$0.59} & 0.98\scriptsize{$\pm$0.58} & 3.40\scriptsize{$\pm$0.80} \\
& HandReach         & 22.11\scriptsize{$\pm$0.55} & \textbf{23.44}\scriptsize{$\pm$0.62} & \textbf{24.93}\scriptsize{$\pm$0.49} & 15.39\scriptsize{$\pm$6.37} & 18.05\scriptsize{$\pm$5.12} & 0.15\scriptsize{$\pm$0.11} & 0.00\scriptsize{$\pm$0.00} & 0.00\scriptsize{$\pm$0.00} & 0.00\scriptsize{$\pm$0.00} & 15.44\scriptsize{$\pm$0.24} \\
\hline \rule{0pt}{2.5ex}%
& Average Rank & \multicolumn{1}{c}{3.5} & \multicolumn{1}{c}{\textbf{2.4}} & \multicolumn{1}{c|}{\textbf{1.7}} & \multicolumn{1}{c}{5.4} & \multicolumn{1}{c}{5.5} & \multicolumn{1}{c}{8.6} & \multicolumn{1}{c|}{6.2} & \multicolumn{1}{c}{9.7} & \multicolumn{1}{c}{6.2} & \multicolumn{1}{c|}{5.4} \\
\hline \rule{0pt}{3ex}%
\multirow{10}{*}{\rotatebox[origin=c]{90}{\textbf{Random}}}
& PointReach        & 29.26\scriptsize{$\pm$0.04} & \textbf{29.36}\scriptsize{$\pm$0.08} & \textbf{29.31}\scriptsize{$\pm$0.04} & 23.96\scriptsize{$\pm$0.93} & 25.76\scriptsize{$\pm$0.96} & 17.74\scriptsize{$\pm$1.84}  & 25.55\scriptsize{$\pm$0.57} & 11.12\scriptsize{$\pm$0.72} & 22.65\scriptsize{$\pm$1.57} & 26.12\scriptsize{$\pm$1.04} \\
& PointRooms        & 24.80\scriptsize{$\pm$0.36} & \textbf{25.17}\scriptsize{$\pm$0.19} & \textbf{25.16}\scriptsize{$\pm$0.59} & 18.09\scriptsize{$\pm$4.13} & 19.41\scriptsize{$\pm$1.01} & 14.69\scriptsize{$\pm$2.51}  & 19.10\scriptsize{$\pm$1.39} & 9.76\scriptsize{$\pm$2.99} & 20.88\scriptsize{$\pm$0.96} & 22.80\scriptsize{$\pm$1.12} \\
& Reacher           &21.09\scriptsize{$\pm$0.65} & \textbf{24.49}\scriptsize{$\pm$0.48} & 22.24\scriptsize{$\pm$0.54} & \textbf{25.20}\scriptsize{$\pm$0.48} & 22.98\scriptsize{$\pm$0.91} & 10.62\scriptsize{$\pm$2.30} & 23.70\scriptsize{$\pm$0.62} & 4.74\scriptsize{$\pm$0.36} & 6.06\scriptsize{$\pm$0.84} & 18.16\scriptsize{$\pm$1.08} \\
& SawyerReach       & 26.70\scriptsize{$\pm$0.14} & \textbf{26.78}\scriptsize{$\pm$0.12} & \textbf{27.07}\scriptsize{$\pm$0.07} & 19.48\scriptsize{$\pm$1.39} & 21.32\scriptsize{$\pm$1.40} & 8.78\scriptsize{$\pm$2.59}  & 25.29\scriptsize{$\pm$0.35} & 3.46\scriptsize{$\pm$0.86} & 2.84\scriptsize{$\pm$0.05} & 21.16\scriptsize{$\pm$0.95} \\
& SawyerDoor        & 19.05\scriptsize{$\pm$0.66} & 20.37\scriptsize{$\pm$1.18} & \textbf{21.69}\scriptsize{$\pm$2.36} & \textbf{20.69}\scriptsize{$\pm$2.14} & 19.58\scriptsize{$\pm$3.55} & 12.47\scriptsize{$\pm$3.08} & 18.82\scriptsize{$\pm$1.67} & 7.92\scriptsize{$\pm$0.86} & 14.77\scriptsize{$\pm$0.51} & 16.56\scriptsize{$\pm$0.92} \\
& FetchReach        &\textbf{30.42}\scriptsize{$\pm$0.04} & \textbf{30.44}\scriptsize{$\pm$0.02} & \textbf{30.42}\scriptsize{$\pm$0.04} & 28.34\scriptsize{$\pm$0.98} & 27.94\scriptsize{$\pm$0.30} & 18.96\scriptsize{$\pm$1.77} & 27.11\scriptsize{$\pm$0.22} & 1.71\scriptsize{$\pm$0.77} & 1.21\scriptsize{$\pm$0.46} & 23.02\scriptsize{$\pm$1.64} \\
& FetchPush         & 5.21\scriptsize{$\pm$0.43} & 6.83\scriptsize{$\pm$0.32} & \textbf{7.22}\scriptsize{$\pm$0.35} & \textbf{6.99}\scriptsize{$\pm$1.27} & 5.35\scriptsize{$\pm$3.36} & 4.22\scriptsize{$\pm$2.19} & 4.53\scriptsize{$\pm$1.94} & 4.49\scriptsize{$\pm$1.34} & 5.35\scriptsize{$\pm$0.23} & 5.10\scriptsize{$\pm$0.62}\\
& FetchPick         & 3.75\scriptsize{$\pm$0.18} & \textbf{4.22}\scriptsize{$\pm$0.16} & \textbf{4.36}\scriptsize{$\pm$0.19} & 3.81\scriptsize{$\pm$3.71} & 1.87\scriptsize{$\pm$1.59} & 0.81\scriptsize{$\pm$0.82} & 3.08\scriptsize{$\pm$1.35} & 2.16\scriptsize{$\pm$0.75} & 2.17\scriptsize{$\pm$0.18} & 3.21\scriptsize{$\pm$0.32} \\
& FetchSlide        & 2.67\scriptsize{$\pm$0.35} & \textbf{2.98}\scriptsize{$\pm$0.21} & \textbf{3.15}\scriptsize{$\pm$0.14} & 1.32\scriptsize{$\pm$1.22} & 1.04\scriptsize{$\pm$0.98} & 0.24\scriptsize{$\pm$0.27} & 1.12\scriptsize{$\pm$0.39} & 1.31\scriptsize{$\pm$0.52} & 0.00\scriptsize{$\pm$0.00} & 0.54\scriptsize{$\pm$0.12} \\
& HandReach         & 14.89\scriptsize{$\pm$2.54} & \textbf{18.24}\scriptsize{$\pm$2.18} & \textbf{20.06}\scriptsize{$\pm$3.06} & 0.08\scriptsize{$\pm$0.07} & 2.54\scriptsize{$\pm$1.42} & 1.41\scriptsize{$\pm$0.51} & 0.00\scriptsize{$\pm$0.00} & 0.00\scriptsize{$\pm$0.00} & 0.00\scriptsize{$\pm$0.00} & 8.36\scriptsize{$\pm$0.18} \\
\hline \rule{0pt}{2.5ex}%
& Average Rank & \multicolumn{1}{c}{3.8} & \multicolumn{1}{c}{\textbf{1.9}} & \multicolumn{1}{c|}{\textbf{1.7}} & \multicolumn{1}{c}{4.5} & \multicolumn{1}{c}{5.5} & \multicolumn{1}{c}{8.6} & \multicolumn{1}{c|}{6.0} & \multicolumn{1}{c}{8.8} & \multicolumn{1}{c}{7.9} & \multicolumn{1}{c|}{5.9} \\
\hline
\end{tabular}
\end{adjustbox}
\end{center}
\label{table:returns}%
\end{minipage}
\hfill
\begin{minipage}{0.272\textwidth}
    \includegraphics[width=\textwidth]{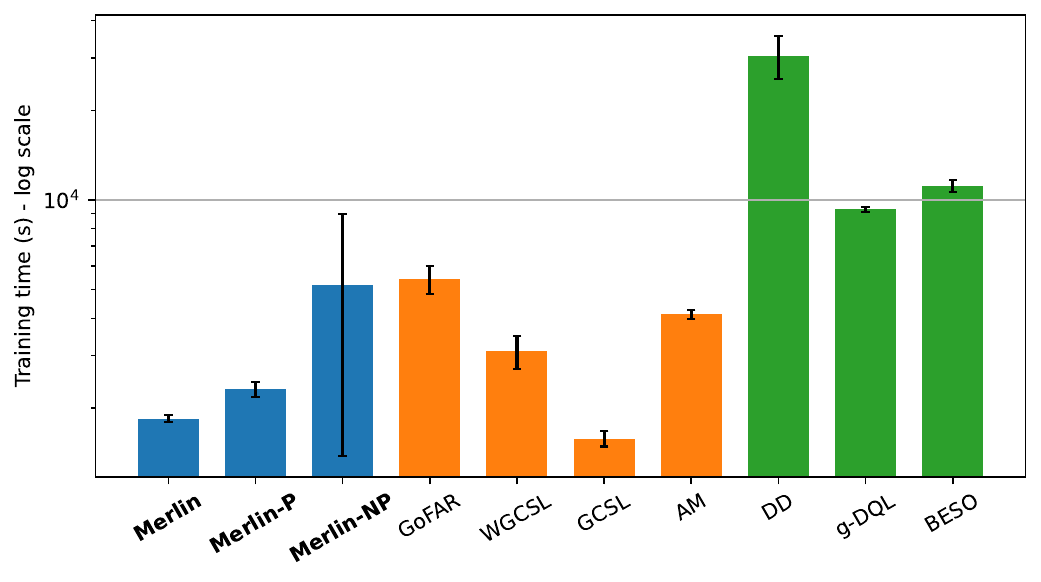}
    \includegraphics[width=\textwidth]{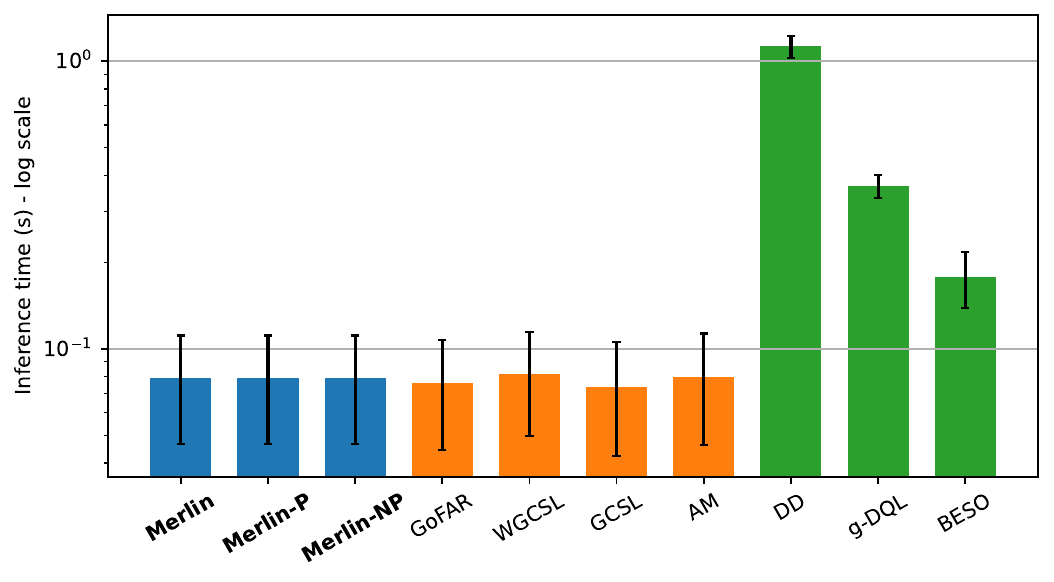}
    \caption{\small{Mean (a) training and (b) inference times for state-space input.}}
    \label{fig:times}
\end{minipage}
\vspace{-1em}
\end{figure*}

\paragraph{Tasks.} We evaluate Merlin on several goal-conditioned control tasks using the benchmark introduced in \citet{yang2021rethinking}. The benchmark consists of 10 different tasks of varying difficulty, with the maximum trajectory length fixed to be $T=50$ for all tasks. The tasks include the relatively easier PointReach, PointRooms, Reacher, SawyerReach, SawyerDoor, and FetchReach tasks with 2000 trajectories each ($1\times 10^5$ transitions); and the harder tasks include FetchPush, FetchPick, FetchSlide, and HandReach with 40000 trajectories ($2\times 10^6$ transitions).  The benchmark consists of two settings `expert' and `random'. The `expert' dataset consists of trajectories collected by a policy trained using online DDPG+HER with added Gaussian noise ($\sigma=0.2$) to increase diversity, while the `random' dataset consists of trajectories collected by sampling random actions. The dataset includes the desired goal for each trajectory in addition to the state-action pairs. The reward for each task is sparse and binary, $+1$ for reaching the goal, and $0$ everywhere else. Further details on these tasks are provided in \Cref{app:envs}.

\paragraph{Algorithms.}  
We compare with state-of-the-art offline GCRL methods, as well as diffusion-based RL methods. The GCRL methods are: (1) GCSL \citep{ghosh2020learning} which uses behavior cloning with hindsight relabeling, (2) WGCSL \citep{yang2021rethinking} that improves upon GCSL by incorporating advantage function weighting, (3) ActionableModel (AM) \citep{chebotar2021actionable} which uses an actor-critic method with conservative Q-learning and goal chaining, and (4) GoFAR \citep{ma2022offline} which uses advantage-weighted regression with $f$-divergence regularization based on state-occupancy matching. The diffusion-based methods are: (1) Decision Diffuser (DD) \citep{ajay2022conditional} which generates full trajectories from random Gaussian noise using classifier-free guidance, (2) Diffusion-QL (DQL) \citep{wang2022diffusion} that represents the policy as a diffusion model guided by a learned value function, modified by additionally conditioning the policy on goals (g-DQL), and (3) BESO \citep{reuss2023goal} that uses a goal-conditioned score-based diffusion model as its policy. \Cref{app:baselines} provides implementation details of the baselines.

We implement three variations of Merlin, all of which use behavior cloning and hindsight relabeling,
\begin{itemize}[align=right,itemindent=0em,labelsep=0.5em,labelwidth=1em,leftmargin=2em,nosep]
\item \textbf{Merlin} uses the offline data loaded to a replay buffer, and samples trajectories for reverse play. 
\item \textbf{Merlin-P} uses a learned parametric reverse dynamics model and reverse policy as described in \Cref{sec:model} to generate additional diffusion trajectories starting from goal states, in addition to the offline data.
\item \textbf{Merlin-NP} uses the non-parametric trajectory stitching method introduced in \Cref{sec:traj} to generate diverse diffusion trajectories, in addition to the offline data. Note that while the nearest-neighbors are computed based on the contrastive representations, the diffusion policy is trained on the original states.
\end{itemize}

We train each method for $500k$ policy updates using a batch size of $512$, and the results are averaged over $10$ seeds. Full implementation details for the three variants of Merlin are provided in \Cref{app:implementation}. We tune two hyperparameters - the hindsight relabeling ratio and the time horizon on each individual task; see \Cref{sec:ablation} for an ablation study. For the baselines, we use the best-reported hyperparameter values.

\begin{figure*}[t!]
\begin{minipage}{0.7\textwidth}
\captionof{table}{\small{Discounted returns for pixel-space input, averaged over 10 seeds.}}%
\vspace{-1em}
\begin{center}
\begin{adjustbox}{width=\textwidth}
\begin{tabular}{|ll|rrr|rrrr|rrr|}
\hline \\[-2.35ex]
\multicolumn{2}{c|}{\multirow{2}{*}{\textbf{Task Name}}} & \multicolumn{3}{c|}{\textbf{Ours}} & \multicolumn{4}{c|}{\textbf{Offline GCRL}} & \multicolumn{3}{c|}{\textbf{Diffusion-based}}\\
\rule{0pt}{2ex}
& & \multicolumn{1}{c}{\textbf{Merlin}} & \multicolumn{1}{c}{\textbf{Merlin-P}} & \multicolumn{1}{c|}{\textbf{Merlin-NP}} & \multicolumn{1}{c}{\textbf{GoFAR}} & \multicolumn{1}{c}{\textbf{WGCSL}} & \multicolumn{1}{c}{\textbf{GCSL}} & \multicolumn{1}{c|}{\textbf{AM}} & \multicolumn{1}{c}{\textbf{DD}} & \multicolumn{1}{c}{\textbf{g-DQL}} & \multicolumn{1}{c|}{\textbf{BESO}} \\
\hline \rule{0pt}{3ex}
\multirow{4}{*}{\rotatebox[origin=c]{90}{\textbf{Expert}}}
& PointReach        & 27.69\scriptsize{$\pm$0.06} & \textbf{28.54}\scriptsize{$\pm$0.08} & \textbf{28.95}\scriptsize{$\pm$0.05} & 25.14\scriptsize{$\pm$0.52} & 24.25\scriptsize{$\pm$0.60} & 21.06\scriptsize{$\pm$1.06} & 25.16\scriptsize{$\pm$1.22} & 8.20\scriptsize{$\pm$0.75} & 26.48\scriptsize{$\pm$0.76} & 27.92\scriptsize{$\pm$0.55} \\
& PointRooms        & 23.76\scriptsize{$\pm$0.19} & \textbf{25.16}\scriptsize{$\pm$0.26}  & \textbf{25.28}\scriptsize{$\pm$0.22} & 20.06\scriptsize{$\pm$0.34} & 19.72\scriptsize{$\pm$0.86} & 18.15\scriptsize{$\pm$1.59} & 22.47\scriptsize{$\pm$1.25}  & 5.54\scriptsize{$\pm$1.88} & 26.28\scriptsize{$\pm$0.64} & 23.80\scriptsize{$\pm$0.62} \\
& SawyerReach       & 26.87\scriptsize{$\pm$0.04} & \textbf{26.98}\scriptsize{$\pm$0.08}  & \textbf{27.15}\scriptsize{$\pm$0.06} & 22.16\scriptsize{$\pm$0.84} & 21.59\scriptsize{$\pm$1.02} & 19.04\scriptsize{$\pm$1.14} & 23.10\scriptsize{$\pm$1.12} & 6.89\scriptsize{$\pm$0.88} & 23.92\scriptsize{$\pm$0.15} & 25.96\scriptsize{$\pm$0.44} \\
& SawyerDoor        & 25.42\scriptsize{$\pm$0.08} & 25.15\scriptsize{$\pm$0.18}  & \textbf{26.08}\scriptsize{$\pm$0.08} & 23.17\scriptsize{$\pm$0.32} & 23.24\scriptsize{$\pm$0.75} & 19.76\scriptsize{$\pm$1.36} & \textbf{25.89}\scriptsize{$\pm$0.48} & 6.06\scriptsize{$\pm$1.12} & 24.44\scriptsize{$\pm$0.85} & 22.78\scriptsize{$\pm$1.28} \\
\hline \rule{0pt}{2.5ex}%
& Average Rank & \multicolumn{1}{c}{3.75} & \multicolumn{1}{c}{\textbf{2.75}} & \multicolumn{1}{c|}{\textbf{1.25}} & \multicolumn{1}{c}{7.0} & \multicolumn{1}{c}{7.5} & \multicolumn{1}{c}{9.0} & \multicolumn{1}{c|}{5.0} & \multicolumn{1}{c}{10.0} & \multicolumn{1}{c}{4.0} & \multicolumn{1}{c|}{4.75} \\
\hline \rule{0pt}{3ex}%
\multirow{4}{*}{\rotatebox[origin=c]{90}{\textbf{Random}}}
& PointReach        & 27.52\scriptsize{$\pm$0.05} & \textbf{28.80}\scriptsize{$\pm$0.08} & \textbf{28.76}\scriptsize{$\pm$0.06} & 23.51\scriptsize{$\pm$0.68} & 25.10\scriptsize{$\pm$0.88} & 17.34\scriptsize{$\pm$1.20} & 24.89\scriptsize{$\pm$0.72} & 6.36\scriptsize{$\pm$1.04} & 22.15\scriptsize{$\pm$1.32} & 25.85\scriptsize{$\pm$0.98} \\
& PointRooms        & 22.40\scriptsize{$\pm$0.07} & \textbf{24.05}\scriptsize{$\pm$0.22}  & \textbf{24.02}\scriptsize{$\pm$0.09} & 17.82\scriptsize{$\pm$1.89} & 19.02\scriptsize{$\pm$1.20} & 14.12\scriptsize{$\pm$1.92} & 18.82\scriptsize{$\pm$1.72} & 4.67\scriptsize{$\pm$2.15} & 20.16\scriptsize{$\pm$0.98} & 22.24\scriptsize{$\pm$1.08} \\
& SawyerReach       & 26.14\scriptsize{$\pm$0.04} & \textbf{26.46}\scriptsize{$\pm$0.10}  & \textbf{26.78}\scriptsize{$\pm$0.05} & 19.22\scriptsize{$\pm$1.08} & 21.04\scriptsize{$\pm$1.18} & 8.64\scriptsize{$\pm$2.44} & 25.01\scriptsize{$\pm$0.42} &  2.02\scriptsize{$\pm$2.59} & 2.32\scriptsize{$\pm$1.01} & 20.89\scriptsize{$\pm$0.98} \\
& SawyerDoor        & 18.99\scriptsize{$\pm$0.08} & 20.10\scriptsize{$\pm$1.78}  & \textbf{21.12}\scriptsize{$\pm$0.09} & \textbf{20.43}\scriptsize{$\pm$1.89} & 19.38\scriptsize{$\pm$1.68} & 12.04\scriptsize{$\pm$2.81} & 17.72\scriptsize{$\pm$0.84} & 4.12\scriptsize{$\pm$1.32} & 14.18\scriptsize{$\pm$0.65} & 16.24\scriptsize{$\pm$1.14} \\
\hline \rule{0pt}{2.5ex}%
& Average Rank & \multicolumn{1}{c}{3.5} & \multicolumn{1}{c}{\textbf{1.75}} & \multicolumn{1}{c|}{\textbf{1.5}} & \multicolumn{1}{c}{6.0} & \multicolumn{1}{c}{5.0} & \multicolumn{1}{c}{8.75} & \multicolumn{1}{c|}{5.75} & \multicolumn{1}{c}{10.0} & \multicolumn{1}{c}{7.5} & \multicolumn{1}{c|}{5.25} \\
\hline
\end{tabular}
\end{adjustbox}
\end{center}
\label{table:returns_img}%
\includegraphics[width=\textwidth]{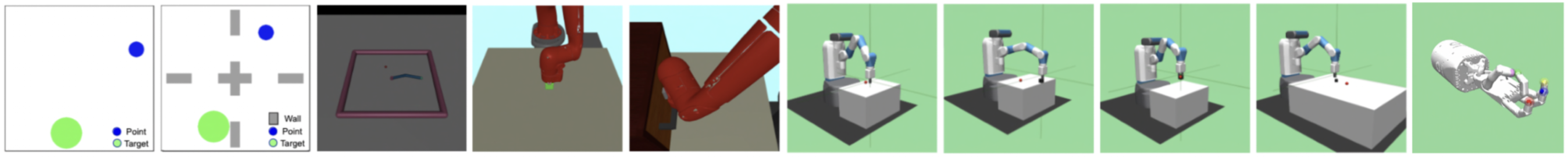}
\captionof{figure}{\small{Goal-conditioned tasks from left to right: PointReach, PointRooms, Reacher, SawyerReach, SawyerDoor, FetchReach, FetchPush, FetchPick, FetchSlide, and HandReach.}}
\end{minipage}
\hfill
\begin{minipage}{0.272\textwidth}
    \includegraphics[width=\textwidth]{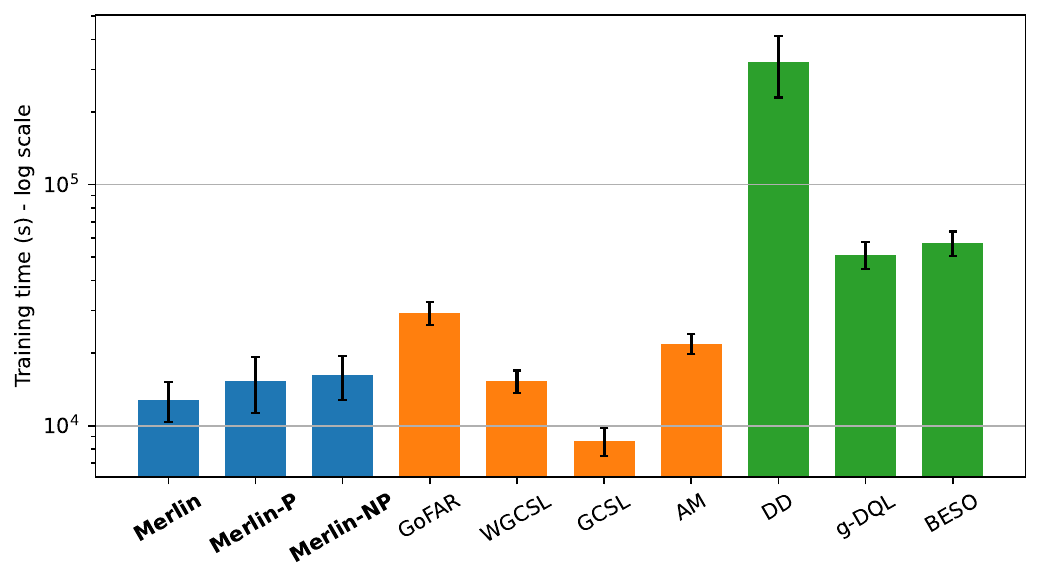}
    \includegraphics[width=\textwidth]{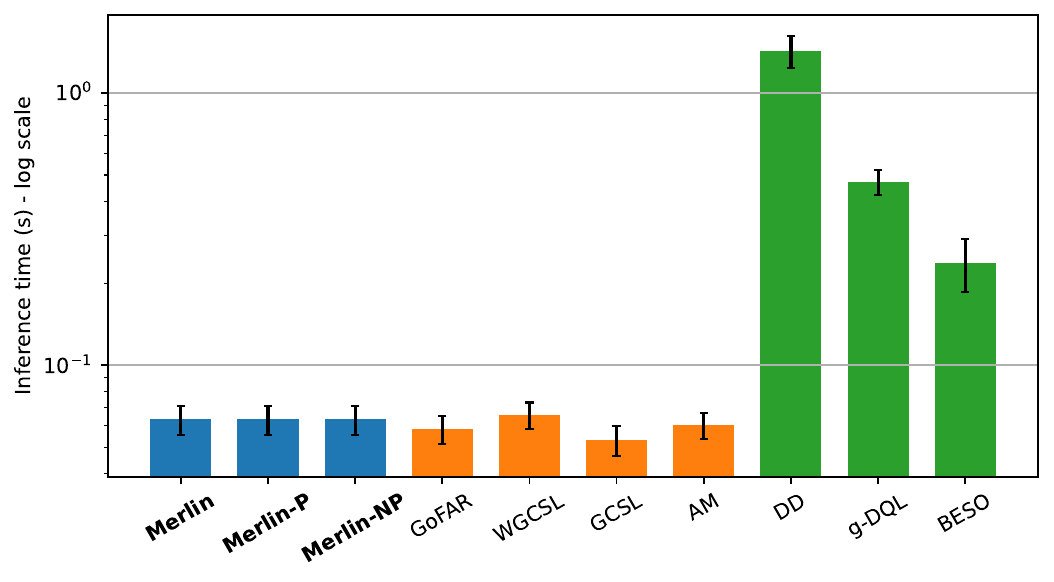}
    \caption{\small{Mean (a) training and (b) inference times for pixel-space input.}}
    \label{fig:times_img}
\end{minipage}
\vspace{-1em}
\end{figure*}

\subsection{Results}

\paragraph{State-space.} 
\Cref{table:returns} presents the discounted returns using the sparse binary task reward. The discounted return takes into account how fast the agent reaches the goal and whether it stays in the goal region thereafter. We also report the final success rate in \Cref{app:exp}. The results show that the basic implementation of Merlin outperforms the baselines on most tasks. Merlin-P and Merlin-NP improve the performance further, achieving the highest discounted returns on most tasks, and are overall the best-performing methods. Since Merlin does not perform multiple denoising steps for each environment step, training and inference are roughly an order of magnitude faster than the other diffusion-based methods (DD, g-DQL and BESO), which is shown in \Cref{fig:times}. A more detailed discussion on the training and inference times for all methods is provided in \Cref{app:compute}.

\paragraph{Pixel-space.}
We also perform experiments with pixel-space observations for the tasks that support image observations. The results show a similar trend as the state-space inputs and are provided in \Cref{table:returns_img}. In terms of computational efficiency, the improvement of Merlin over the diffusion-based methods is even more pronounced (\Cref{fig:times_img}), since the pixel-space observations are much higher dimension than the state observations.

\begin{figure*}[!b]
  \centering
  \begin{subfigure}[b]{0.4\linewidth}
      \centering
      \includegraphics[width=\linewidth]{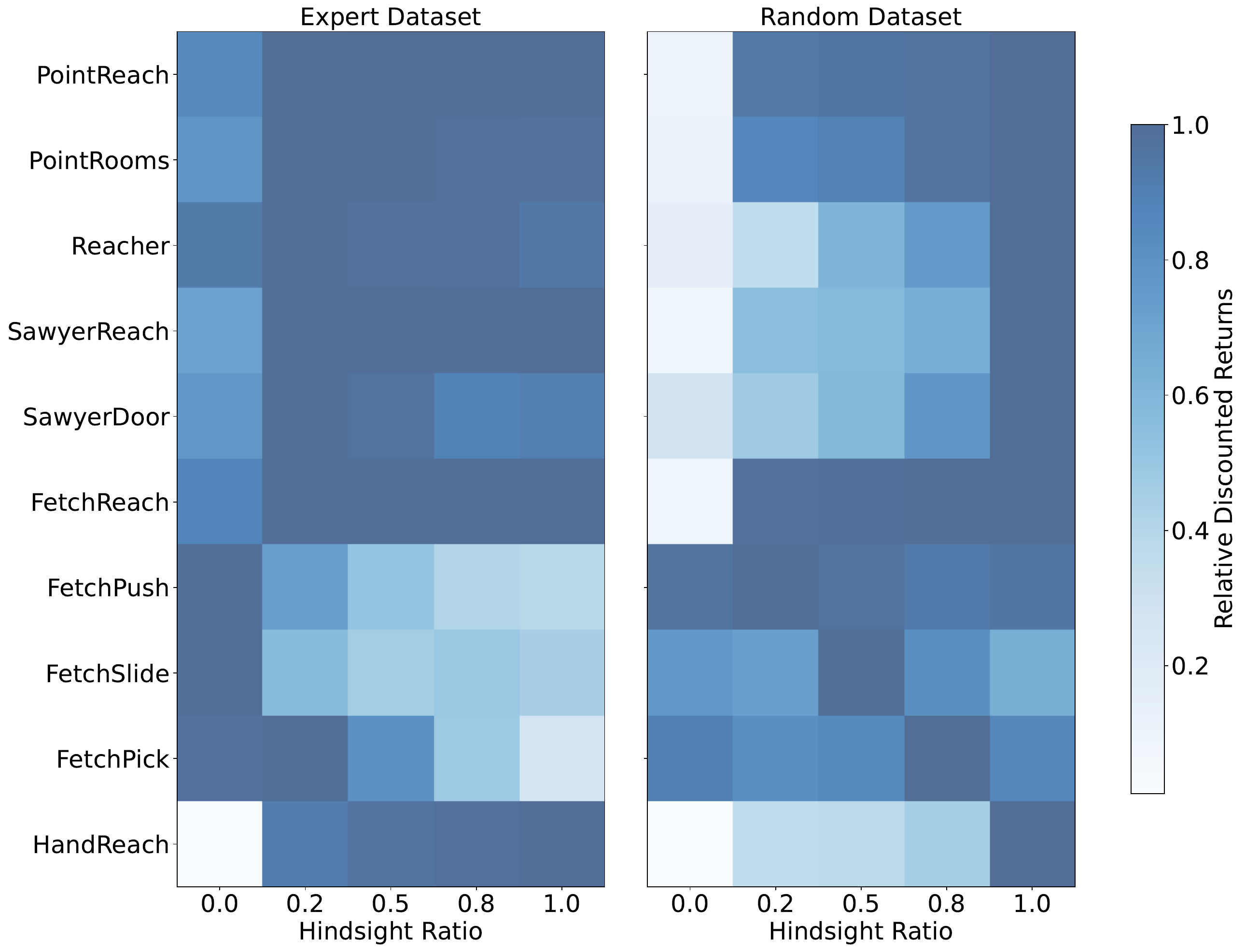}
      \caption{\texttt{}}
      \label{fig:abl_ratio}
  \end{subfigure}
  \hspace{1em}
  \begin{subfigure}[b]{0.465\linewidth}
      \centering
      \includegraphics[width=\linewidth]{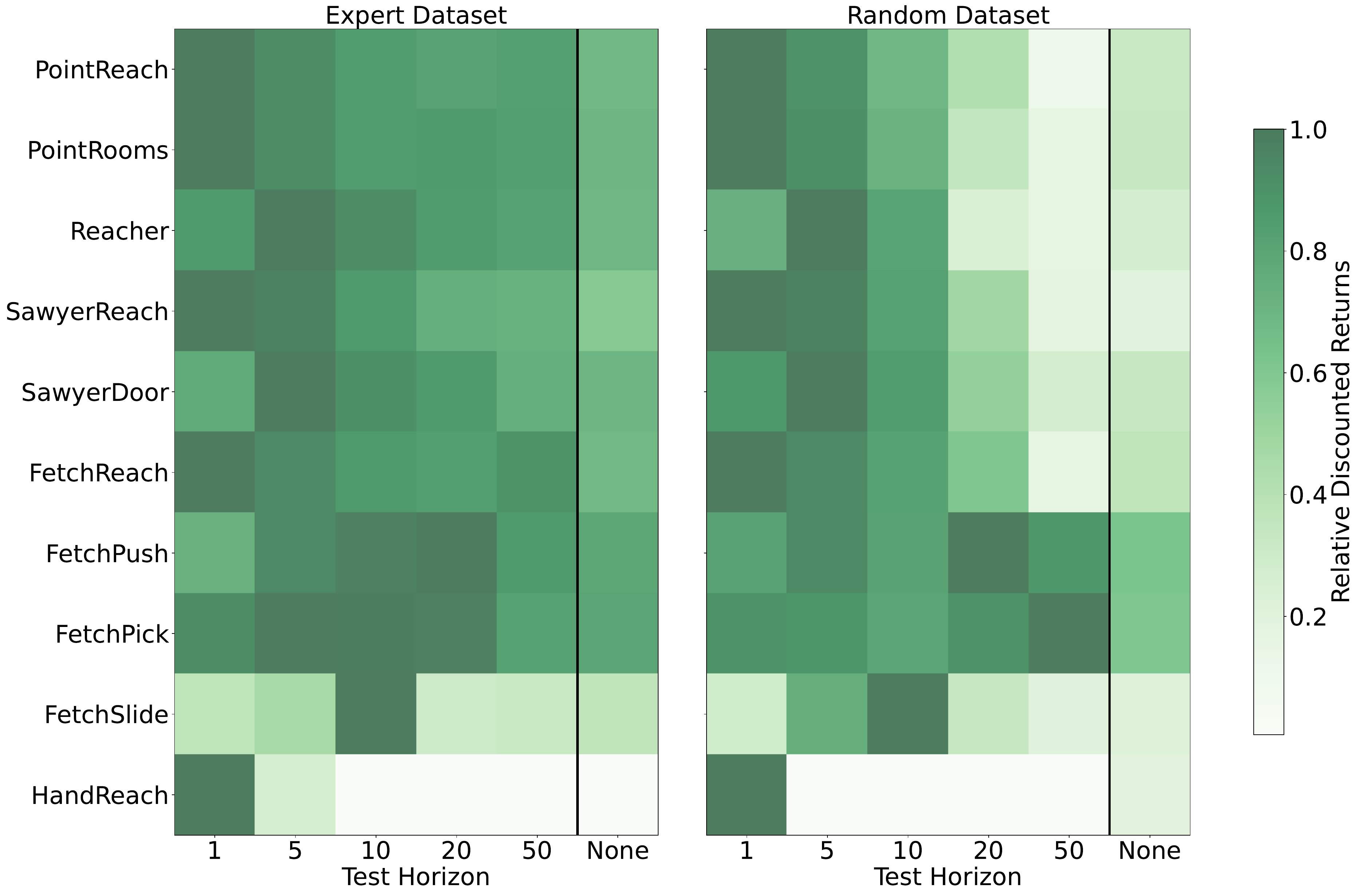}
      \caption{\texttt{}}
      \label{fig:abl_horizon}
  \end{subfigure}
  \caption{\small{Discounted returns for each dataset with different values of (a) hindsight ratio and (b) time horizon during evaluation. Values are normalized with respect to the maximum value in each row.}}
  \vspace{-1em}
  \label{fig:ablation}
\end{figure*}

\paragraph{Comparison to GCSL.}
Both Merlin and GCSL employ behavior cloning with hindsight relabeling, with two key differences: (1) Merlin learns the variance of the policy in addition to the mean, which provides additional flexibility during optimization, and (2) Merlin conditions the policy on the time horizon similar to the denoising function in diffusion models. GCSL also allows for this conditioning, however, it does not learn the variance, resulting in similar performance with and without conditioning on the time horizon.
The significance of this difference between proper conditioning on the horizon is comparable to the difference between denoising diffusion model and earlier score based methods for generative modelling.
\Cref{fig:diff_toy_horizon} illustrates the effect of time horizon on the learned variance, and \Cref{sec:ablation} further demonstrates its effect on the performance of Merlin. 
Beyond the vanilla setting of reverse play from the buffer, the forward view of GCSL and the backward view of Merlin, which is inspired by diffusion, can result in very different outcomes.
Consider the model-based approach: a forward dynamics model generates trajectories without guarantees on the distribution over the goal state. In contrast, in a reverse dynamics model, one has control over this distribution. We apply a modified version of the nearest-neighbor trajectory stitching operation to GCSL by constructing forward-looking trajectories and report the performance in \Cref{app:traj}. Here again, we see that GCSL's performance remains inferior to Merlin-NP. We also compare Merlin-P and Merlin-NP with their corresponding forward view variants in \Cref{app:forward_backward} demonstrating that the backward view is superior on most tasks.

\subsection{Ablation Studies}
\label{sec:ablation}

\paragraph{Hindsight Relabeling.} During training, we employ hindsight relabeling to replace the desired goals with a future state further along the trajectory. A hyperparameter, which we call the \emph{hindsight ratio}, specifies the fraction of each sampled batch of transitions that are subjected to this relabeling operation. As shown in \Cref{fig:abl_ratio}, this ratio can significantly affect performance depending on the dataset. In general, a low-to-moderate value for the expert datasets and a high value for the random datasets seem to result in good performance. This observation can be explained by the fact that a large number of the expert trajectories reach the desired goals hence the original state-goal pairs provide good quality data for training the policy. On the other hand, the random trajectories benefit more from hindsight relabeling since state-goal pairs in the original dataset are sub-optimal, and relabeling provides realistic state-goal pairs to the policy. For the baselines that use hindsight relabeling, we use the optimal hindsight ratio as reported in their works.

\paragraph{Time Horizon.} During training, the time horizon indicates the time difference between the current and desired goal states. During evaluation, the optimal value of the time horizon depends on the environment, as shown in \Cref{fig:abl_horizon}. The last column, labeled `None' shows the performance without conditioning on the time horizon, and for all tasks conditioning on the time horizon performs much better than without.  For the easier tasks, a time horizon of $h=1$ or $h=5$ seems to work best, whereas for the more complex tasks, a higher value seems optimal. This can be attributed to the fact that for the more difficult tasks, the policy is expected to require more time steps to successfully reach the goal. In particular, the HandReach task seems especially sensitive to the time horizon, as using $h=1$ performs significantly better than other values or without using time horizon conditioning. The optimal values for the hindsight ratio and the time horizon are provided in \Cref{app:merlin}.

\section{Related Works}
\label{sec:related}

\paragraph{Offline Goal-Conditioned Reinforcement Learning.} 
Offline GCRL methods generally aim to address the sparse nature of the rewards and the limited state-goal pairs present in the dataset. One widely used technique for goal-reaching problems is hindsight experience relabeling (HER) \citep{andrychowicz2017hindsight}, which replaces the desired goals with the achieved goals that appear later along the same trajectory. 
Closely related to our work is Goal-Conditioned Supervised Learning (GCSL) \citep{ghosh2020learning} which uses HER to train the policy on data collected by the agent itself. In the offline setting, this takes the form of behavior cloning combined with HER. Several other works employ goal-conditioned behavior cloning \citep{ding2019goal, lynch2020learning} to learn performant policies. \citet{yang2021rethinking, chebotar2021actionable} incorporate value learning methods and adapt them to the offline goal-conditioned setting. A different line of work in offline RL is based on distribution matching of the state-action visitation distribution of the learned policy and the expert policy \citep{ghasemipour2020divergence, ni2021f}. \citet{ma2022offline} apply this state-occupancy matching perspective to the offline goal-conditioned setting.

\paragraph{Diffusion-based Reinforcement Learning.} Recent works have leveraged diffusion models for offline RL by generating trajectory segments from random Gaussian noise. Diffuser \citep{janner2022planning} employs classifier-based guidance using a learned value function to guide the diffusion process to generate high-return trajectories. In contrast, Decision Diffuser (DD) \citep{ajay2022conditional} uses classifier-free guidance by learning a denoising function conditioned on returns, goals, or constraints. These methods operate similarly to model predictive control \citep{garcia1989model}, where only the first action of the generated trajectory is performed. A different line of work, including Diffusion-QL (DQL) \citep{wang2022diffusion}, represents the \emph{policy} as a diffusion model where actions are sampled by denoising random Gaussian noise, conditioned on the states and guided by a learned value function. BESO \citep{reuss2023goal} uses a goal-conditioned score-based diffusion model as its policy while decoupling the score function learning and inference sampling process.

\section{Conclusion}

We introduce Merlin, a goal-conditioned RL method that draws inspiration from generative diffusion models. 
Its main appeal is simplicity and potential for scalability, as proven by denoising diffusion models. 
Distinct from other works that use diffusion for RL, we construct trajectories that ``diffuse away'' from potential goals and train a policy to reverse them, analogous to the score function. We discuss several choices to construct the forward diffusion process and introduce a novel latent space trajectory stitching method that can be applied to most offline RL algorithms. We demonstrate that Merlin is performant and efficient on various offline goal-conditioned control tasks. 

\section*{Acknowledgements}
We thank the anonymous reviewers for their feedback on the earlier version of the paper.
This research is in part supported by Canada CIFAR AI Chair and NSERC Discovery.  
Mila and the Digital Research Alliance of Canada provided computational resources.

\section*{Impact Statement}
This paper contributes to the advancement of both reinforcement learning and the broader field of machine learning. In particular, we foresee it to have a positive impact on the scale of problems addressed by reinforcement learning. Although the scalability of such applications in real-world domains from robotics and recommendation systems, to autonomous driving may have significant societal impacts, we find that our proposed method is agnostic to the positivity of this impact, and similar to many other methodologies can be deployed in applications with positive or negative societal impacts. 

\bibliography{example_paper}
\bibliographystyle{icml2024}

\newpage
\appendix
\onecolumn

\section{Proof of Theorem \ref{thm}}
\label{sec:app_proof}

\paragraph{Setting.} Consider a dataset $\mathcal{D}(g)$ where each trajectory is of the form $\tau = \{s_1, a_1, \ldots, s_T\}$ generated by some unknown behavior policy $\pi_\beta$. The final states are such that $g=\phi(s_T)$. We view these trajectories in reverse - starting from the final state $s_T$, we apply some unknown transformation to state $s_{t+1}$ to obtain state $s_t$. The corresponding forward diffusion process is denoted by $q(s_{t} | s_{t+1})$.

\paragraph{Outline.} The basic steps involved in the proof are:
\begin{enumerate}[align=right,itemindent=0em,labelsep=0.5em,labelwidth=1em,leftmargin=2em,nosep]
\item Define the forward and reverse diffusion processes.
\item Obtain the distribution of final states achieved by the reverse diffusion process.
\item Define the log-likelihood of final states under the reverse diffusion process. Lower bound the log-likelihood using Jensen's inequality and simplify the resulting expression.
\item Obtain the optimal policy parameters by maximizing the lower bound for (a) deterministic and (b) stochastic MDPs.
\end{enumerate}

\paragraph{Proof.}
\begin{enumerate}[align=right,itemindent=0em,labelsep=0.5em,labelwidth=1em,leftmargin=2em,nosep]
\item Let $q(s_T|g)$ denote the target distribution of final states corresponding to the goal $g$. For brevity, we denote it simply as $q(s_T)$, since in this setting the goal $g$ is fixed. The forward diffusion trajectory, starting at $s_T$ and performing $T$ steps of diffusion is thus,
\begin{align*}
    q(s_1,\ldots,s_T) = q(s_T) \prod_{t=1}^{T-1} q(s_{t} | s_{t+1}),
\end{align*}
We train a policy denoted by $\pi_\theta(\cdot | s_t)$ to reverse this diffusion. The corresponding reverse diffusion process is given by,
\begin{align*}
    p_\theta(s_{t+1} | s_t) = \mcP(s_{t+1} | s_t, \pi_\theta(\cdot | s_t)),    
\end{align*}
The generative process corresponding to this reverse diffusion is,
\begin{align*}
    p_\theta(s_1,\ldots,s_T) = p(s_1) \prod_{t=1}^{T-1} p_\theta(s_{t+1} | s_{t}),
\end{align*}
where $p(s_1)$ is the distribution of initial states. \\

\item The distribution of final states achieved by the reverse diffusion process,
\begin{align*}
    p_\theta(s_T) &= \int ds_1\ldots ds_{T-1} \; p_\theta(s_1,\ldots,s_{T})\\
    &= \int ds_1\ldots ds_{T-1} \; q(s_1,\ldots,s_{T-1} | s_T) \frac{p_\theta(s_1,\ldots,s_{T})}{q(s_1,\ldots,s_{T-1} | s_T)}\\
    &= \int ds_1\ldots ds_{T-1} \; q(s_1,\ldots,s_{T-1} | s_T) p(s_1) \prod_{t=1}^{T-1} \frac{p_\theta(s_{t+1} | s_{t})}{q(s_t | s_{t+1})}
\end{align*}\\

\item During training, the objective is to maximize the log-likelihood of final states given by the reverse diffusion process, with final states sampled from the target state distribution $q(s_T|g)$,
\begin{align*}
    L(\theta) &= \mathbb{E}_{s_T \sim q(s_T)} \left[\log p_\theta(s_T)\right] = \int ds_T \; q(s_T) \log p_\theta(s_T)\\
    &= \int ds_T \; q(s_T) \log \left[\int ds_1\ldots ds_{T-1} \; q(s_1,\ldots,s_{T-1} | s_T) p(s_1) \prod_{t=1}^{T-1} \frac{p_\theta(s_{t+1} | s_{t})}{q(s_t | s_{t+1})}\right]\\
    &\geq \int ds_1\ldots ds_T \; q(s_1,\ldots,s_T) \log \left[ p(s_1) \prod_{t=1}^{T-1} \frac{p_\theta(s_{t+1} | s_{t})}{q(s_t | s_{t+1})}\right]
\end{align*}
where the lower bound is provided by Jensen's inequality.

We separate the term corresponding to the initial state $s_1$,
\begin{align*}
    L(\theta) &\geq \int ds_1\ldots ds_T \; q(s_1,\ldots,s_T) \sum_{t=1}^{T-1} \log \left[  \frac{p_\theta(s_{t+1} | s_{t})}{q(s_t | s_{t+1})}\right] + \int ds_1 \; q(s_1) \log p(s_1)\\
    &= \sum_{t=1}^{T-1} \int ds_t ds_{t+1} \; q(s_t,s_{t+1}) \log \left[  \frac{p_\theta(s_{t+1} | s_{t})}{q(s_t | s_{t+1})}\right] + \int ds_1 \; q(s_1) \log p(s_1)
\end{align*}
We apply Bayes' rule to rewrite in terms of posterior of the forward diffusion,
\begin{align*}
    L(\theta) &\geq \sum_{t=1}^{T-1} \int ds_t ds_{t+1} \; q(s_t,s_{t+1}) \log \left[  \frac{p_\theta(s_{t+1} | s_{t})}{q(s_{t+1} | s_{t})} \frac{q(s_{t+1})}{q(s_{t})}\right] + \int ds_1 \; q(s_1) \log p(s_1) \\
    &= \sum_{t=1}^{T-1} \int ds_t ds_{t+1} \; q(s_t,s_{t+1}) \log \left[  \frac{p_\theta(s_{t+1} | s_{t})}{q(s_{t+1} | s_{t})}\right]\\
    &\qquad \qquad \qquad \qquad + \sum_{t=1}^{T-1} \left[H_q(S_t) - H_q(S_{t+1})\right] + \int ds_1 \; q(s_1) \log p(s_1) \\
    &\qquad \qquad \qquad \qquad + H_q(S_1) - H_q(S_{T}) + \int ds_1 \; q(s_1) \log p(s_1)\\
    &= - \sum_{t=1}^{T-1} \int ds_t ds_{t+1} \; q(s_t)q(s_{t+1}|s_t) \log \left[  \frac{q(s_{t+1} | s_{t})}{p_\theta(s_{t+1} | s_{t})}\right] \\
    &\qquad \qquad \qquad \qquad + H_q(S_1) - H_q(S_{T}) + \int ds_1 \; q(s_1) \log p(s_1) \\
    &= - \sum_{t=1}^{T-1} \int ds_t \; q(s_t) D_{KL}\big(q(s_{t+1}|s_t) \| p_\theta(s_{t+1}|s_t)\big)\\
    &\qquad \qquad \qquad \qquad + H_q(S_1) - H_q(S_{T}) + \int ds_1 \; q(s_1) \log p(s_1)
\end{align*}
\item We maximize the log-likelihood with respect to the policy parameters $\theta$, which is equivalent to minimizing the first term,
\begin{align*}
    \theta^* &= \argmax_\theta L(\theta) \equiv \argmin_\theta \sum_{t=1}^{T-1} \int ds_t \; q(s_t) D_{KL}\big(q(s_{t+1}|s_t) \| p_\theta(s_{t+1}|s_t)\big)
\end{align*}
The posterior of the forward diffusion is simply the state transition using the behavior policy $\pi_\beta$,
\begin{align*}
    \theta^* = \argmin_\theta \sum_{t=1}^{T-1} \int ds_t \; q(s_t) D_{KL}\big(\mcP(s_{t+1}|s_t, \pi_\beta(\cdot)) \| \mcP(s_{t+1}|s_t, \pi_\theta(\cdot | s_t))\big)
\end{align*}
\begin{enumerate}
\item For deterministic state transitions, the next state $s_{t+1}$ is given by the dynamics function $f$ of the MDP, $s_{t+1} = f(s_t, a_t)$. For a given state $s_t$, this dynamics function represents a fixed parameter transformation of the policy function. We exploit the property that KL divergence is invariant under parameter transformations. Thus for a deterministic MDP,
\begin{align*}
    \theta^* &= \argmin_\theta \sum_{t=1}^{T-1} \int ds_t \; q(s_t) D_{KL}\big(f(s_t, \pi_\beta(\cdot)) \| f(s_t, \pi_\theta(\cdot | s_t))\big)\\
    &= \argmin_\theta \sum_{t=1}^{T-1} \int ds_t \; q(s_t) D_{KL}\big(\pi_\beta(\cdot) \| \pi_\theta(\cdot | s_t)\big)
\end{align*}

\item For stochastic state transitions, the next state $s_{t+1}$ is given by a noisy dynamics function $s_{t+1} = f(s_t, a_t, \epsilon)$, where $\epsilon \sim \xi(\epsilon)$ denotes random noise to account for the stochasticity. For a given state $s_t$, this dynamics function represents a fixed parameter transformation of the joint distribution of the policy and the noise distribution. Abusing notation, we denote this joint distribution as $p(\pi, \xi)$. Since KL divergence is invariant under parameter transformations, for a stochastic MDP,
\begin{align*}
    \theta^* &= \argmin_\theta \sum_{t=1}^{T-1} \int ds_t \; q(s_t) D_{KL}\big(f(s_t, \pi_\beta(\cdot), \xi) \| f(s_t, \pi_\theta(\cdot | s_t), \xi)\big)\\
    &= \argmin_\theta \sum_{t=1}^{T-1} \int ds_t \; q(s_t) D_{KL}\big(p(\pi_\beta(\cdot), \xi) \| p(\pi_\theta(\cdot | s_t), \xi)\big)
\end{align*}
Since the policy and the noise distribution $\xi$ are independent, the KL divergence decomposes,
\begin{align*}
    \theta^* &=\argmin_\theta \sum_{t=1}^{T-1} \int ds_t \; q(s_t) D_{KL}\big(\pi_\beta(\cdot) \| \pi_\theta(\cdot | s_t)\big) + \sum_{t=1}^{T-1} D_{KL}\big(\xi \| \xi\big)\\
    &= \argmin_\theta \sum_{t=1}^{T-1} \int ds_t \; q(s_t) D_{KL}\big(\pi_\beta(\cdot) \| \pi_\theta(\cdot | s_t)\big)
\end{align*}

\end{enumerate}
Minimizing the KL divergence between the policies is equivalent to maximizing the log-likelihood of the behavior policy action under the parameterized policy. Therefore, given state-action-goal tuples $(s,a) \sim \mathcal{D}(g)$,
\begin{align*}
    \theta^* &= \argmax_\theta \mathbb{E}_{(s,a) \sim \mathcal{D}(g)} \left[ \log \pi_\theta(a | s)\right]
\end{align*}
Therefore, behavior cloning is equivalent to maximizing a lower bound on the log-likelihood of the target final states achieved by the reverse diffusion process.
\end{enumerate}

\section{Additional Illustrative Experiments}
\label{app:add_toy}

\subsection{Four Rooms Navigation}
\label{app:fourrooms}

The illustrative example presented in \Cref{sec:toy} considered a simple navigation problem. In this section, we extend the analysis to the four rooms variant, which adds walls that the agent must navigate around in order to reach the goal. We seek to understand whether Merlin can learn policies that produce more complex behavior compared to simply heading straight toward the goal.

\begin{figure}[h]
  \centering
  \begin{subfigure}[b]{0.32\textwidth}
      \centering
      \includegraphics[width=\textwidth]{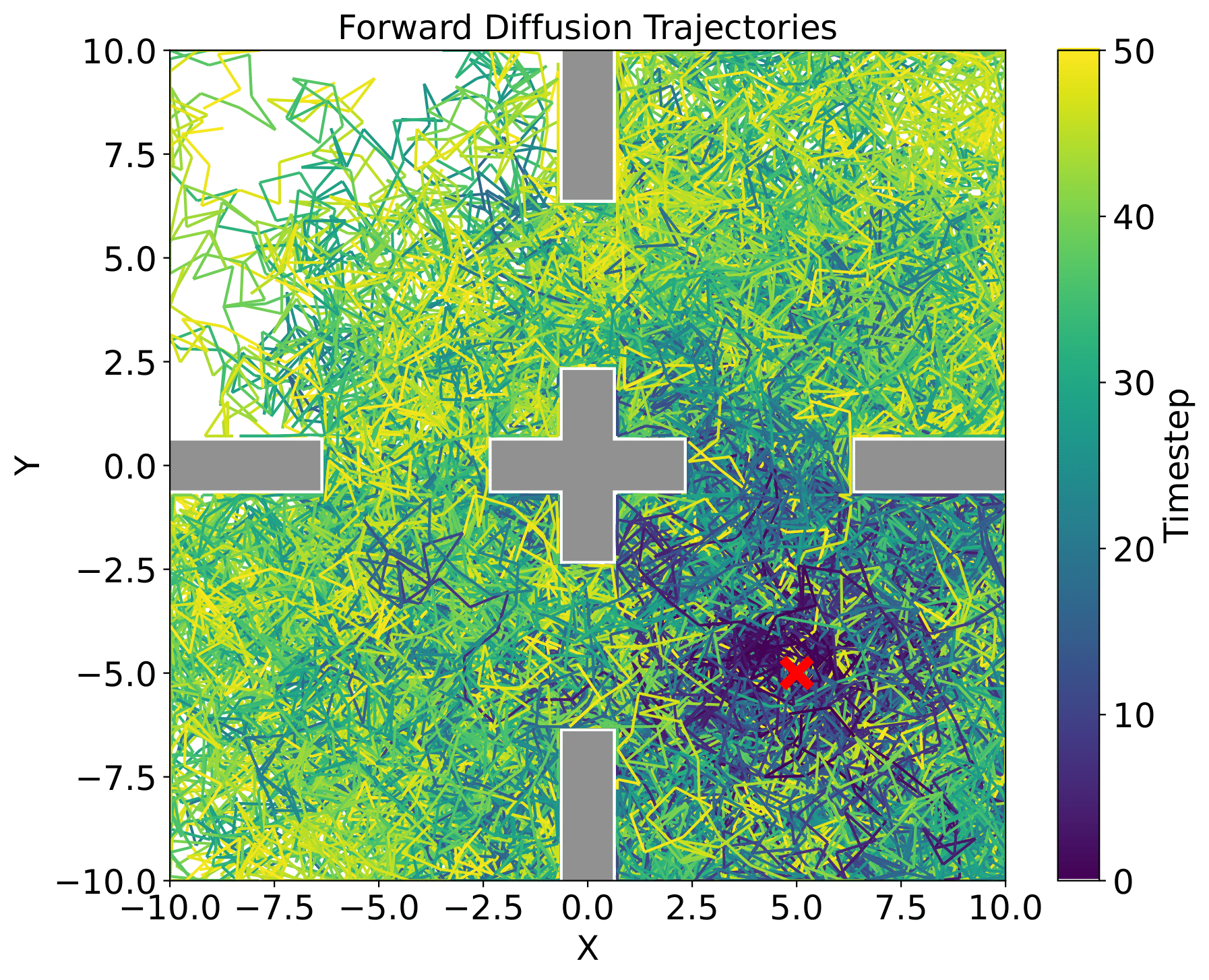}
      \caption{\small{}}
      \label{fig:diff_fourrooms_traj}
  \end{subfigure}
  \begin{subfigure}[b]{0.27\textwidth}
      \centering
      \includegraphics[width=\textwidth]{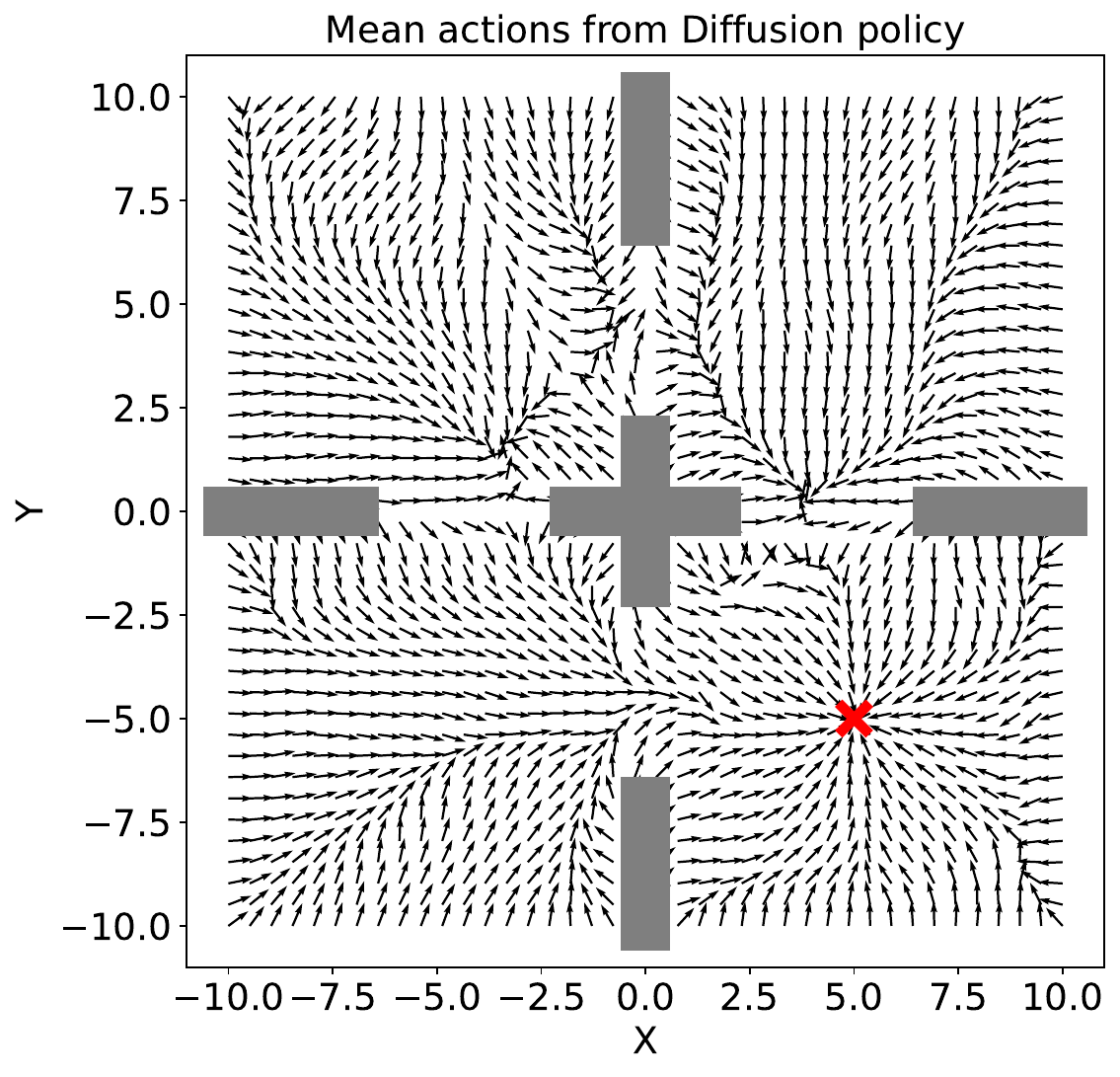}
      \caption{\small{}}
      \label{fig:diff_fourrooms_main}
  \end{subfigure}
  \begin{subfigure}[b]{0.27\textwidth}
      \centering
      \includegraphics[width=\textwidth]{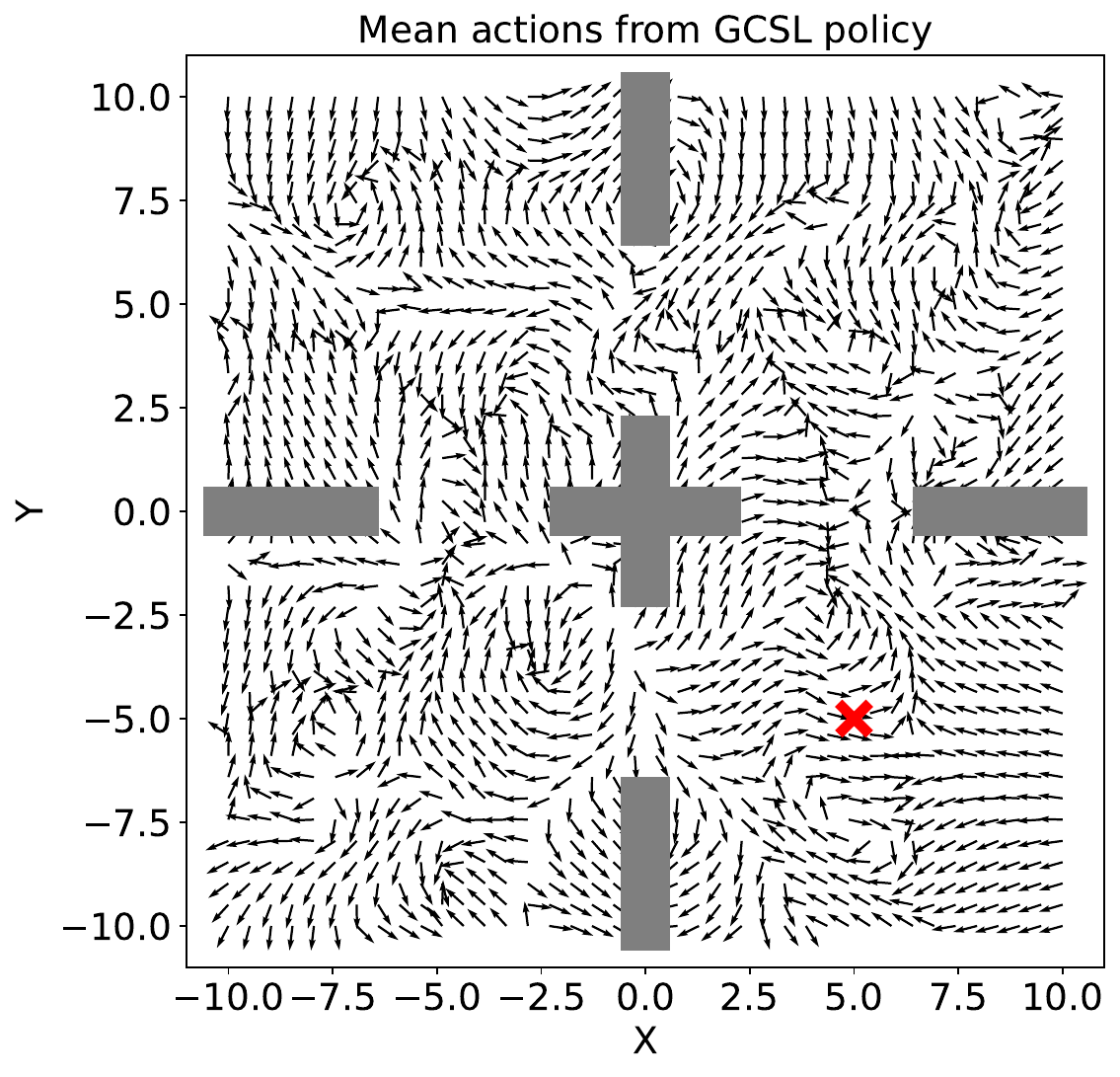}
      \caption{\small{}}
      \label{fig:diff_fourrooms_gcsl}
  \end{subfigure}
  \vspace{-0.5em}
  \caption{\small{(a) Visualization of trajectories starting from the goal \textbf{\textcolor{red}{X}} generated during the forward process, (b) Predicted actions from policy trained via diffusion, (c) Predicted actions from policy trained using GCSL.}}
  \vspace{-0.25em}
  \label{fig:fourrooms}
\end{figure}

The goal state during training is fixed to $g=(5,-5)$ in one of the quadrants. 
\Cref{fig:diff_fourrooms_traj} visualizes the trajectories during forward diffusion by taking random actions starting from the goal. \Cref{fig:diff_fourrooms_main} and \Cref{fig:diff_fourrooms_gcsl} visualize the policy learned by Merlin and GCSL, respectively. Both methods were trained for $100k$ policy updates. Merlin effectively learns to navigate around the walls, while still managing to reach the goal. In contrast, GCSL often navigates directly into the walls and in some areas wanders away from the goal.

\begin{figure}[h]
  \centering
  \begin{subfigure}[b]{0.245\textwidth}
      \centering
      \includegraphics[width=\linewidth]{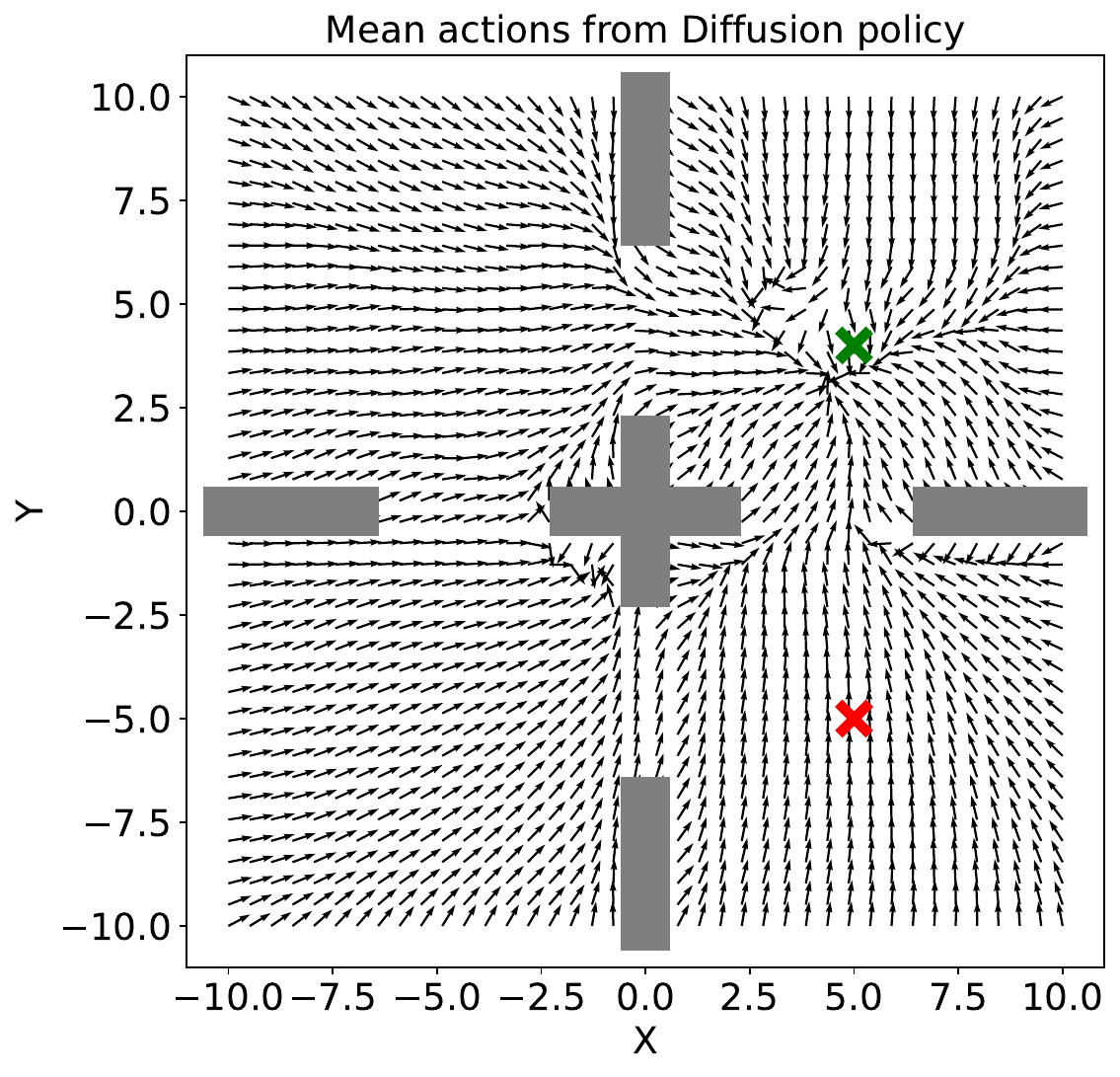}
  \end{subfigure}
  \begin{subfigure}[b]{0.245\textwidth}
      \centering
      \includegraphics[width=\textwidth]{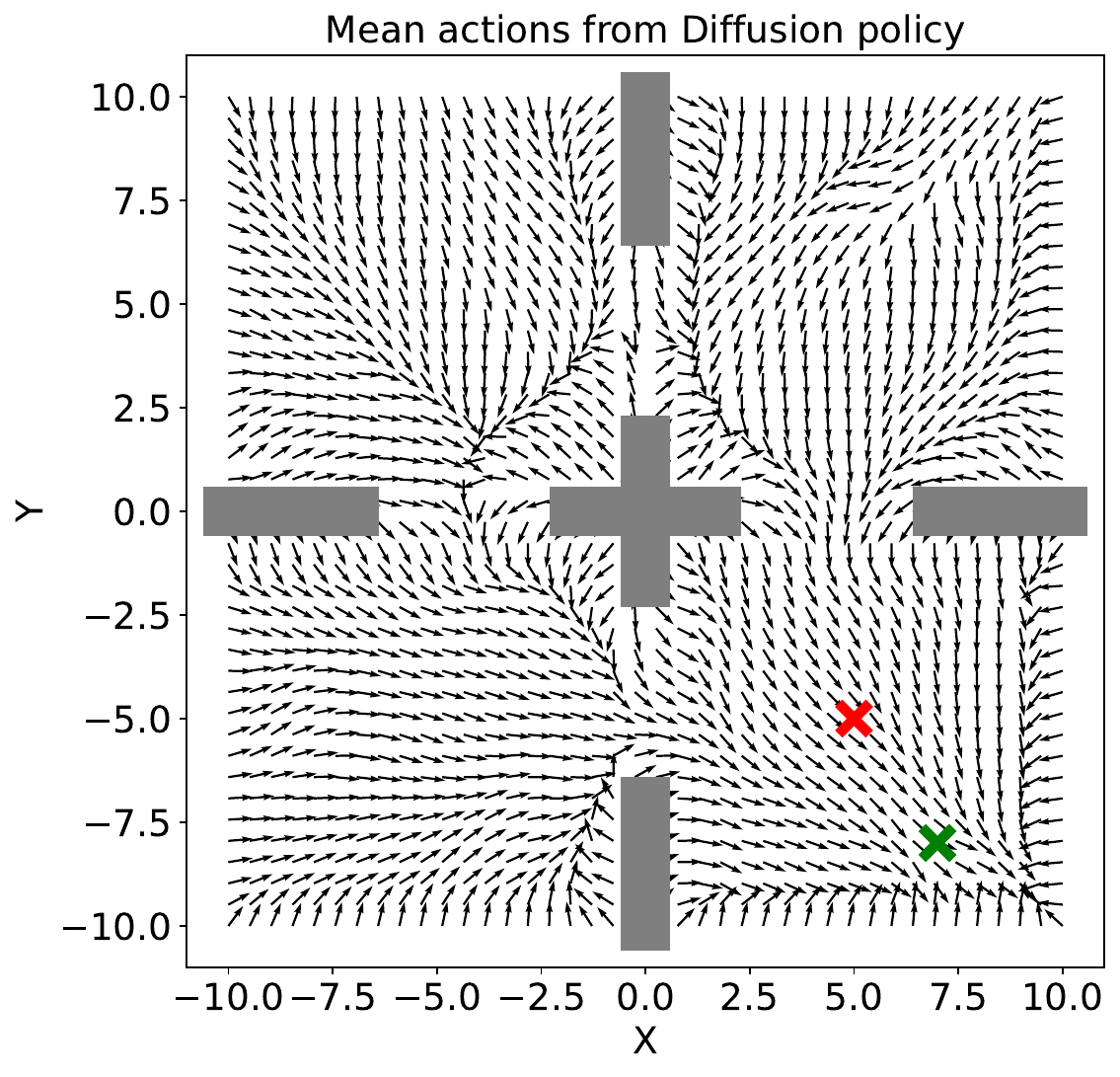}
  \end{subfigure}
  \begin{subfigure}[b]{0.245\textwidth}
      \centering
      \includegraphics[width=\textwidth]{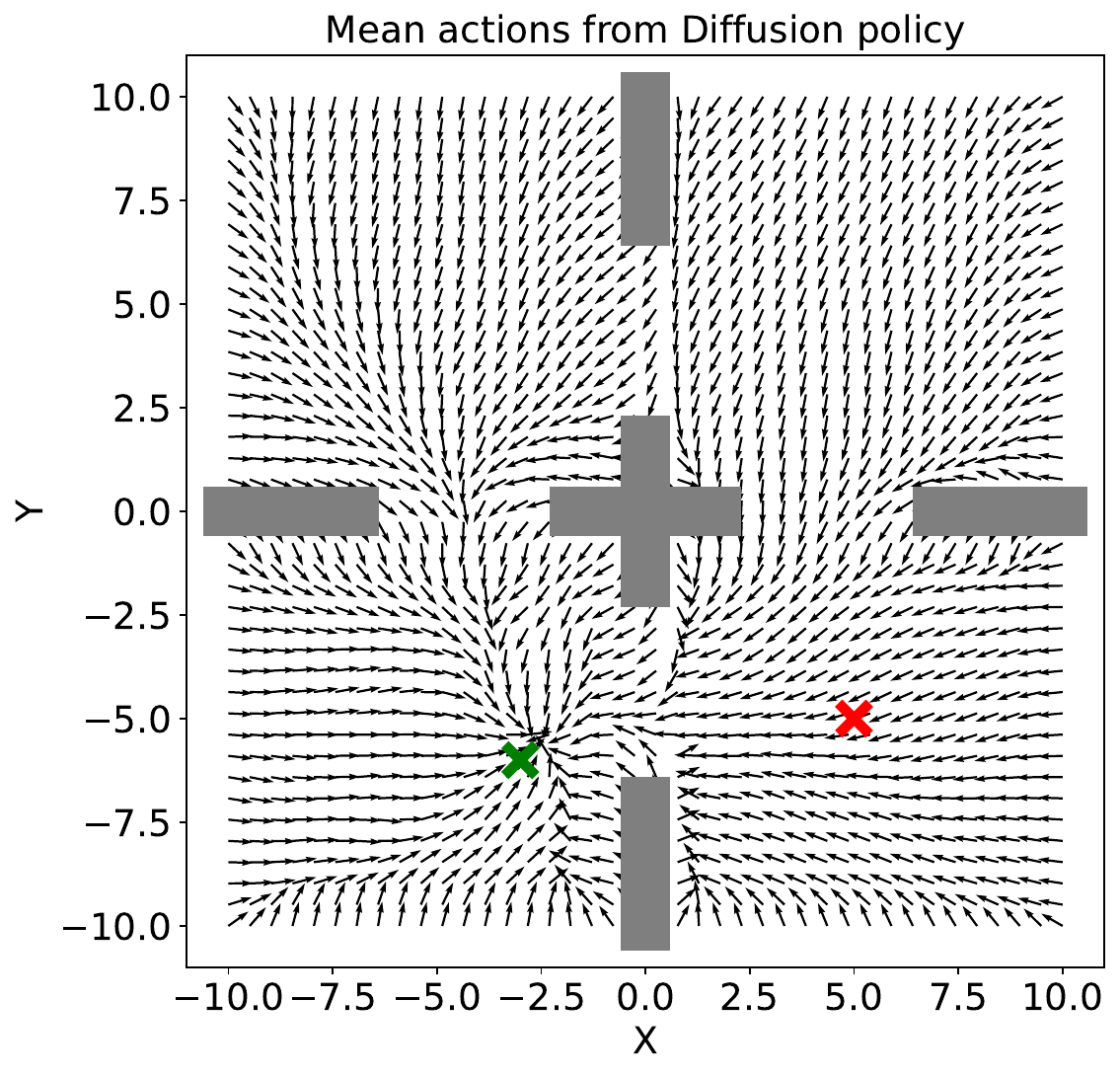}
  \end{subfigure}
  \begin{subfigure}[b]{0.245\textwidth}
      \centering
      \includegraphics[width=\textwidth]{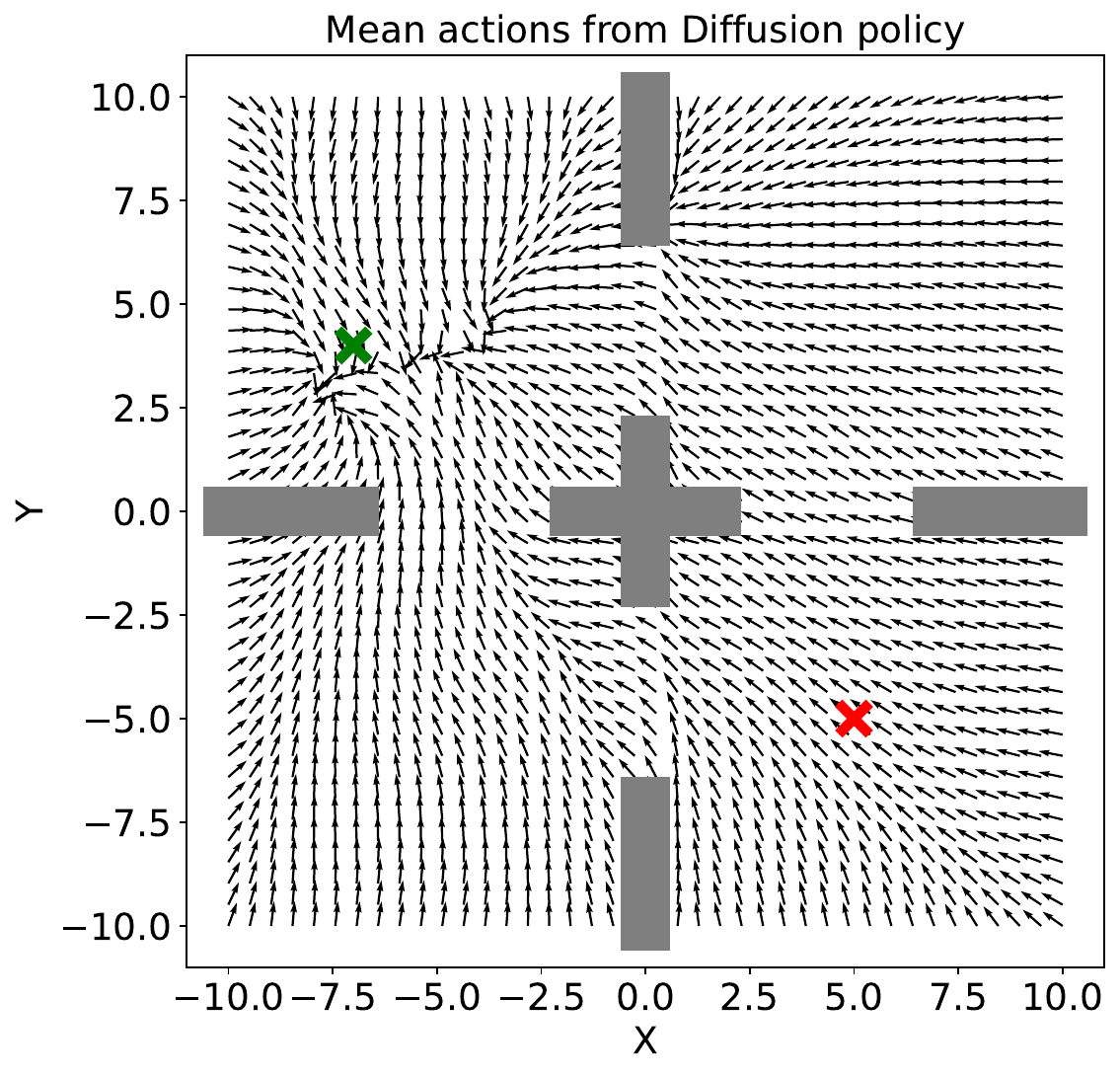}
  \end{subfigure}
  \newline
  \begin{subfigure}[b]{0.245\textwidth}
      \centering
      \includegraphics[width=\textwidth]{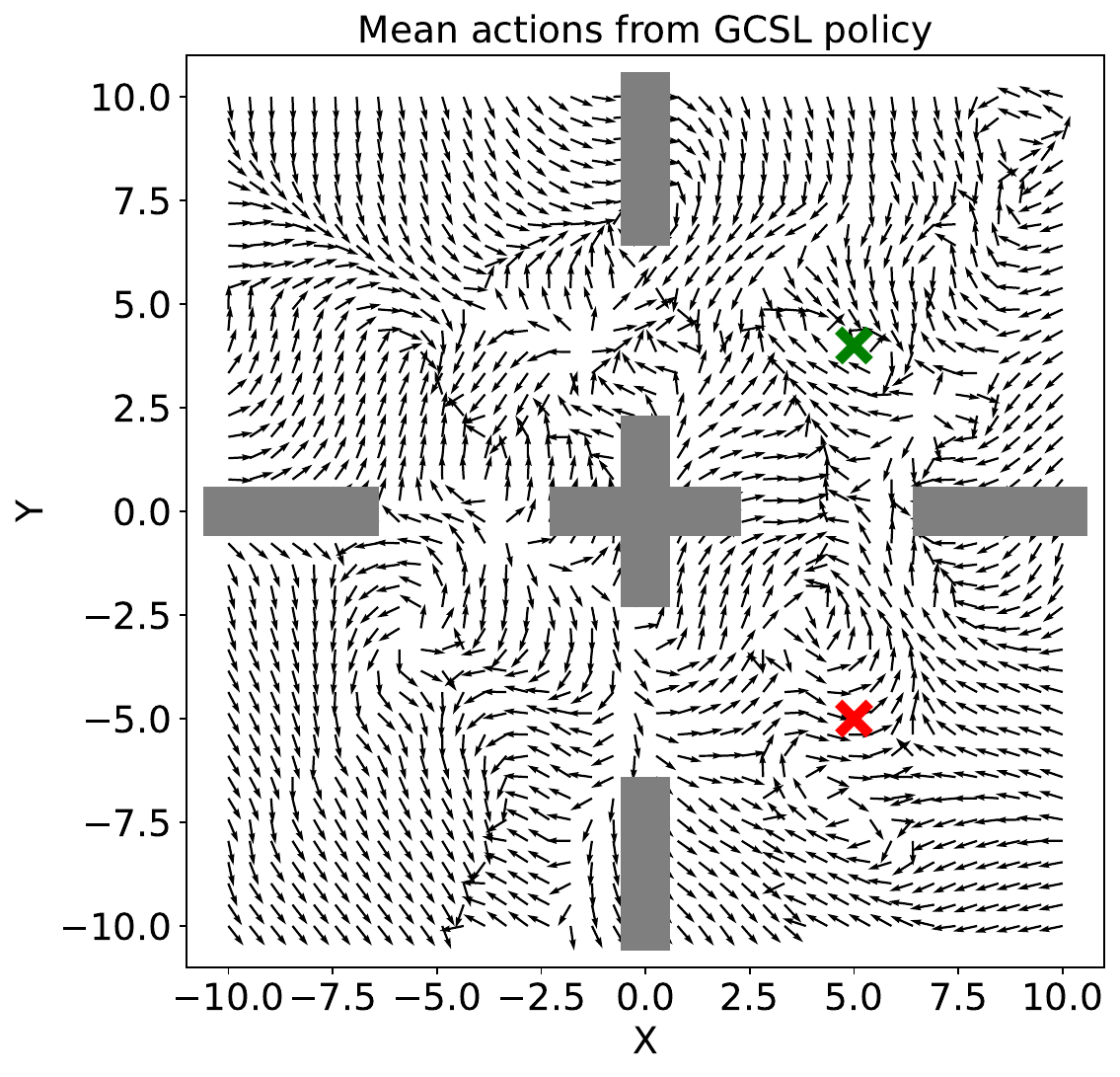}
  \end{subfigure}
  \begin{subfigure}[b]{0.245\textwidth}
      \centering
      \includegraphics[width=\textwidth]{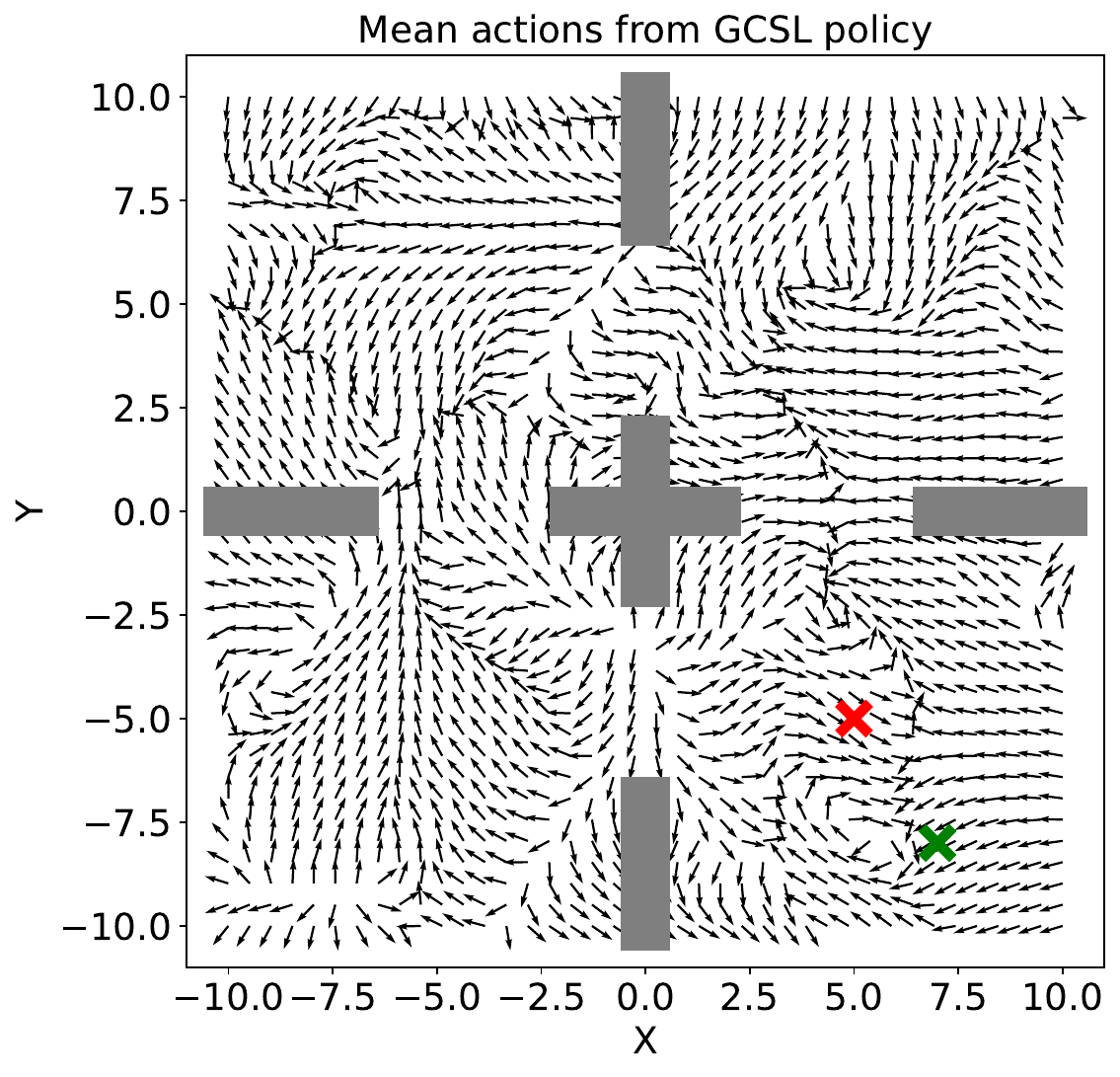}
  \end{subfigure}
  \begin{subfigure}[b]{0.245\textwidth}
      \centering
      \includegraphics[width=\textwidth]{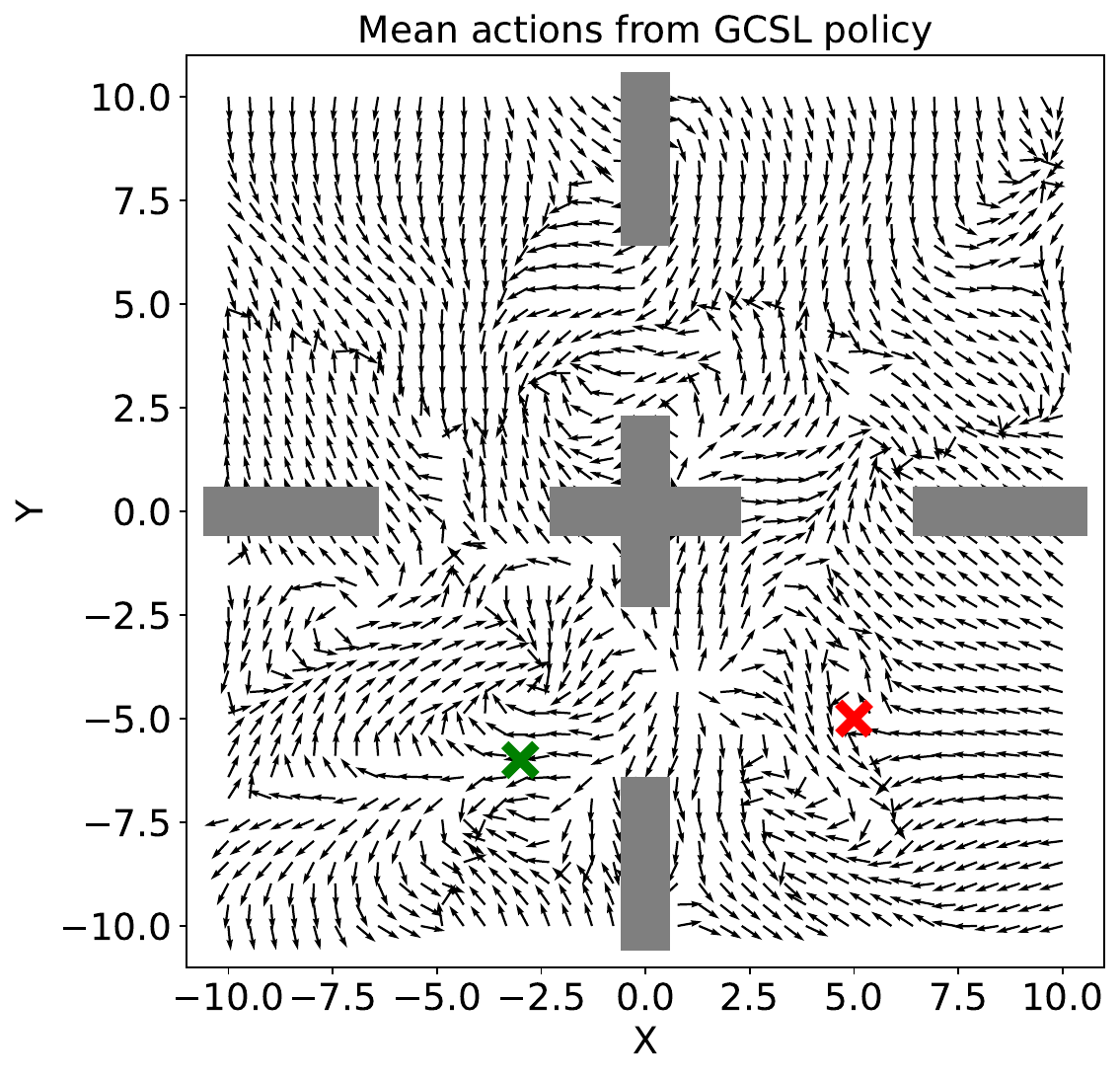}
  \end{subfigure}
  \begin{subfigure}[b]{0.245\textwidth}
      \centering
      \includegraphics[width=\textwidth]{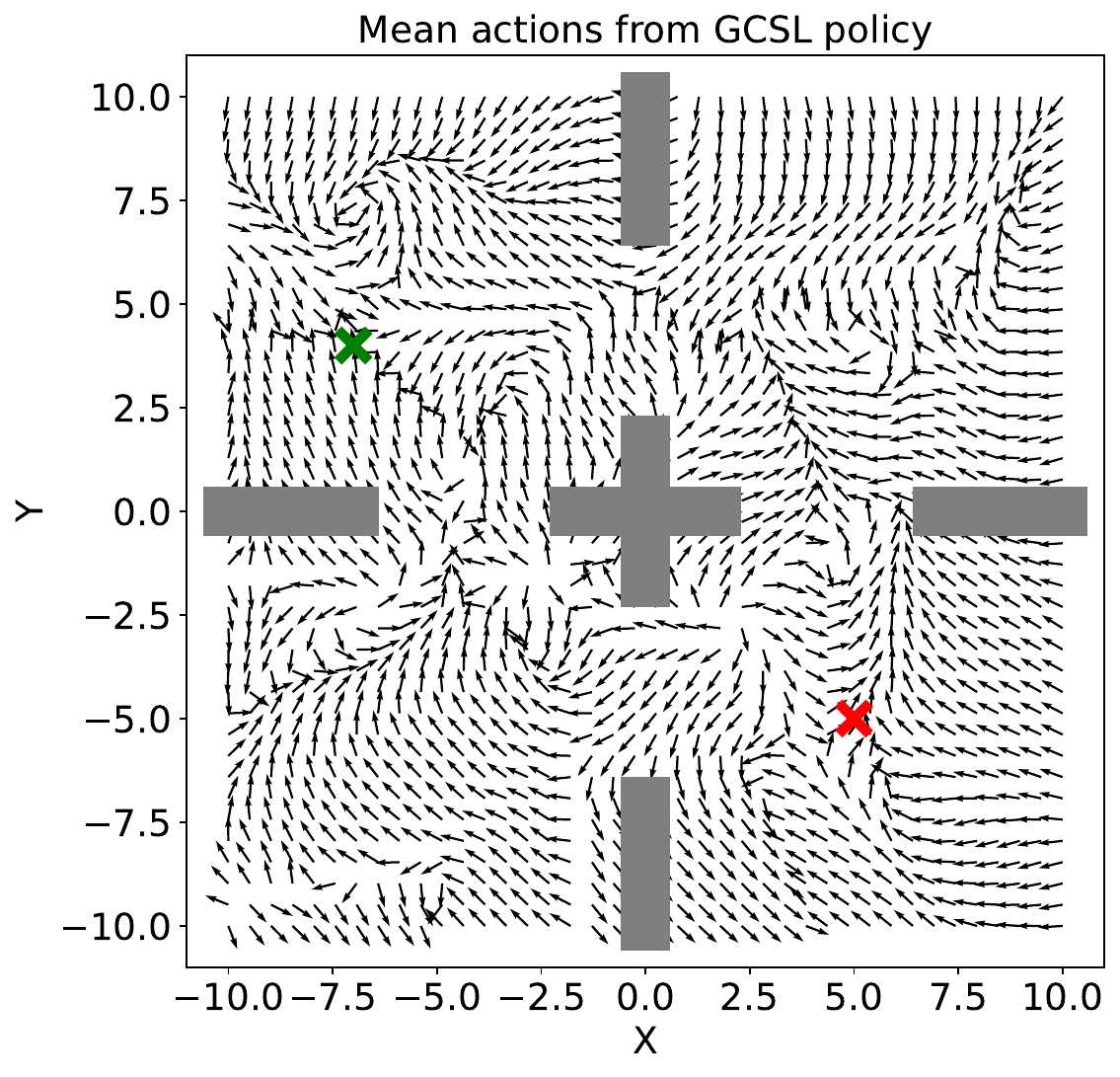}
  \end{subfigure}
  \vspace{-0.75em}
  \caption{\small{Evaluating the trained policy on out-of-distribution goals. Red \textbf{\textcolor{red}{X}} denotes the goal used during training, and green \textbf{\textcolor{teal}{X}} denotes the goal used for evaluation. \textbf{Top:} Diffusion; \textbf{Bottom:} GCSL.}}
  \vspace{-1.8em}
  \label{fig:diff_fourrooms_ood}
\end{figure}

We then evaluate the trained policy on out-of-distribution goals. During training, the goal is fixed to $g=(5,-5)$ but during evaluation, we condition the policy on random goals. As shown in \Cref{fig:diff_fourrooms_ood}, Merlin effectively generalizes this complex navigation behavior to new goals, learning to avoid the walls in most cases. One interesting case is when the goal is in the quadrant furthest away from the training goal. Here, Merlin has some difficulty navigating around the walls, particularly the walls in the center. This is likely due to insufficient data generated by the forward diffusion process in this quadrant (see \Cref{fig:diff_fourrooms_traj}).

\subsection{Multiple Goal Setting}
\label{app:multigoal}

\begin{figure}[!h]
  \centering
  \hspace{-3pt}\begin{subfigure}[t]{0.25\textwidth}
      \centering
      \includegraphics[width=\linewidth]{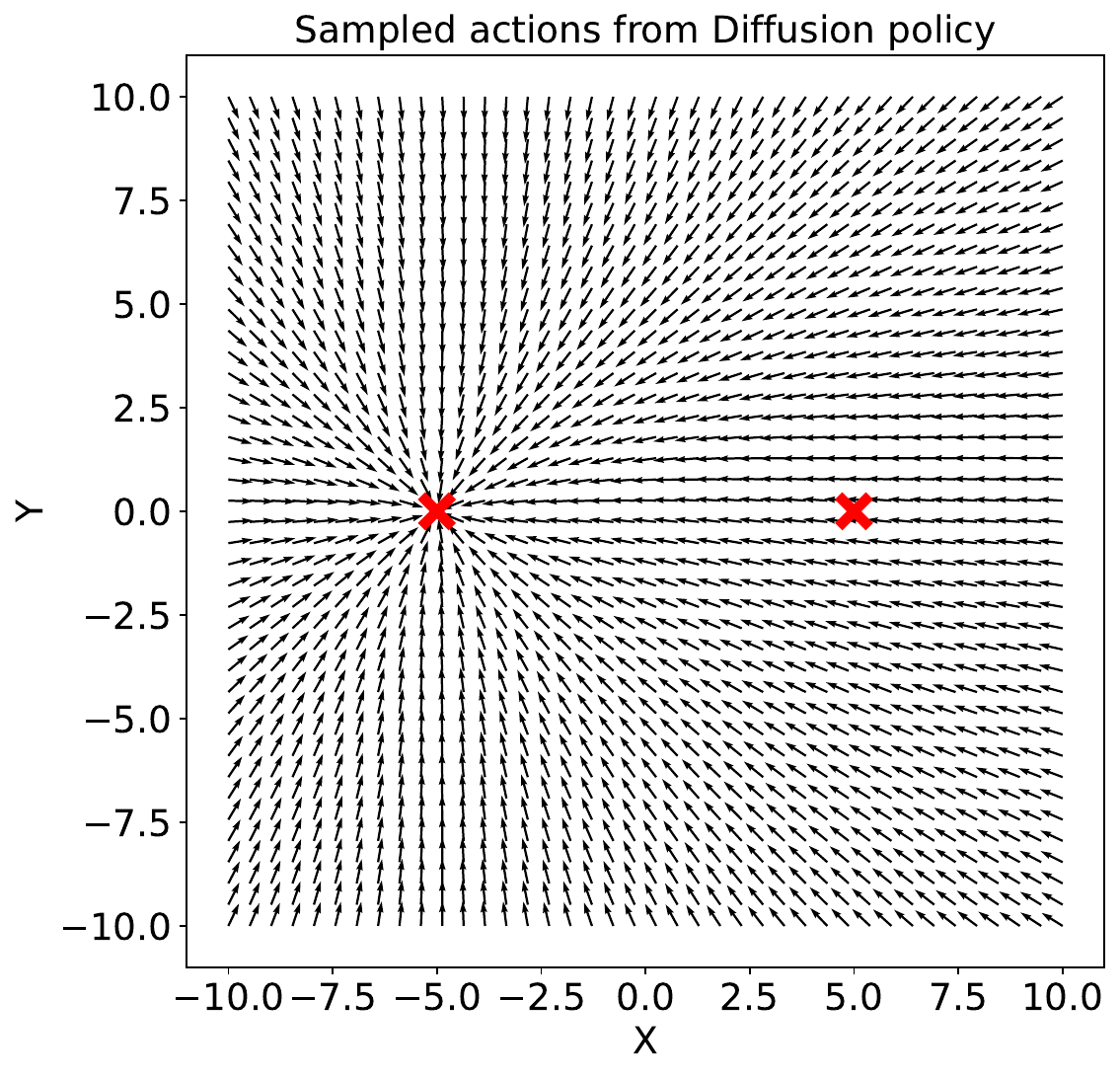}
  \end{subfigure}
  \centering
  \begin{subfigure}[t]{0.25\textwidth}
      \centering
      \includegraphics[width=\textwidth]{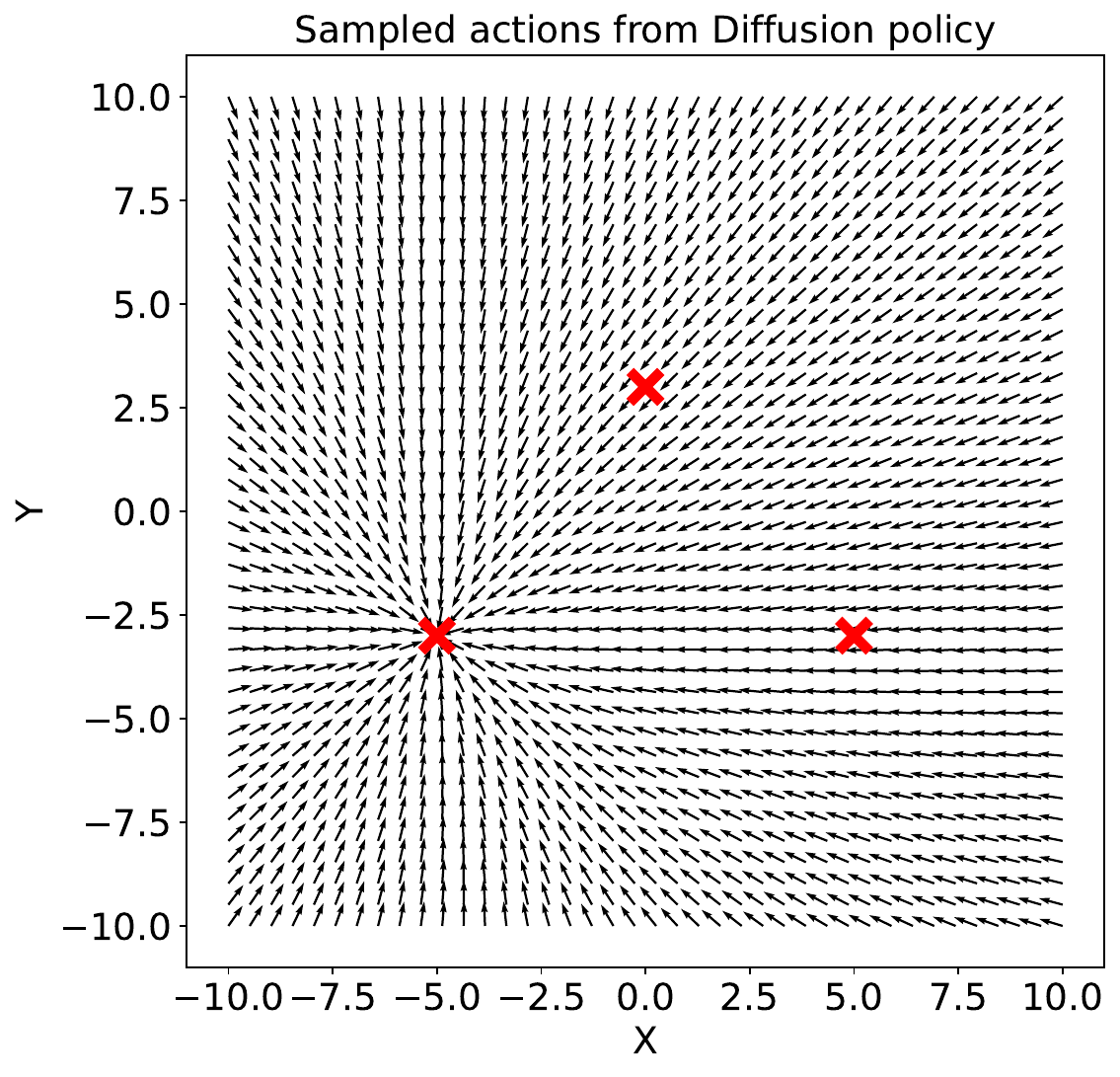}
  \end{subfigure}
  \centering
  \begin{subfigure}[t]{0.25\textwidth}
      \centering
      \includegraphics[width=\textwidth]{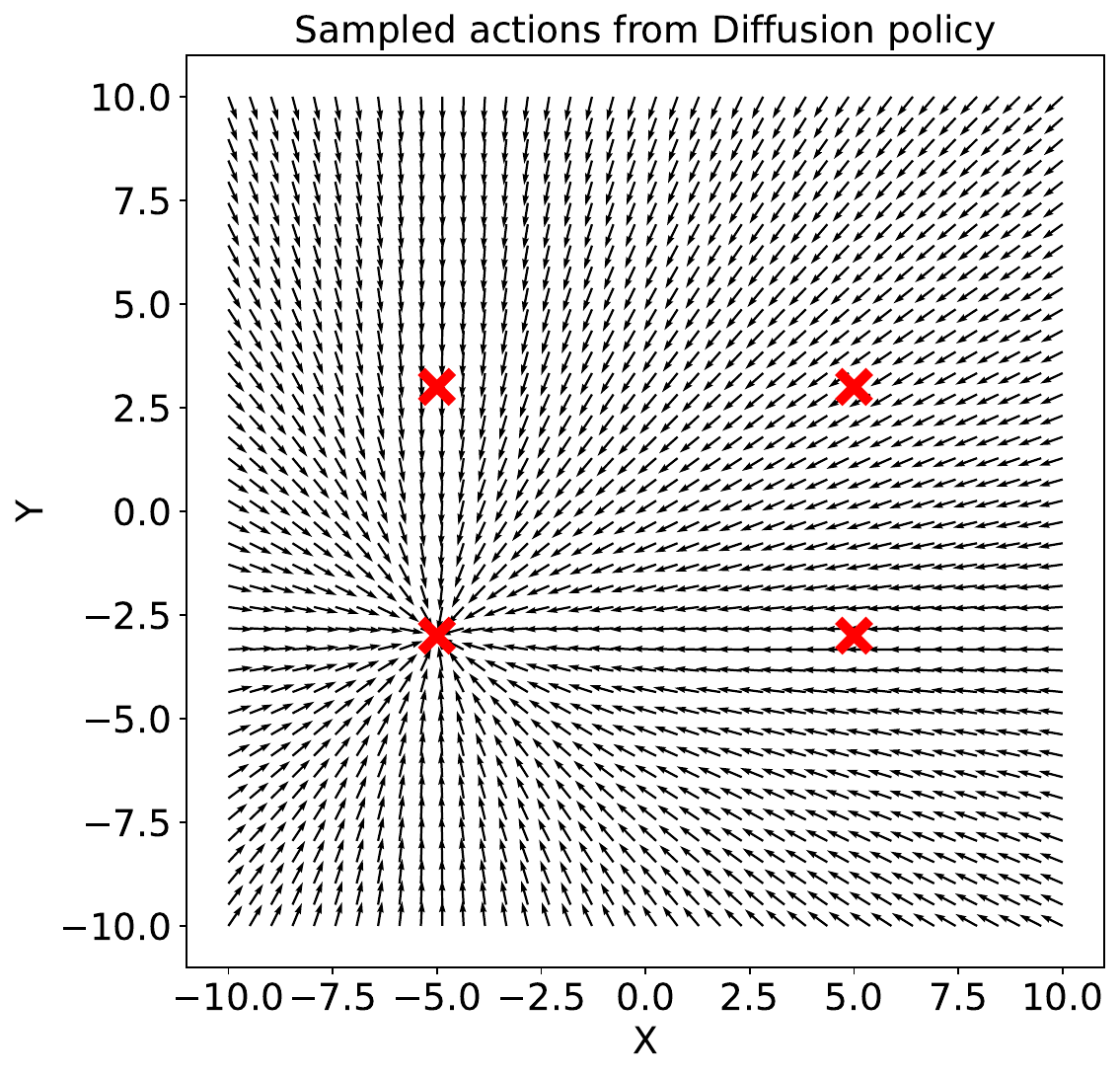}
  \end{subfigure}
  \centering
  \begin{subfigure}[t]{0.25\textwidth}
      \centering
      \includegraphics[width=\textwidth]{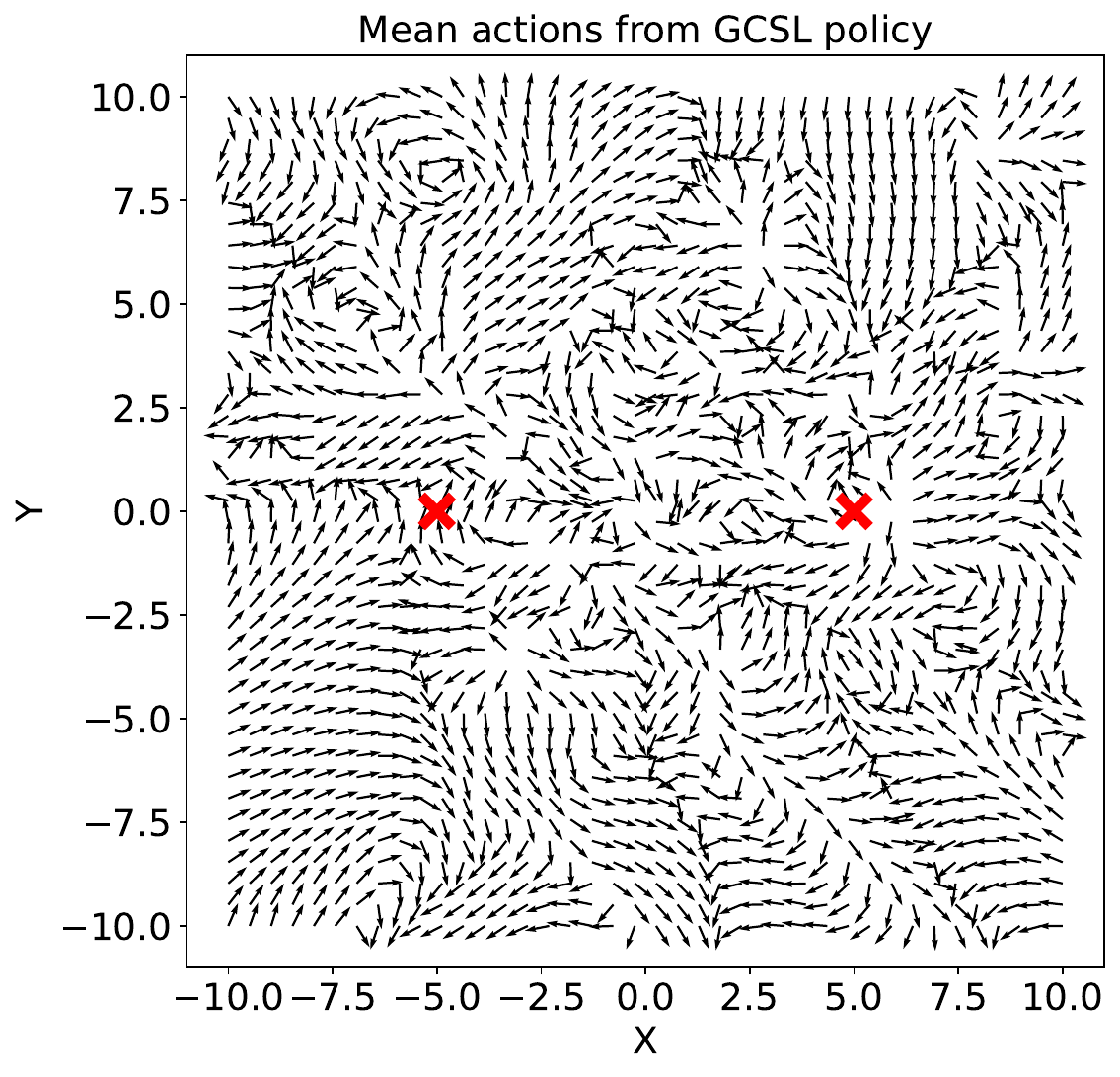}
  \end{subfigure}
  \centering
  \begin{subfigure}[t]{0.25\textwidth}
      \centering
      \includegraphics[width=\textwidth]{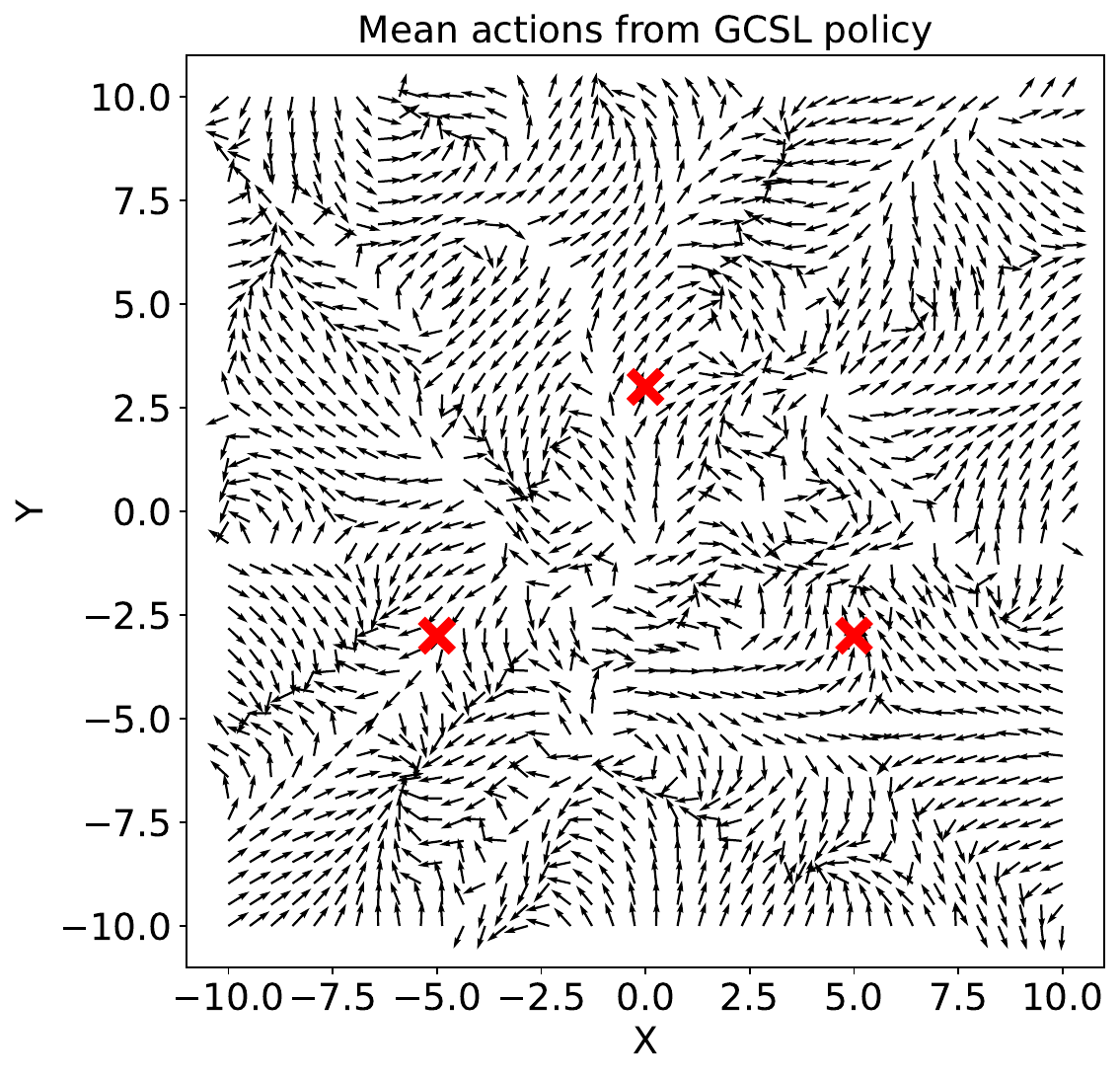}
  \end{subfigure}
  \centering
  \begin{subfigure}[t]{0.25\textwidth}
      \centering
      \includegraphics[width=\textwidth]{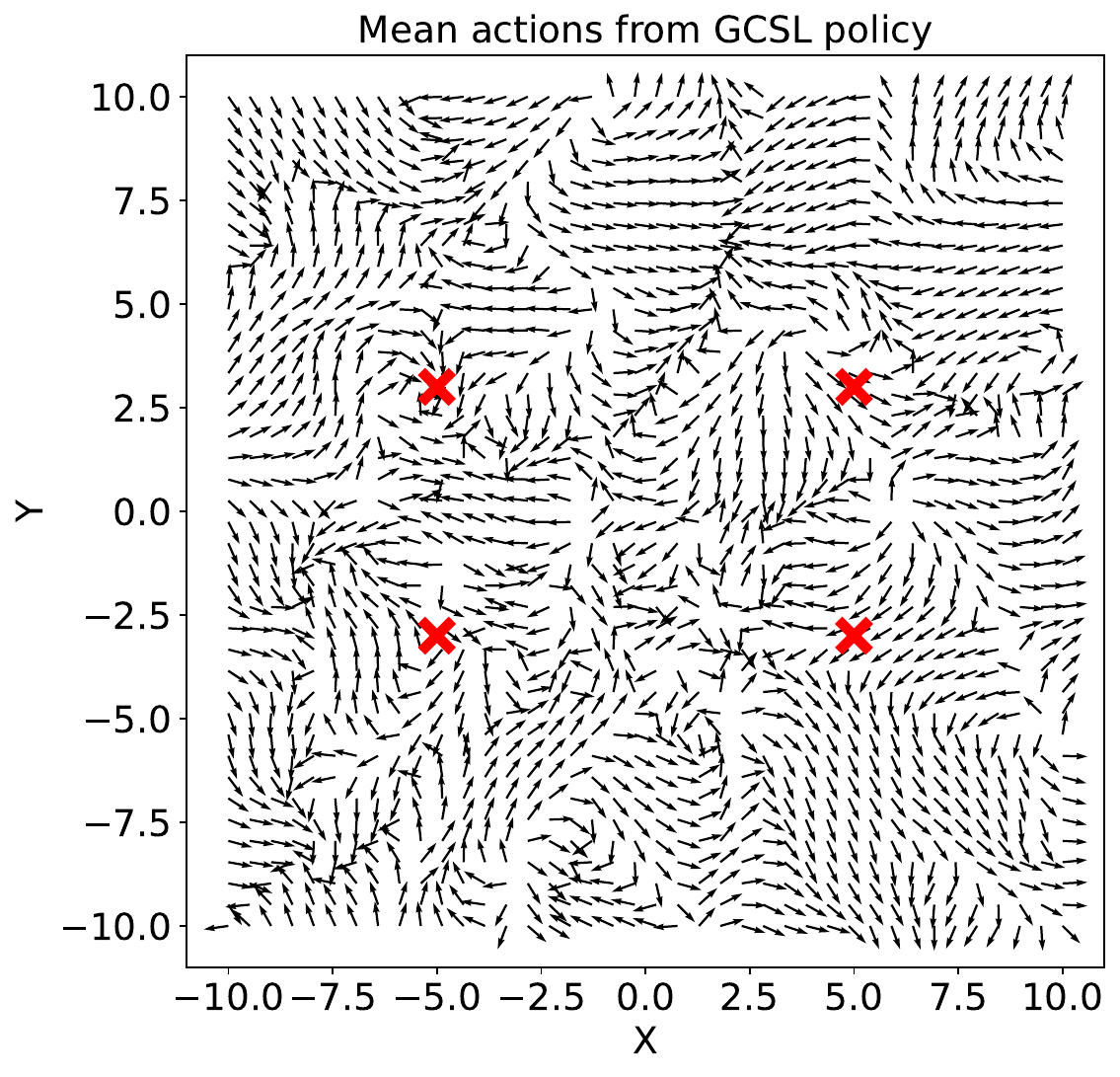}
  \end{subfigure}
  \vspace{-0.75em}
  \caption{\small{The policy is conditioned on the bottom leftmost goal in all cases. \textbf{Top:} Diffusion; \textbf{Bottom:} GCSL.}}
  \vspace{-1em}
  \label{fig:multidiff}
\end{figure}

The illustrative example presented in \Cref{sec:toy} considered a single goal setting. In this section, we verify that Merlin works as expected in the multiple goal setting. In these experiments, the forward diffusion process comprises taking random actions starting from one of the goals, which is picked randomly. A trained agent should be able to effectively navigate towards any one of these goals.

\Cref{fig:multidiff} shows the predicted actions from Merlin and GCSL for navigating towards one particular goal (fixed to be the bottom leftmost goal) among two, three, and four possible goals. Merlin successfully navigates to the specified goal taking the most optimal path in all cases, whereas GCSL struggles to reach the specified goal.

\section{Relation to Diffusion Probabilistic Models}
\label{app:relation}
Merlin takes inspiration from generative diffusion models, resulting in several parallels as highlighted earlier. Notably, \Cref{fig:diff_toy} visualizes the noisy trajectories generated via the forward diffusion process and conveys the similarity of the learned policy to the score function in generative diffusion models. However, there are several key differences:
\begin{itemize}[align=right,itemindent=0em,labelsep=0.5em,labelwidth=1em,leftmargin=2em,nosep]
\item The noise in diffusion probabilistic models is fixed to be Gaussian, whereas in application to RL, the noise corresponds to taking actions and reversing the dynamics, which is dependent on the properties of the MDP. 
\item The noisy samples for diffusion models lie outside the data manifold and hold no significance, while in this case, the noisy samples are valid states of the MDP.
\item Lastly, conditioning the policy on goals is different from class conditioning in diffusion - here, any state in the diffusion path is a potential goal.
\end{itemize}

\section{Implementation Details}
\label{app:implementation}

\subsection{Merlin Algorithm}
\label{app:algo}

\begin{algorithm}[h!]
    \centering
    \begin{algorithmic}
    \label{algo:merlin_full}
        \STATE \textbf{Input:} Dataset $\mathcal{D}$, hindsight ratio $p$, number of training steps $N$, \textcolor{purple}{number of new trajectories to collect $M$}.
        \STATE \textbf{Output: } Policy $\pi_\theta$
        \STATE \textcolor{red}{Train $f_\psi$, $E_\omega$, $D_\xi$ on $\mathcal{D}$ by minimizing Equation (6,7)}
        \STATE \textcolor{blue}{Train $h_\phi$ on $\mathcal{D}$ by minimizing Equation (8)}
        \STATE \textcolor{blue}{Construct ball tree $T$ for all states encoded using $h_\phi$}
        \STATE \texttt{\# Simulate forward diffusion process}
        \textcolor{purple}{\FOR{$m\gets 1$ {\bfseries to} $M$}
        \STATE \textcolor{purple}{Sample random final state $s_T$ from $\mathcal{D}$}
        \STATE $\tau_\textrm{new} \gets \{s_T\}$, $s_{\textrm{current}} \gets s_T$
        \FOR{$t\gets T$ {\bfseries to} $1$}
        \STATE \textcolor{red}{Sample $z \sim \mathcal{N}(0, \mathbf{I})$}
        \STATE \textcolor{red}{$a_\textrm{prev} \gets \hat{D}_\xi(s_{\textrm{current}},z)$, $s_\textrm{prev} \gets f_\psi(s_{\textrm{current}},a_\textrm{prev})$}
        \STATE \textcolor{blue}{$s_{\textrm{nbr}}, \textrm{sim} \gets T.\textrm{query}(h_\phi(s_{\textrm{current}}), k=1)$}
        \STATE \textcolor{blue}{$(s_\textrm{prev}, a_\textrm{prev}) \gets [s_\textrm{nbr}]_\textrm{prev}$ \textbf{if} $\textrm{sim} \geq \delta$ \textbf{else} $[s_\textrm{current}]_\textrm{prev}$}
        \STATE $\tau_\textrm{new} \gets \{s_\textrm{prev},a_\textrm{prev}\} \cup \tau_\textrm{new}$, $s_\textrm{current} \gets s_\textrm{prev}$  
        \ENDFOR
        \STATE $\mathcal{D}_\textrm{new} \gets \mathcal{D}_\textrm{new} \cup \tau_\textrm{new}$
        \ENDFOR}
        \STATE \textcolor{purple}{$\mathcal{D} \gets \mathcal{D}_\textrm{new} \cup \mcD$}
        \STATE \texttt{\# Train policy via reverse diffusion}
        \FOR{$n\gets 1$ {\bfseries to} $N$}
        \STATE Sample batch $(s,a,g)$ from $\mathcal{D}$
        \STATE Relabel fraction $p$ of batch 
        \STATE Update policy $\pi_\theta$ as per \Cref{eq:bc}
        \ENDFOR
        \STATE \textbf{Return:} $\pi_\theta$
    \end{algorithmic}
    \caption{Detailed Merlin algorithm. \textcolor{red}{Red} and \textcolor{blue}{blue} statements apply only for Merlin-P and Merlin-NP, respectively. \textcolor{purple}{Purple} statements apply to both.}
\end{algorithm}

\subsection{Merlin: Details of Policy Network and Hyperparameters}
\label{app:merlin}

The policy is parameterized as a diagonal Gaussian distribution using an MLP with three hidden layers of $256$ units each with the ReLU activation function, except for the final layer. The input to the policy comprises the state, the desired goal, and the time horizon. The time horizon is encoded using sinusoidal positional embeddings of $32$ dimensions with the maximum period set to $T=50$ since that is the maximum length of the trajectory for all our tasks. The output of the policy is the mean and the standard deviation of the action. The $\mathrm{tanh(\cdot)}$ function is applied to the mean and it is multiplied by the maximum value of the action space to ensure the mean is within the correct range. The softplus function, $\mathrm{softplus}(x) = \log(1+\exp(x))$ is applied to the standard deviation to ensure non-negativity.

The policy was trained for $500k$ mini-batch updates using Adam optimizer with a learning rate of $5\times 10^{-4}$ and a batch size of $512$. The same policy network architecture and corresponding hyperparameters are used for all variations of Merlin. Merlin involves two main hyperparameters - the hindsight ratio and the time horizon used during evaluation. We perform ablations in \Cref{sec:ablation} and report the tuned values for each task below.

\begin{table}[h]
\caption{Optimal values for the hindsight ratio and time horizon for Merlin.}
\vspace{-1em}
\label{table:hyperparam}%
\begin{center}
\begin{tabular}{|l|cc|cc|}
\hline \\[-2.55ex]
\multicolumn{1}{c}{\multirow{2}{*}{\bf Task Name}} & \multicolumn{2}{c|}{\bf Hindsight Ratio} & \multicolumn{2}{c|}{\bf Time Horizon} \\
\rule{0pt}{2ex}
& \multicolumn{1}{|c}{\bf Expert} & \multicolumn{1}{c|}{\bf Random} & \multicolumn{1}{c}{\bf Expert} & \multicolumn{1}{c|}{\bf Random}\\
\hline \rule{0pt}{3ex}%
PointReach & 0.2 & 1.0 & 1 & 1 \\
PointRooms & 0.2  & 1.0 & 1 & 1 \\
Reacher & 0.2 & 1.0 & 5 & 5 \\
SawyerReach & 0.2 & 1.0 & 1 & 1 \\
SawyerDoor & 0.2 & 1.0 & 5 & 5 \\
FetchReach & 0.2 & 1.0 & 1 & 1 \\
FetchPush & 0.0 & 0.2 & 20 & 20 \\
FetchPick & 0.0 & 0.5 & 10 & 50 \\
FetchSlide & 0.2 & 0.8 & 10 & 10 \\
HandReach & 1.0 & 1.0 & 1 & 1 \\
\hline
\end{tabular}
\end{center}
\end{table}

\subsection{Merlin-P: Details of Reverse Dynamics model and Reverse Policy}
\label{app:merlinp}

Merlin-P uses a learned parametric reverse dynamics model and a reverse policy to simulate the forward diffusion process starting from potential goal states. We use a reverse policy as described in \citet{wang2021offline} to generate previous actions given a state, and use a non-Markovian reverse model to generate forward diffusion trajectories.

To generate diverse candidate actions for reverse rollouts, the reverse policy is parameterized as a conditional variational autoencoder (CVAE), consisting of an action encoder $E_\omega(s_{t+1},a_t)$ that outputs a latent vector $z$, and an action decoder $D_\xi(s_{t+1},z)$ which reconstructs the action given latent vector $z$. The reverse policy is trained by maximizing the variational lower bound,
\begin{align*}
    \mathcal{L}(\omega, \xi) = \mathbb{E}_{(s_t,a_t,s_{t+1}) \sim \mathcal{D}, z \sim E_\omega(s_{t+1},a_t)} \left[ \left(a_t - D_\xi(s_{t+1},z)\right)^2 + D_{KL}\left(E_\omega(s_{t+1},a_t) || \mathcal{N}(0, \mathbf{I})\right)\right].
\end{align*}

The encoder is an MLP with two hidden layers of $256$ units each using the ReLU activation function. The latent space dimension is twice the action space dimension. The encoder outputs the mean and log standard deviation, the latter is clamped to $[-4,15]$ for numerical stability. The decoder is also an MLP with two hidden layers of $256$ units each using the ReLU activation function. The $\mathrm{tanh(\cdot)}$ function is applied to the action output of the decoder and it is multiplied by the maximum value of the action space to ensure it is within the correct range. The CVAE is trained for $20$ epochs using Adam optimizer with a learning rate of $3 \times 10^{-4}$ and a batch size of $256$.

The reverse dynamics model $\mcP_\psi(\cdot)$ produces the previous state given the future states and actions in a trajectory. The model parameters are optimized by minimizing the negative log-likelihood, which is equivalent to the mean squared error for deterministic environments,
\begin{align*}
    \mathcal{L}(\psi) = \mathbb{E}_{\tau \sim \mathcal{D}} \left[-\log \mcP_\psi(s|s',a)\right] = \mathbb{E}_{\tau \sim \mathcal{D}} \|s_t - f_\psi(a_t,s_{t+1},\ldots,s_T)\|_2^2\;,
\end{align*}
where $f_\psi(\cdot,\cdot)$ denotes the deterministic reverse dynamics function. In our implementation, the reverse dynamics model $f_\psi$ predicts the state difference $s_\Delta = s' - s$ instead of the absolute state $s$. The network architecture uses GRU \citep{cho2014learning} with one hidden layer of $256$ units with the ReLU activation function. The dynamics model is trained for $20$ epochs using Adam optimizer with a learning rate of $3 \times 10^{-4}$ and a batch size of $256$. 

In order to generate a rollout starting from state $s_{t+1}$, a latent vector is drawn from the standard Gaussian distribution, $z \sim \mathcal{N}(0,\mathbf{I})$. The action decoder is used to obtain a candidate action $a_t = D_\xi(s_{t+1},z)$, and finally the reverse dynamics model produces the previous state $s_t = f_\psi(a_t,s_{t+1},\ldots,s_T)$. 

For both the reverse policy and the reverse dynamics model with image inputs, we use a Convolutional Neural Network (CNN) to produce latent representations of the images, and use the latent states as inputs or outputs to the CVAE or the GRU network. The CNN architecture is given in \Cref{table:cnn}. 

\subsection{Merlin-NP: Details of nearest-neighbor trajectory stitching}
\label{app:merlinnp}

We propose a novel trajectory stitching method in \Cref{sec:traj} which is used for Merlin-NP. The method is based on finding the nearest neighbor of states along a trajectory. However, nearby states might not necessarily be connected. For example, consider the walled 2D navigation example where an agent must navigate in a room with walls towards a goal position. Positions on two sides of a wall are nearby based on a distance metric (such as Euclidean distance) but are not reachable directly. Therefore, we learn state representations using contrastive learning, such that consecutive states are positive pairs. This encourages the representations corresponding to connected states to be nearby in the latent space. The loss function to train the contrastive encoder $h_\phi$ is given by,
\begin{align*}
    \mathcal{L}(\phi) = \mathbb{E}_{(s,s') \sim \mathcal{D}}\left[-\log\frac{\exp(h_\phi(s)^\top h_\phi(s'))}{\sum_{s'_\textrm{neg} \in \mathcal{D}} \exp(h_\phi(s)^\top h_\phi(s'_\textrm{neg}))}\right].
\end{align*}
The encoder is an MLP with three hidden layers of $256$ units each with the ReLU activation function. For image states, we use a CNN encoder with the architecture described in \Cref{table:cnn}. We train the encoder for $20$ epochs and the parameters are optimized using the Adam optimizer with a learning rate of $5\times 10^{-4}$. We can then apply the method described in \Cref{sec:traj} in the latent space. Since contrastive representations lie on a $d$-dimensional hypersphere, where $d$ is the latent space dimension, we choose cosine similarity to identify nearest neighbors. The use of distance metrics in high dimensions can be unreliable, however, note that all methods implicitly assume a metric. The policy and the value function assume a metric to determine which states are similar, and therefore, should yield similar actions or values respectively. 

In order to search for nearest neighbors as efficiently as possible, we construct a ball tree from all the states in the dataset, instead of a KD tree. The ball tree partitions the space using a series of hyperspheres instead of partitioning along the Cartesian axes, which leads to more efficient queries in higher dimensions. The query time for the ball tree grows as approximately $O(d \log N)$ for $N$ samples of $d$-dimensional data. For a KD tree, the query time is the same as a ball tree for lower dimensions ($<20$), however, it quickly becomes comparable to a brute force search for higher dimensions.

\begin{table}[h]
    \caption{CNN Architecture.}
    \vspace{-1em}
    \label{table:cnn}%
    \begin{center}
    \begin{tabular}{|l|c|}
    \hline \rule{0pt}{3ex}%
    {\bf Parameter} & {\bf Value}\\
    \hline \rule{0pt}{3ex}%
    Channels & $32, 64, 64$\\
    Filter size & $8, 4, 3$\\
    Stride & $4, 2, 1$\\
    Non-linearity & ReLU\\
    Linear layer & 256\\
    \hline
    \end{tabular}
    \end{center}
\end{table}

For the trajectory stitching method, we also choose a similarity threshold $\delta$ which determines which states are considered similar enough to allow stitching. If the cosine similarity between two states is greater than $\delta$, we consider those states as candidates for stitching. Very small values of $\delta$ would result in constant switching between trajectories even when the states are considerably dissimilar, leading to mismatched state-action pairs at the stitching point. Empirically, we observed that setting $\delta = 0.9999$ seemed to be appropriate, where this value was chosen such that there would be 2-3 stitching operations per trajectory.

We collect $2000$ stitched trajectories for the simpler tasks (PointReach, PointRooms, Reacher, SawyerReach, SawyerDoor and FetchReach), which effectively doubles the amount of offline data. For the harder tasks (FetchPush, FetchPick, FetchSlide and HandReach), we collect $10000$ stitched trajectories to augment the $40000$ trajectories in the original dataset.

\subsection{Compute}
\label{app:compute}

\begin{figure}[h]
  \centering
  \begin{subfigure}[b]{0.4\textwidth}
      \centering
      \includegraphics[width=\linewidth]{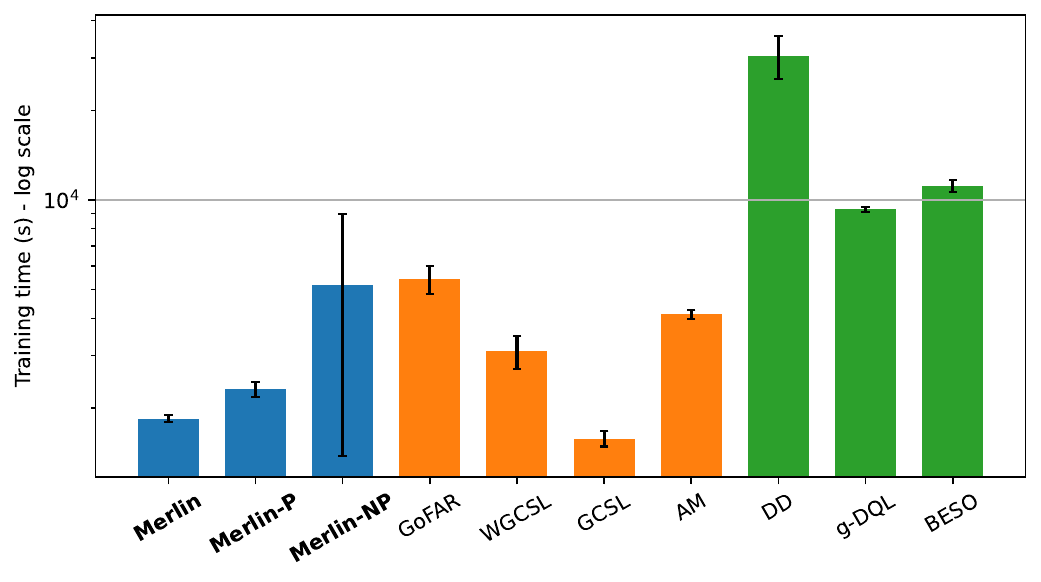}
  \end{subfigure}
  \begin{subfigure}[b]{0.405\textwidth}
      \centering
      \includegraphics[width=\textwidth]{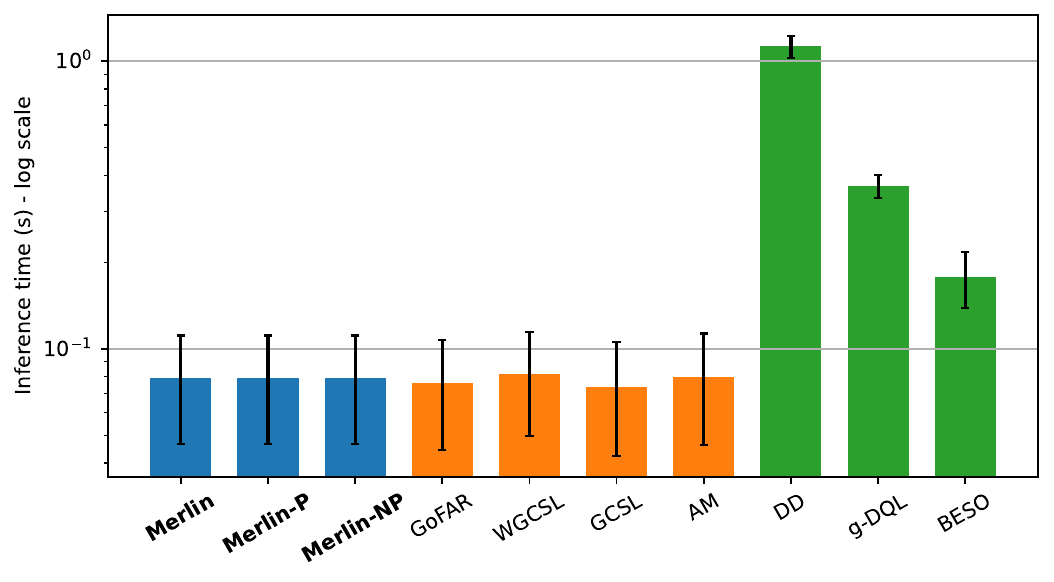}
  \end{subfigure}
  \vspace{-0em}
  \caption{\small{Mean training and inference times over different tasks for each method. Training times are reported for $500k$ policy updates and inference times are reported for one episode comprising of $50$ time steps.}}
  \vspace{-1em}
  \label{fig:times_app}
\end{figure}

\Cref{fig:times_app} shows the training and inference times averaged over all tasks for each method in \Cref{sec:exp}. The diffusion-based baselines (DD, g-DQL and BESO) have significantly higher training and inference times since each environment step requires denoising the entire reverse diffusion chain. In contrast, Merlin has comparable training and inference times to non-diffusion-based offline GCRL methods, which is roughly an order of magnitude lower. Merlin-P suffers from an overhead compared to Merlin due to training the reverse dynamics model and the reverse policy. The training time overhead for Merlin-NP is during the trajectory stitching phase for nearest-neighbors search. The large variation in training time for Merlin-NP is because trajectory stitching for the harder tasks (FetchPush, FetchPick, FetchSlide and HandReach) takes more time owing to the higher state-space dimension and a larger number of collected trajectories.

\begin{figure}[h]
  \centering
  \begin{subfigure}[b]{0.4\textwidth}
      \centering
      \includegraphics[width=\linewidth]{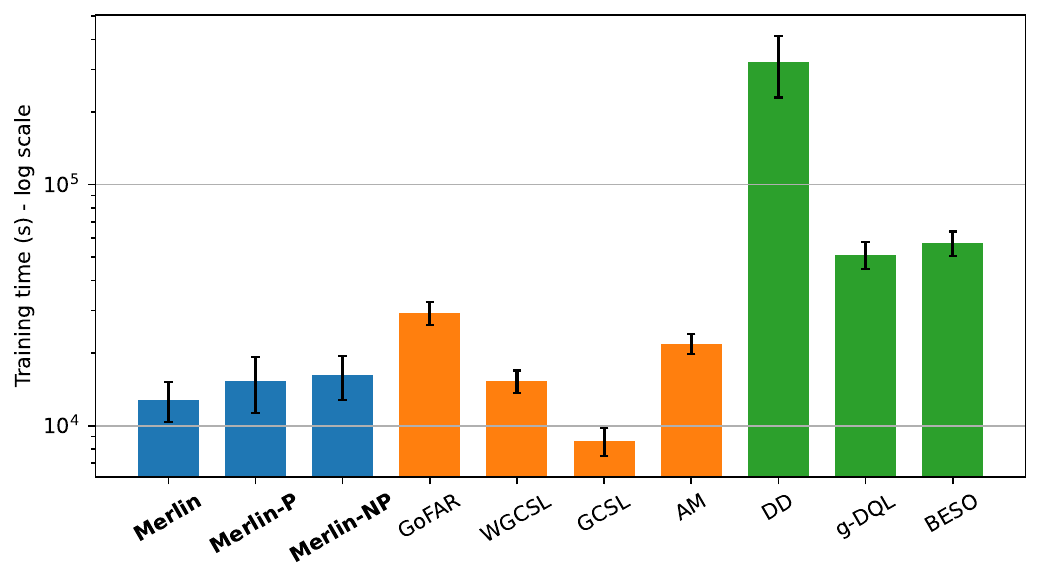}
  \end{subfigure}
  \begin{subfigure}[b]{0.405\textwidth}
      \centering
      \includegraphics[width=\textwidth]{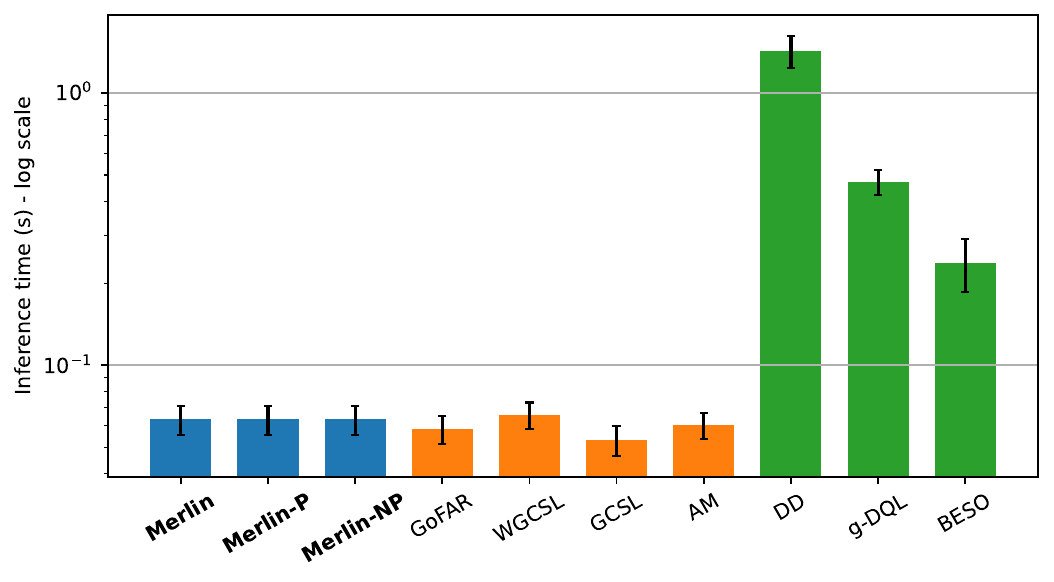}
  \end{subfigure}
  \vspace{-0em}
  \caption{\small{Mean training and inference times over different tasks for each method, using image observations. Training times are reported for $500k$ policy updates and inference times are reported for one episode comprising of $50$ time steps.}}
  \vspace{-1em}
  \label{fig:times_app_img}
\end{figure}

\Cref{fig:times_app_img} shows the training and inference times averaged over all tasks for image observations. We observe a similar trend as before, except that the efficiency improvement of Merlin over the baseline diffusion-based methods is even more pronounced. 

\section{Baseline Implementation Details}
\label{app:baselines}

\subsection{Offline GCRL methods}

For all the methods described in this section, the policy architecture is identical to the one used for Merlin, described in \Cref{app:implementation}. Wherever applicable, the critic architecture is an MLP with three hidden layers of $256$ units each with the ReLU activation. All of these methods were fine-tuned in \citet{ma2022offline}, and we used their implementation (\href{https://github.com/JasonMa2016/GoFAR/tree/main}{https://github.com/JasonMa2016/GoFAR/tree/main}) to produce the baseline results in \Cref{sec:exp}.

\paragraph{GCSL.} GCSL uses hindsight relabeling by setting the goal to be a future state within the same trajectory, where future states are sampled uniformly from possible choices. The policy is learned using behavior cloning,
\begin{align*}
    \max_\pi \mathbb{E}_{(s,a,g) \sim \mathcal{D}_\textrm{relabel}}\left[\log \pi(a|s,g)\right]
\end{align*}

\paragraph{WGCSL.} This method builds upon GCSL and learns a Q-function with standard TD learning, where the dataset $\mathcal{D}$ uses hindsight relabeling and $\bar{Q}$ denotes the stop-gradient operation,
\begin{align*}
    \min_Q \mathbb{E}_{(s_t, a_t, s_{t+1}, g) \sim \mathcal{D}} \left[ \left(r(s_t, g) + \gamma \bar{Q}(s_{t+1}, \pi(s_{t+1}, g), g) - Q(s_t, a_t, g)\right)^2 \right]
\end{align*}
The advantage function is defined as $A(s_t, a_t, g) = r(s_t, g) + \gamma Q(s_{t+1}, \pi(s_{t+1}, g), g) - Q(s_t, \pi(s_t, g), g)$, and is used to weight the regression loss for policy updates,
\begin{align*}
    \max{\pi} \mathbb{E}_{(s_t, a_t, \phi(s_i)) \sim \mathcal{D}} \left[ \gamma^{i-t} \exp_\textrm{clip}(A(s_t, a_t, \phi(s_i))) \log \pi(a_t | s_t, \phi(s_i)) \right]
\end{align*}

\paragraph{Actionable Model.} AM employs an actor-critic framework similar to DDPG \citep{lillicrap2015continuous}, but uses conservative critic updates by adding a regularization term to the regular TD updates,
\begin{align*}
    \min_Q \mathbb{E}_{(s_t, a_t, s_{t+1}, g) \sim \mathcal{D}} \left[ \left(r(s_t, g) + \gamma \bar{Q}(s_{t+1}, \pi(s_{t+1}, g), g) - Q(s_t, a_t, g)\right)^2  +  \mathbb{E}_{a \sim \exp(\bar{Q})}[Q(s, a, g)] \right]
\end{align*}
The policy updates are similar to DDPG, where gradients are backpropagated through the critic,
\begin{align*}
    \max_\pi \mathbb{E}_{(s_t, a_t, s_{t+1}, g) \sim \mathcal{D}} \left[ Q(s_t, \pi(s_t, g), g) \right]
\end{align*}
In addition to hindsight relabeling, AM uses a goal-chaining technique where for half of the relabeled transitions in each minibatch, the relabelled goals are randomly sampled from the offline dataset.

\paragraph{GoFAR.} GoFAR takes a state-occupancy matching perspective by training a discriminator to define a reward function that encourages visiting states that occur more often in conjunction with the desired goal,
\begin{align*}
    \min_c \mathbb{E}_{g \sim p(g)} \left[ \mathbb{E}_{p(s,g)} \left[ \log c(s, g) \right] + \mathbb{E}_{(s,g) \sim \mathcal{D}} \left[ \log(1 - c(s, g)) \right] \right]
\end{align*}
where $p(s,g) = \exp(r(s,g))/Z$ and $Z=\int \exp(r(s,g))$. The reward function used for learning the critic is $R(s,g) = -\log\left(1/c(s,g)-1\right)$.
GoFAR also uses $f$-divergence regularization to learn a value function,
\begin{align*}
    \min_{V(s, g) \geq 0} (1 - \gamma) \mathbb{E}_{s \sim \mu(s), g \sim p(g)} [V(s, g)] + \mathbb{E}_{(s, a, g) \sim \mathcal{D}} [f_\star(R(s, g) + \gamma \mathbb{E}_{s' \sim \mcP(\cdot | s,a)}[V(s', g)] - V(s, g))]
\end{align*}
where $f_\star$ denotes the convex conjugate of $f$. The policy is updated using regression weights that are first-order derivatives of $f_\star$ evaluated at the optimal advantage,
\begin{align*}
    \max_{\pi} \mathbb{E}_{g \sim p(g)} \mathbb{E}_{(s, a) \sim \mathcal{D}} \left[f'_\star(R(s, g) + \gamma \mathbb{E}_{s' \sim \mcP(\cdot | s,a)}[V^*(s', g)] - V^*(s, g)) \log \pi(a | s, g)\right]
\end{align*}
where $V^*$ denotes the optimal value function obtained after training. GoFAR does not use hindsight relabeling.

\subsection{Diffusion-based methods}

\paragraph{Decision Diffuser.} Decision diffuser models sequential decision-making as a conditional generative modeling problem,
\begin{align*}
    \max_{\theta} E_{\tau \sim \mcD} [\log p_{\theta}(\bx_0(\tau) | \by(\tau))]
\end{align*}
where $\by(\tau)$ denotes the conditioning variable representing returns, goals, or other constraints that are desirable in the generated trajectory. The forward and reverse diffusion process are $q(\bx_{k+1}(\tau) | \bx_k(\tau))$ and $p_{\theta}(\bx_{k-1}(\tau) | \bx_k(\tau), \by(\tau))$. The diffusion is performed on state sequences, and actions are obtained using an inverse dynamics model $a_t = f_\psi(s_t, s_{t+1})$. The reverse diffusion process learns a conditional denoising function by sampling noise $\epsilon \sim \mathcal{N(\mathbf{0},\mathbf{I})}$ and a timestep $k \sim \mathcal{U}\{1,\ldots, K\}$,
\begin{align*}
    \min_{\theta, \psi} \mathbb{E}_{k,\tau \in D, \beta \sim \textrm{Bern}(p)}\left[ \left\| \epsilon - \epsilon_{\theta}\left(\bx_k(\tau), (1-\beta)\by(\tau) + \beta\varnothing, k\right) \right\|_2^2 \right] + \mathbb{E}_{(s,a,s')\in \mcD}\left[ \left\| a - f_{\psi}(s, s') \right\|_2^2 \right]
\end{align*}
Classifier-free guidance is employed during planning to generate trajectories respecting the conditioning variable $\by(\tau)$ by starting with Gaussian noise $\bx_K(\tau)$ and refining
$\bx_k(\tau)$ into $\bx_{k-1}(\tau)$ at each intermediate timestep with the perturbed noise,
\begin{align*}
    \epsilon = \epsilon_{\theta}\left(\bx_k(\tau), \varnothing, k\right) + \omega\left(\epsilon_{\theta}\left(\bx_k(\tau), \by(\tau), k\right) - \epsilon_{\theta}\left(\bx_k(\tau), \varnothing, k\right)\right)
\end{align*}
The denoising function is a temporal U-Net model with residual blocks. We used the official implementation provided here: \href{https://github.com/anuragajay/decision-diffuser/tree/main}{https://github.com/anuragajay/decision-diffuser/tree/main}.

\paragraph{Diffusion QL.} The policy is represented via the reverse process of a conditional diffusion model, where the end sample of the reverse chain, $a^0$, is the action used for RL evaluation,
\begin{align*}
    \pi_{\theta}(a | s) = p_{\theta}(a^{0:N} | s) = \mathcal{N}(a^N; \mathbf{0}, \mathbf{I}) \prod_{i=1}^{N} p_{\theta}(a^{i-1} | a^i, s)
\end{align*}
The reverse process is modeled as a noise prediction model with fixed variance $\Sigma_i = \beta_i \mathbf{I}$. The mean is reparameterized in terms of a learned denoising function $\epsilon_\theta$, which is trained using the simplified objective proposed in \citet{ho2020denoising},
\begin{align*}
    \min_\theta \mathbb{E}_{i \sim U\{1,\ldots,N\}, \epsilon \sim \mathcal{N}(\mathbf{0},\mathbf{I}), (s,a) \sim \mcD} \left[ \left\| \epsilon - \epsilon_{\theta}\left(\sqrt{\bar{\alpha}_i}a + \sqrt{1 - \bar{\alpha}_i}\epsilon, s, i\right) \right\|_2^2 \right]
\end{align*}
To sample actions from the policy, first sample $a^N \sim \mathcal{N}(\mathbf{0}, \mathbf{I})$ and denoise using the denoising model for $N$ steps,
\begin{align*}
    a^{i-1} \mid a^i = \frac{a^i}{\sqrt{\alpha_i}} - \frac{\beta_i}{\sqrt{\alpha_i(1 - \bar{\alpha}_i)}} \epsilon_{\theta}(a^i, s, i) + \sqrt{\beta_i}\epsilon, \quad \epsilon \sim \mathcal{N}(\mathbf{0}, \mathbf{I}), \text{ for } i = N, \ldots, 1.
\end{align*}
Similar to DDPG, Diffusion-QL also uses a learned critic function trained using the standard TD error and backpropagates through the critic during training to prefer actions with high Q-values. In order to apply this method to goal-conditioned tasks, we additionally condition the policy on the goal. The architecture of the policy is a three-layer MLP with $256$ hidden units each and the Mish activation function. The critic similarly has three layers of $256$ units each and Mish activations. We used the official implementation provided here: \href{https://github.com/Zhendong-Wang/Diffusion-Policies-for-Offline-RL}{https://github.com/Zhendong-Wang/Diffusion-Policies-for-Offline-RL}.

\paragraph{BESO.} Given a dataset $\mcD$, BESO proposes to model the distribution of actions conditioned on the states and goals, $p(a|s,g)$ using a continuous time stochastic differential equation (SDE),
\begin{align*}
    da = (\beta_t \sigma_t - \dot{\sigma}_t) \sigma_t \nabla_a \log p_t(a | s, g) \, dt + \sqrt{2 \beta_t} \sigma_t \, d\omega_t,
\end{align*}
where $\nabla_a \log p_t(a | s, g)$ refers to the score function, $\omega_t$ is the Standard Wiener process, $\sigma_t$ is the noise scheduler, and $\beta(t)$ describes the relative rate at which the current noise is replaced by new noise. The corresponding probability flow ODE is,
\begin{align*}
    da = - \dot{\sigma}_t \sigma_t \nabla_a \log p_t(a | s, g) \, dt \;.
\end{align*}
The score function is parameterized using a neural network $D_\theta(a,s,g,\sigma_t)$, which matches the score function,
\begin{align*}
    \nabla_a \log p_t(a | s, g) = (D_\theta(a,s,g,\sigma_t) - a)/\sigma_t \; .
\end{align*}
The network is trained using the denoising score matching objective, where Gaussian noise $\epsilon \sim \mathcal{N}(\mathbf{0}, \sigma_t\mathbf{I})$ is added to the actions and the mean squared error with the original actions is minimized,
\begin{align*}
    \mathcal{L}(\theta) = \mathbb{E}_{\sigma_t, a, \epsilon} \left[ \alpha(\sigma_t) \lVert D_\theta (a + \epsilon, s, g, \sigma_t) - a \rVert^2 \right]
\end{align*}
During the action prediction, a random Gaussian sample is iteratively denoised N-discrete noise levels using the DDIM solver \citep{song2020denoising} for fast, deterministic sampling. We used the official implementation provided here: \href{https://github.com/intuitive-robots/beso}{https://github.com/intuitive-robots/beso}.

\section{Task Descriptions}
\label{app:envs}

We consider 10 different goal-conditioned tasks with sparse and binary rewards. The state, action, and goal spaces are continuous, and the maximum length of each episode is set as 50.
We use the offline benchmark introduced in \citet{yang2021rethinking}. The relatively easier tasks (PointReach, PointRooms, Reacher, SawyerReach, SawyerDoor, and FetchReach) have 2000 trajectories each ($1\times 10^5$ transitions); and the harder tasks (FetchPush, FetchPick, FetchSlide, and HandReach) have 40000 trajectories ($2\times 10^6$ transitions). 

The benchmark consists of two settings `expert' and `random'. The `expert' dataset consists of trajectories collected by a policy trained using online DDQPG+HER with added Gaussian noise ($\sigma=0.2$) to increase diversity, while the `random' dataset consists of trajectories collected by sampling random actions.  

\begin{figure}[h]
  \centering
  \begin{subfigure}[b]{0.19\textwidth}
      \centering
      \includegraphics[width=\linewidth]{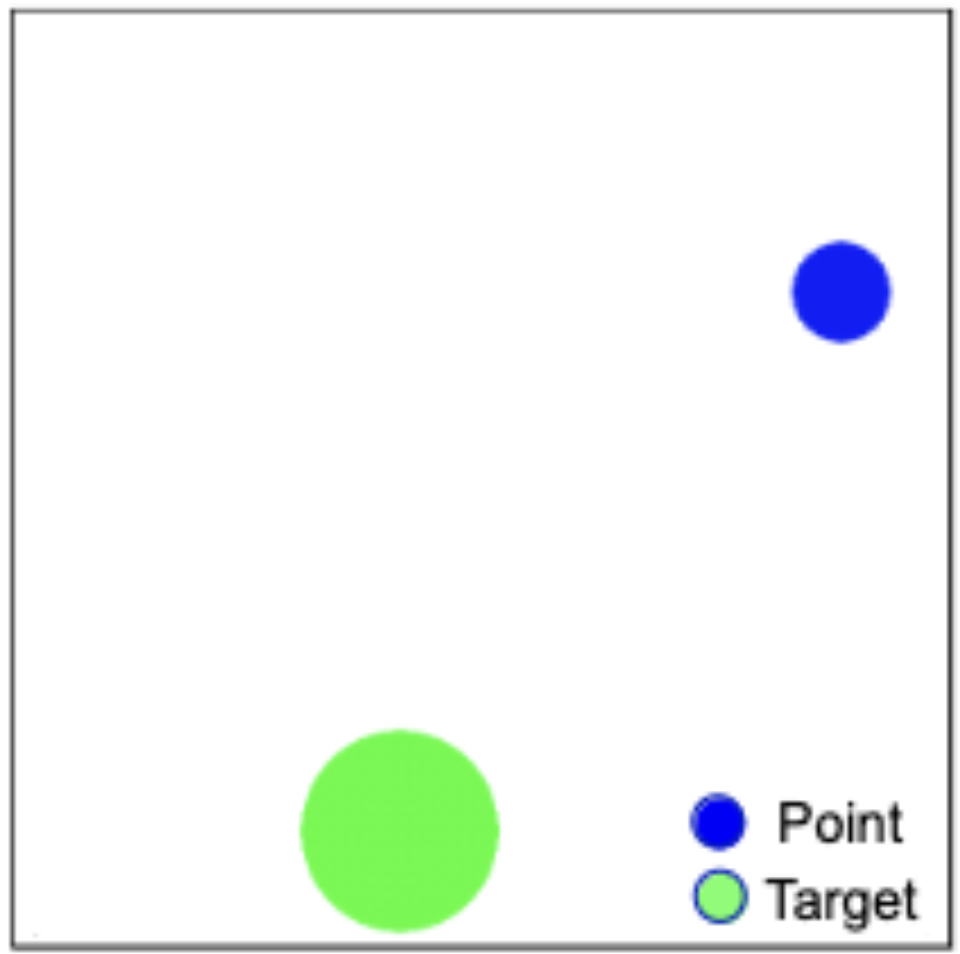}
  \end{subfigure}
  \begin{subfigure}[b]{0.19\textwidth}
      \centering
      \includegraphics[width=\textwidth]{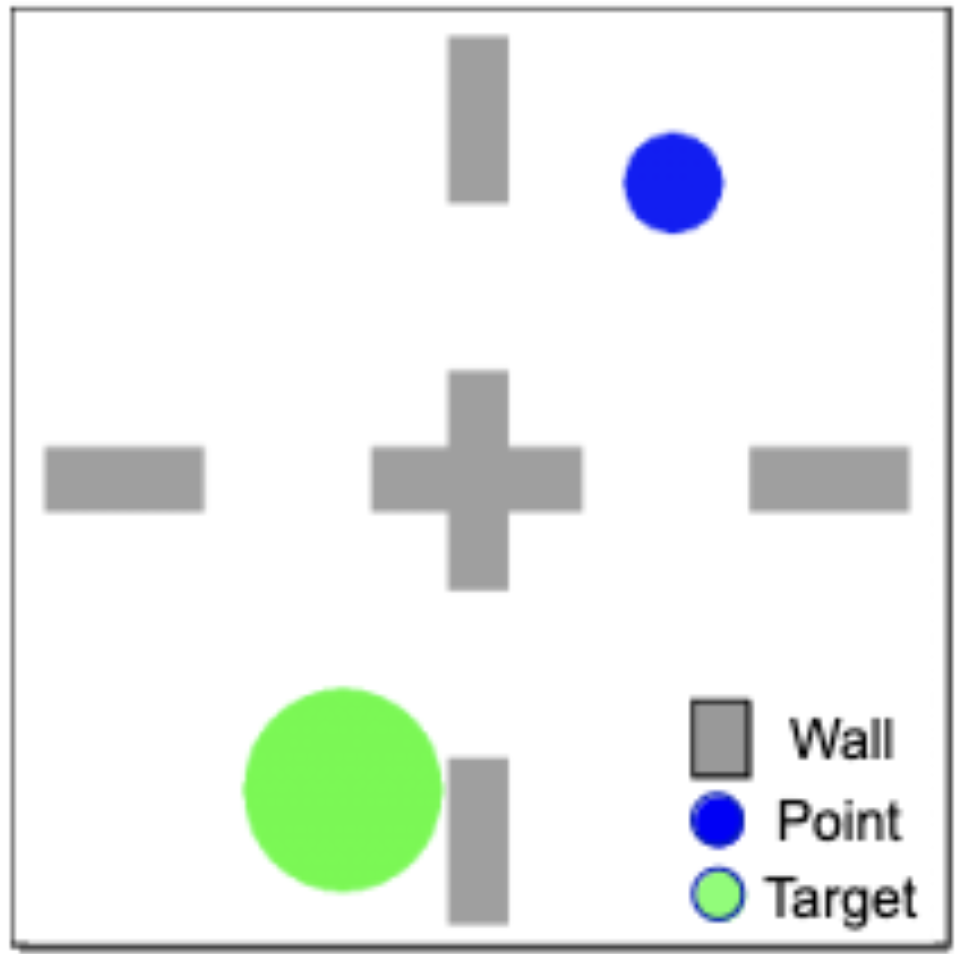}
  \end{subfigure}
  \begin{subfigure}[b]{0.19\textwidth}
      \centering
      \includegraphics[width=\textwidth]{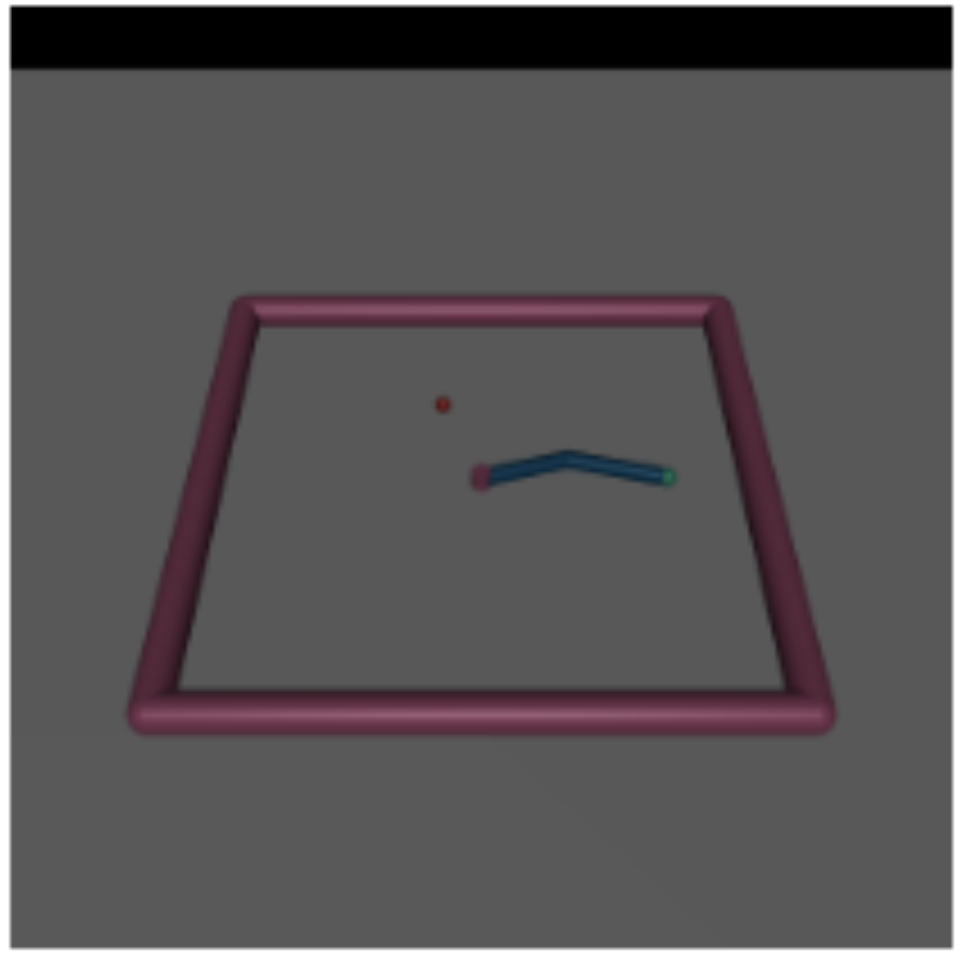}
  \end{subfigure}
  \begin{subfigure}[b]{0.19\textwidth}
      \centering
      \includegraphics[width=\textwidth]{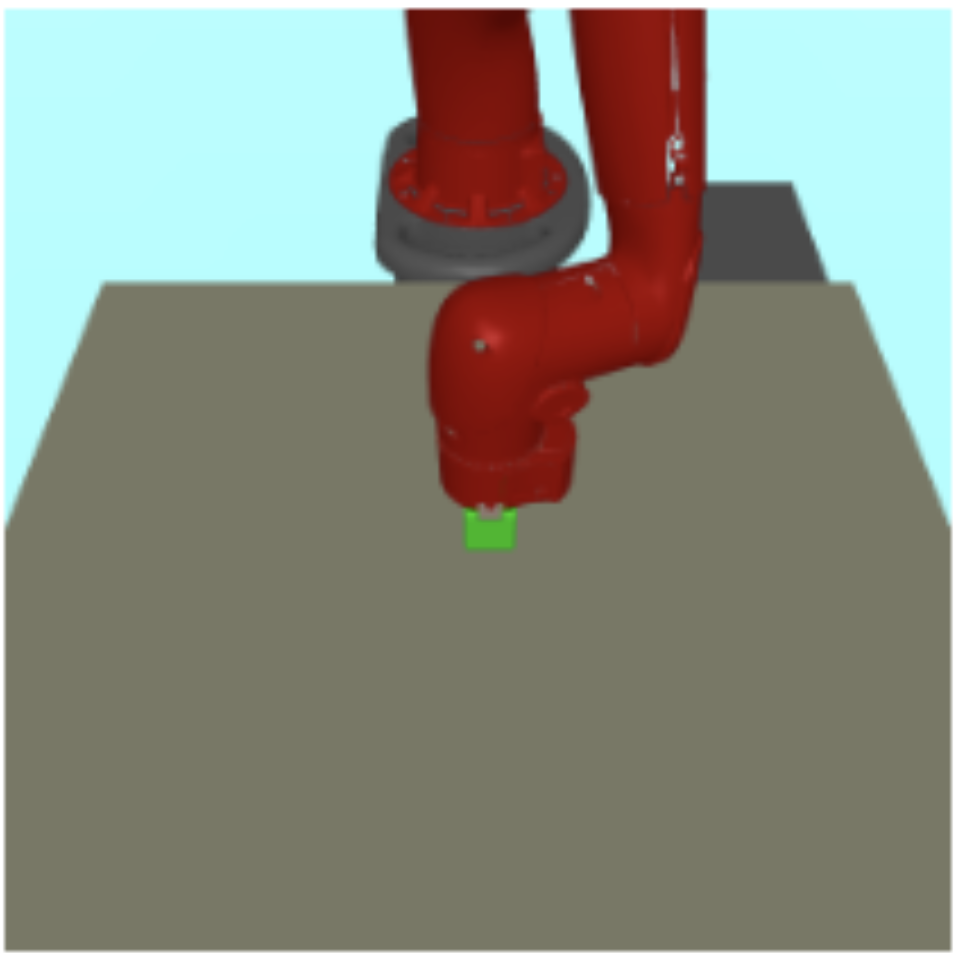}
  \end{subfigure}
  \begin{subfigure}[b]{0.19\textwidth}
      \centering
      \includegraphics[width=\textwidth]{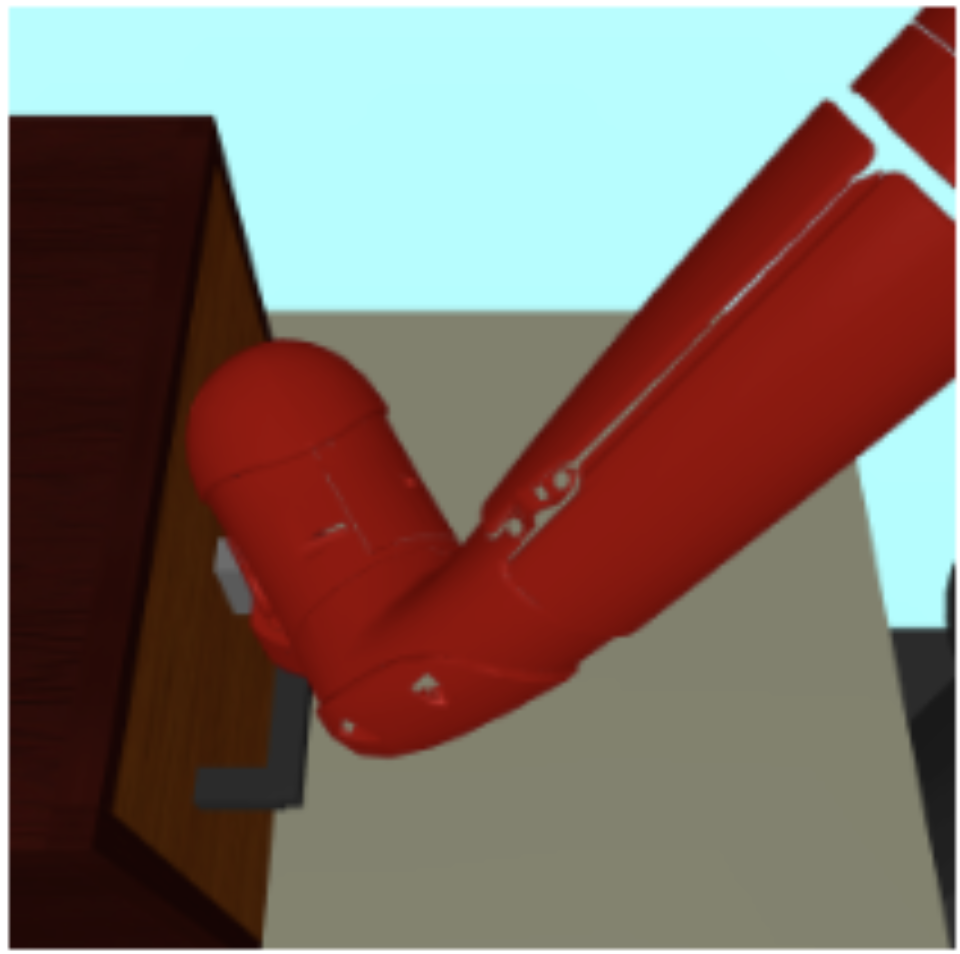}
  \end{subfigure}
  \newline
  \begin{subfigure}[b]{0.19\textwidth}
      \centering
      \includegraphics[width=\textwidth]{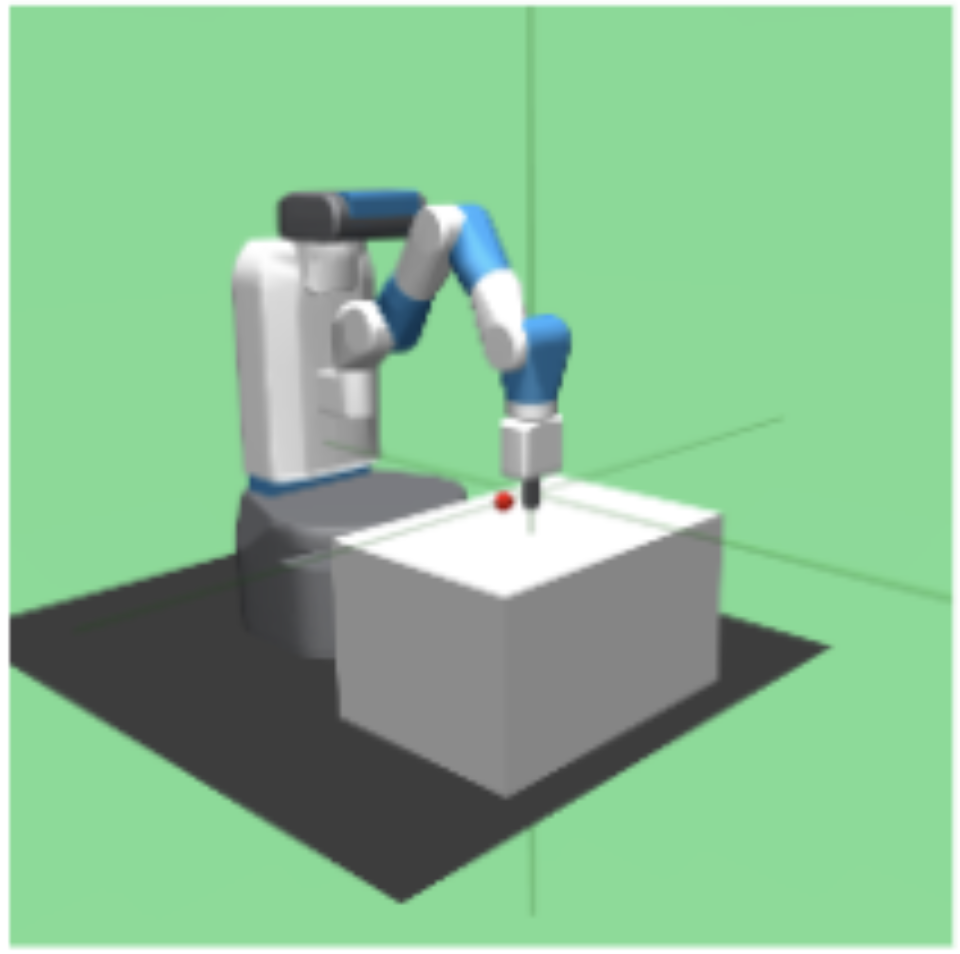}
  \end{subfigure}
  \begin{subfigure}[b]{0.19\textwidth}
      \centering
      \includegraphics[width=\textwidth]{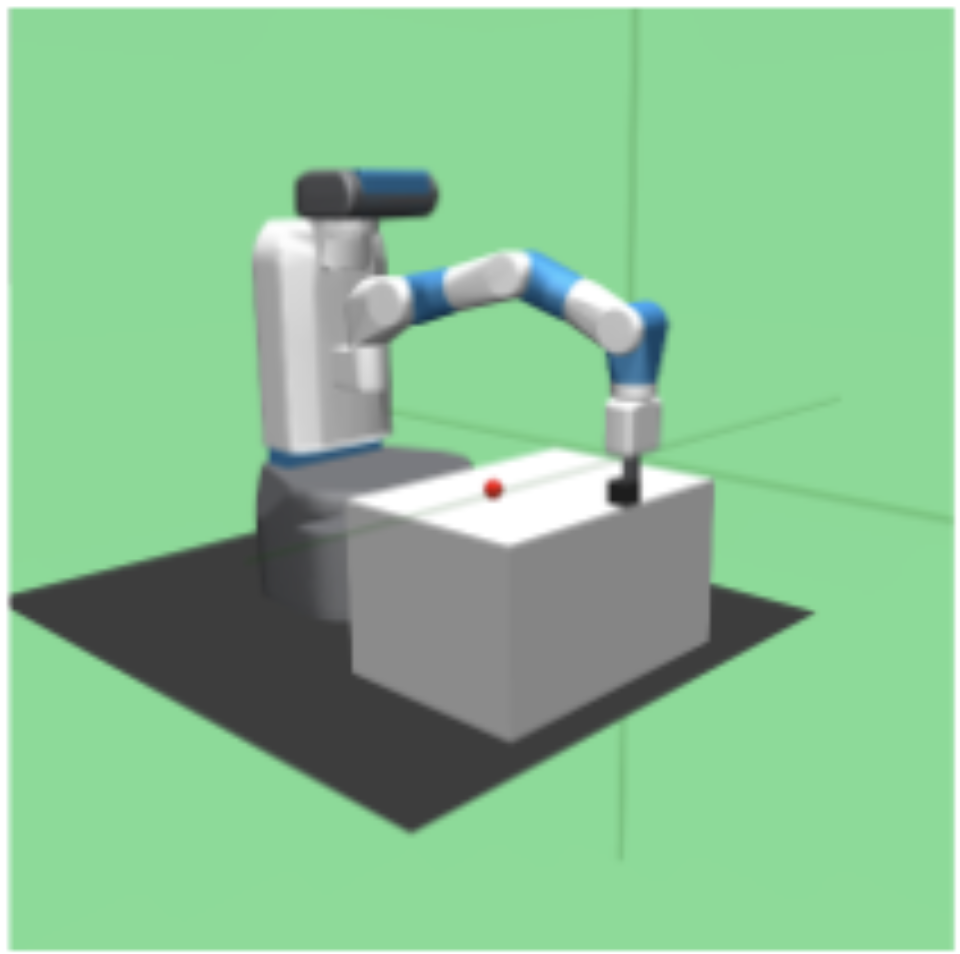}
  \end{subfigure}
  \begin{subfigure}[b]{0.19\textwidth}
      \centering
      \includegraphics[width=\textwidth]{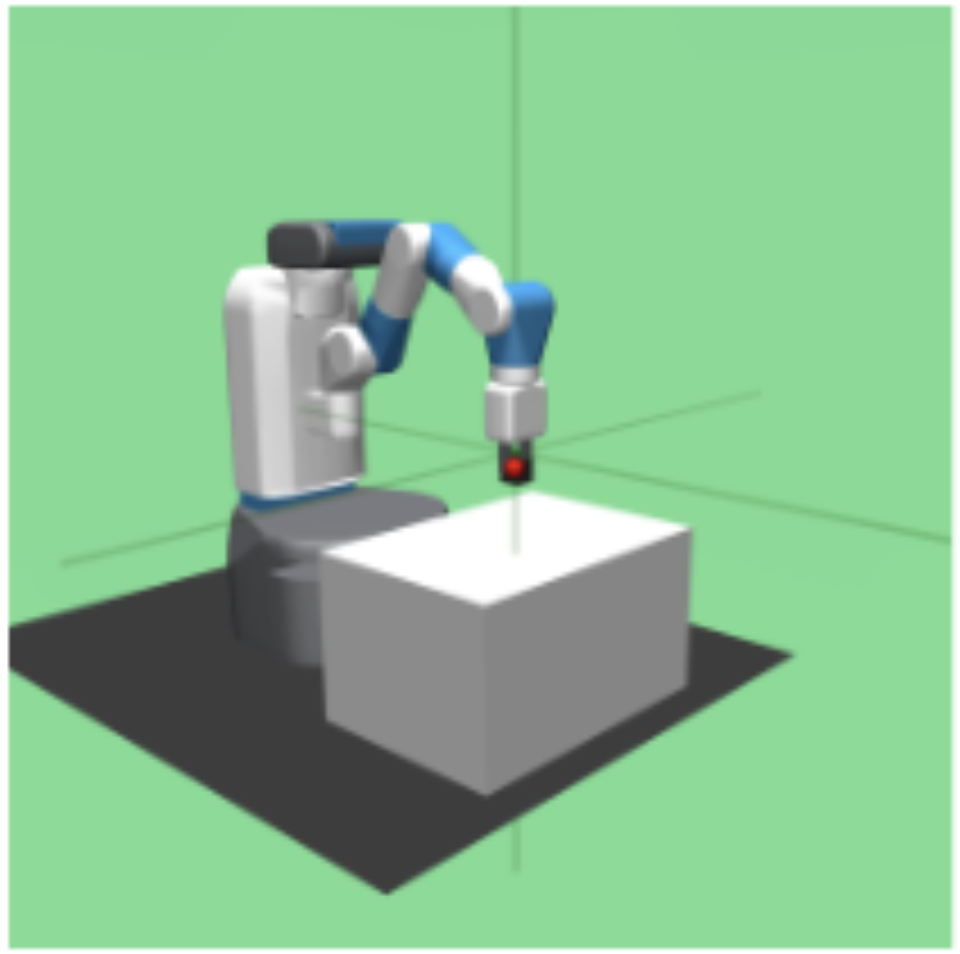}
  \end{subfigure}
  \begin{subfigure}[b]{0.19\textwidth}
      \centering
      \includegraphics[width=\textwidth]{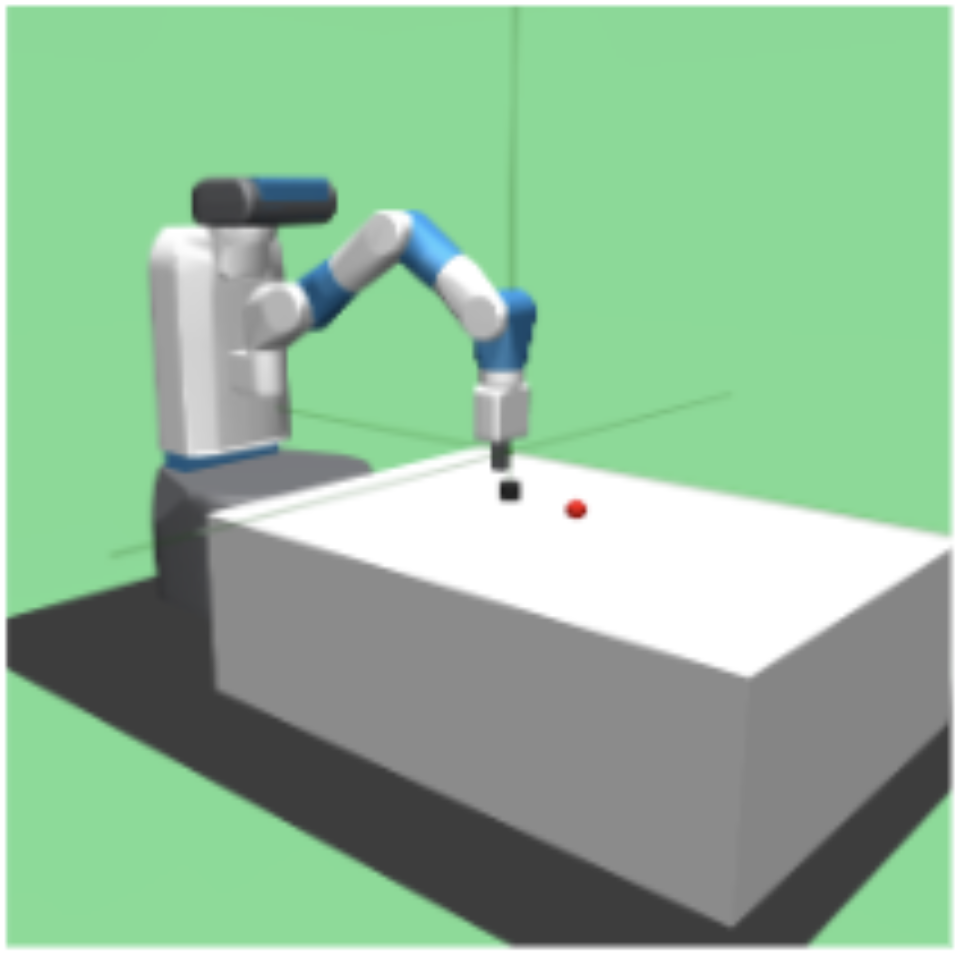}
  \end{subfigure}
  \begin{subfigure}[b]{0.19\textwidth}
      \centering
      \includegraphics[width=\textwidth]{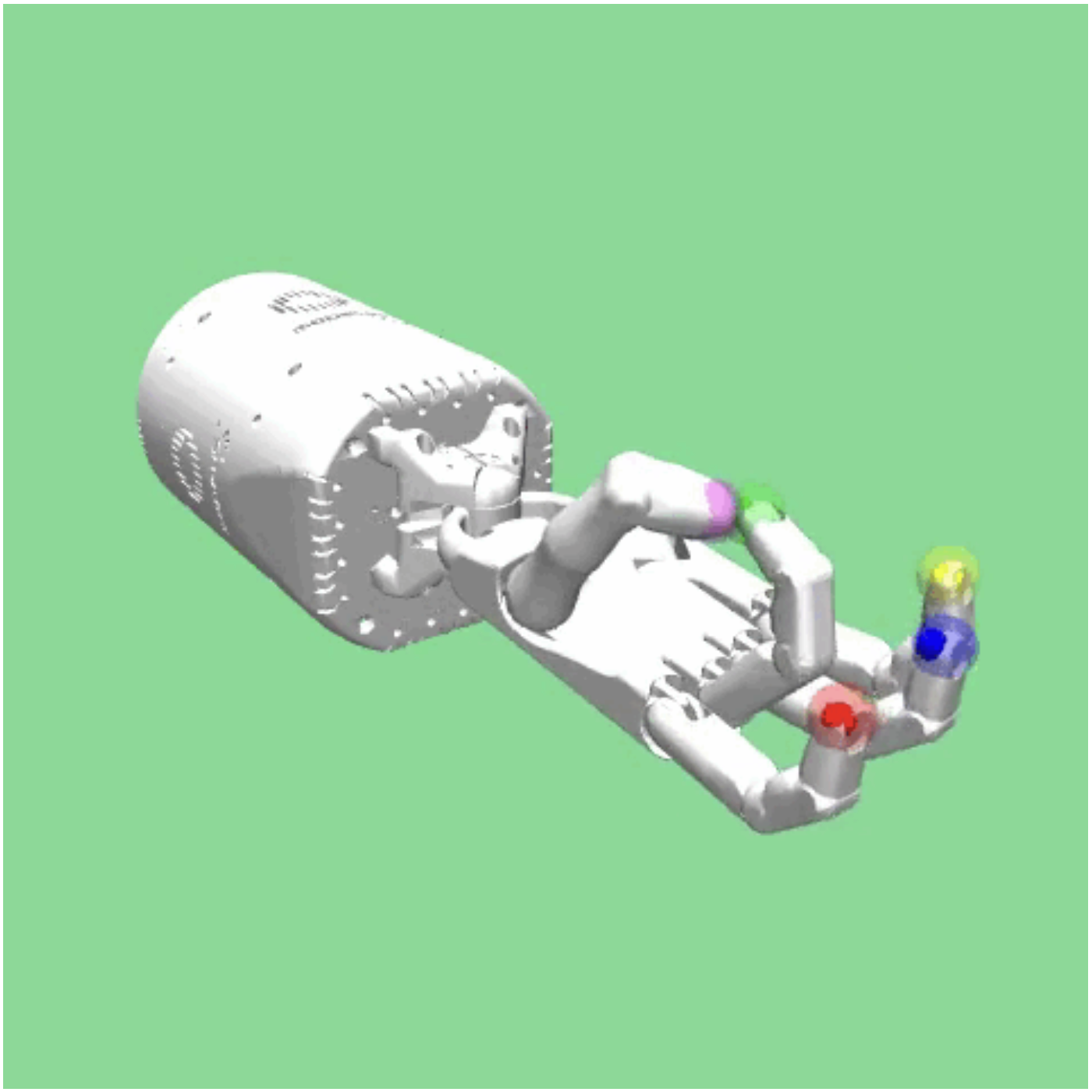}
  \end{subfigure}
  \vspace{-0em}
  \caption{\small{Goal-conditioned tasks from left to right, top to bottom: PointReach, PointRooms, Reacher, SawyerReach, SawyerDoor, FetchReach, FetchPush, FetchPick, FetchSlide, and HandReach.}}
  \vspace{-0.25em}
  \label{fig:tasks}
\end{figure}

\paragraph{PointReach.} The environment is adapted from \emph{multiworld}\footnote{\href{https://github.com/vitchyr/multiworld}{https://github.com/vitchyr/multiworld}}. The blue point represents the agent which is tasked with reaching the green circle representing the goal. The state space is two dimensional representing the $(x,y)$ coordinates of the blue point, where $(x,y) \in [-5,5] \times [-5,5]$. The actions space is also two dimensional representing the displacement in $x$ and $y$ directions, $a \in [-1,1]\times [-1,1]$. The goal space is the same as the state space, $\phi(s) = s$. The initial position of the agent and the goal are randomly initialized. Success is defined if the agent reaches within a certain radium of the goal. The reward function is defined as,
\begin{align*}
    r(s_{XY}, a, g_{XY}) = \mathbb{I}[\|s_{XY} - g_{XY}\|_2^2 \leq \epsilon]
\end{align*}
where the tolerance is $\epsilon=1$.

\paragraph{PointRooms.} The environment is a variation of PointReach environment. The task is again for the blue dot representing the agent to reach the green circle, however, there are vertical and horizontal walls forming four rooms, which make navigation more challenging. The reward function, and the state, action, and goal spaces are the same as in PointReach. 

\paragraph{Reacher.} The environment is included in Gymnasium\footnote{\href{https://github.com/Farama-Foundation/Gymnasium}{https://github.com/Farama-Foundation/Gymnasium}}. Reacher is a two-jointed robot arm tasked with moving the robot’s end effector close to a target that is spawned at a random position. The state space is 11-dimensional representing the angles, positions and velocities of the joints. The goals are $(x,y)$ coordinates of the target, and $\phi(s) = s[4:6]$. The two-dimensional actions represent the torque applied at each joint. The reward function is defined as,
\begin{align*}
    r(s_{XY}, a, g_{XY}) = \mathbb{I}[\|s_{XY} - g_{XY}\|_2^2 \leq \epsilon]
\end{align*}
where the tolerance is $\epsilon=0.05$.

\paragraph{SawyerReach.} The environment is adapted from \emph{multiworld}. The Sawyer robot is tasked with reaching a target position using its end effector. The state space is 3-dimensional representing the $(x,y,z)$ coordinates of the end effector, and the goal space is also 3-dimensional representing the $(x,y,z)$ coordinates of the target position, $\phi(s) = s$. The 3-dimensional actions describe the coordinates of the next position of the end effector. The reward function is defined as,
\begin{align*}
    r(s_{XYZ}, a, g_{XYZ}) = \mathbb{I}[\|s_{XYZ} - g_{XYZ}\|_2^2 \leq \epsilon]
\end{align*}
where the tolerance is $\epsilon=0.06$.

\paragraph{SawyerDoor.} The environment is adapted from \emph{multiworld}. The Sawyer robot is tasked with opening a door to a specified angle. The 4-dimensional state space represents the coordinates of the end effector of the robot and the angle of the door. The action space is the 3-dimensional next position of the end effector. The goal is the desired angle of the door, $\phi(s) = s[-1]$, which is between $[0,0.83]$ radians. The reward function is defined as,
\begin{align*}
    r(s, a, g) = \mathbb{I}[|\phi(s) - g| \leq \epsilon]
\end{align*}
where the tolerance is $\epsilon=0.06$.

\paragraph{FetchReach. } The environment is included in Gymnasium-Robotics\footnote{\href{https://github.com/Farama-Foundation/Gymnasium-Robotics}{https://github.com/Farama-Foundation/Gymnasium-Robotics}}. It consists of a 7-DoF robotic arm, with a two-fingered parallel gripper attached to it. The task is to reach a target location which is specified as a 3-dimensional goal representing the $(x,y,z)$ coordinates of the target location. The states are 10-dimensional represent the kinematic information of the end effector, including the positions and velocities of the end effector and the gripper joint displacement. The actions are 4-dimensional which describes the displacement of the end effector, and the last dimension which represents the gripper opening/closing is not used for this task. The state-to-goal mapping is $\phi(s) = s[0:3]$. The reward function is defined as,
\begin{align*}
    r(s, a, g_{XYZ}) = \mathbb{I}[\|\phi(s) - g_{XYZ}\|_2^2 \leq \epsilon]
\end{align*}
where the tolerance is $\epsilon=0.05$.

\paragraph{FetchPush.} The environment is included in Gymnasium-Robotics. The 7-DoF robotic arm from FetchReach is tasked with pushing a block to a target location. The state space is 25-
dimensional, including the gripper’s position, linear velocities, and the box’s position, rotation, linear and angular velocities. The 4-dimensional state space describes the displacement of the end effector and the gripper opening/closing. The goal is defined as the target $(x,y,z)$ position of the block, and the mapping is $\phi(s) = s[3:6]$. In this task, the block is always on top of the table, hence the block $z$ coordinate is always fixed. The reward function is defined as,
\begin{align*}
    r(s, a, g_{XYZ}) = \mathbb{I}[\|\phi(s) - g_{XYZ}\|_2^2 \leq \epsilon]
\end{align*}
where the tolerance is $\epsilon=0.05$.

\paragraph{FetchPick.} The environment is included in Gymnasium-Robotics. The 7-DoF robotic arm from FetchReach is tasked with picking up a block and taking it to a target location specified as $(x,y,z)$ coordinates. The target $z$ coordinate of the block is not fixed and may be in the air above the table, requiring the robotic arm to pick up the block using the gripper. The state space, action space, goal space, state-to-goal mapping, and reward function are the same as FetchPush.

\paragraph{FetchSlide.} The environment is included in Gymnasium-Robotics. The 7-DoF robotic arm from FetchReach is tasked with moving a block to a target position specified as $(x,y,z)$ coordinates. The block is always on top of the table, hence the $z$ coordinate of the block is always fixed. However, the $(x,y)$ coordinates of the target position are out of reach of the robotic arm, hence it must hit the block with the appropriate amount of force for it to slide and then stop at the goal position. The state space, action space, goal space, state-to-goal mapping, and reward function are the same as FetchPush.

\paragraph{HandReach.} The environment is included in Gymnasium-Robotics. A 24-DoF anthropomorphic hand is tasked with manipulating its fingers to reach a target configuration. The state space is 63-dimensional comprising two 24-dimensional vectors describing the positions and velocities of the joints, and five 3-dimensional vectors describing the  $(x,y,z)$ positions of each fingertip. The 20-dimensional actions describe the absolute angular positions of the actuated joints. The goals are specified as 15-dimensional vectors, describing the $(x,y,z)$ coordinates of the fingertips with $\phi(s) = s[-15:]$. The reward function is defined as,
\begin{align*}
    r(s, a, g) = \mathbb{I}[\|\phi(s) - g\|_2^2 \leq \epsilon]
\end{align*}
where the tolerance is $\epsilon=0.01$.

\section{Forward vs. Backward view}
\label{app:forward_backward}

We implement forward-view variants of Merlin-P and Merlin-NP to explicitly compare the benefits of the backward view proposed in our work. We compare two variants:
\begin{itemize}
    \item MB (model-based) samples an initial state and performs model-based rollout to provide additional trajectories to train the policy. The network architecture and hyperparameters are the same as Merlin-P, except for the forward view rollouts.
    \item TS (trajectory stitching) samples an initial state and performs trajectory stitching by sequentially searching for nearest-neighbors in the latent state space, forward in time.
\end{itemize}

We post experimental results on all tasks for the expert and random settings in \Cref{table:returns3}. The results show that the diffusion-inspired backward view of Merlin-P and Merlin-NP, where we have control over the goal state distribution, performs better on a majority of tasks.

\begin{table*}[h]
\captionof{table}{\small{Discounted returns for state-space input, averaged over 10 seeds.}}
\vspace{-1em}
\begin{center}
\begin{adjustbox}{width=0.7\textwidth}
\begin{tabular}{|ll|r|rr|rr|}
\hline \\[-2.35ex]
\multicolumn{2}{c|}{\textbf{Task Name}} & \multicolumn{1}{c|}{\textbf{Merlin}} & \multicolumn{1}{c}{\textbf{Merlin-P}} & \multicolumn{1}{c|}{\textbf{Merlin-MB}} & \multicolumn{1}{c}{\textbf{Merlin-NP}} & \multicolumn{1}{c|}{\textbf{Merlin-TS}}  \\
\hline \rule{0pt}{3ex}
\multirow{10}{*}{\rotatebox[origin=c]{90}{\textbf{Expert}}}
& PointReach & 29.26\scriptsize{$\pm$0.04} & \textbf{29.34}\scriptsize{$\pm$0.15} & 29.30\scriptsize{$\pm$0.14} & \textbf{29.34}\scriptsize{$\pm$0.05} & 29.24\scriptsize{$\pm$0.10} \\
& PointRooms & \textbf{25.38}\scriptsize{$\pm$0.37} & 25.25\scriptsize{$\pm$0.27} & 25.22\scriptsize{$\pm$0.28} & \textbf{25.63}\scriptsize{$\pm$0.32} & 25.28\scriptsize{$\pm$0.30} \\
& Reacher & 22.75\scriptsize{$\pm$0.59} & \textbf{24.25}\scriptsize{$\pm$0.47} & 23.89\scriptsize{$\pm$0.58} & \textbf{24.97}\scriptsize{$\pm$0.54} & 24.08\scriptsize{$\pm$0.51} \\
& SawyerReach & 26.89\scriptsize{$\pm$0.07} & \textbf{26.92}\scriptsize{$\pm$0.09} & 26.67\scriptsize{$\pm$0.09} & \textbf{27.35}\scriptsize{$\pm$0.06} & 26.69\scriptsize{$\pm$0.05} \\
& SawyerDoor & \textbf{26.18}\scriptsize{$\pm$2.19} & 25.85\scriptsize{$\pm$0.97} & 25.89\scriptsize{$\pm$1.26} & \textbf{26.15}\scriptsize{$\pm$2.08} & 26.06\scriptsize{$\pm$1.58} \\
& FetchReach & 30.29\scriptsize{$\pm$0.03} & \textbf{30.34}\scriptsize{$\pm$0.02} & 30.32\scriptsize{$\pm$0.03} & \textbf{30.42}\scriptsize{$\pm$0.04} & 30.29\scriptsize{$\pm$0.03} \\
& FetchPush & 19.91\scriptsize{$\pm$1.20} & \textbf{22.13}\scriptsize{$\pm$1.41} & 21.08\scriptsize{$\pm$1.34} & \textbf{21.58}\scriptsize{$\pm$1.63} & 20.99\scriptsize{$\pm$1.58} \\
& FetchPick & 19.66\scriptsize{$\pm$0.78} & \textbf{21.78}\scriptsize{$\pm$1.01} & \textbf{21.14}\scriptsize{$\pm$1.05} & 20.41\scriptsize{$\pm$0.92} & 20.50\scriptsize{$\pm$1.09} \\
& FetchSlide & 4.19\scriptsize{$\pm$1.89} & \textbf{4.98}\scriptsize{$\pm$1.46} & 4.11\scriptsize{$\pm$1.52} & \textbf{5.19}\scriptsize{$\pm$2.02} & 4.55\scriptsize{$\pm$1.84} \\
& HandReach & 22.11\scriptsize{$\pm$0.55} & \textbf{23.44}\scriptsize{$\pm$0.62} & 21.97\scriptsize{$\pm$0.46} & \textbf{24.93}\scriptsize{$\pm$0.49} & 22.81\scriptsize{$\pm$0.55} \\
\hline \rule{0pt}{3ex}%
\multirow{10}{*}{\rotatebox[origin=c]{90}{\textbf{Random}}}
& PointReach & 29.26\scriptsize{$\pm$0.04} & \textbf{29.36}\scriptsize{$\pm$0.08} & 29.29\scriptsize{$\pm$0.12} & \textbf{29.31}\scriptsize{$\pm$0.04} & 29.26\scriptsize{$\pm$0.11} \\
& PointRooms & 24.80\scriptsize{$\pm$0.36} & \textbf{25.17}\scriptsize{$\pm$0.19} & 25.01\scriptsize{$\pm$0.18} & \textbf{25.16}\scriptsize{$\pm$0.59} & 25.03\scriptsize{$\pm$0.41} \\
& Reacher & 21.09\scriptsize{$\pm$0.65} & \textbf{24.49}\scriptsize{$\pm$0.48} & \textbf{23.72}\scriptsize{$\pm$0.44} & 22.24\scriptsize{$\pm$0.54} & 21.38\scriptsize{$\pm$0.36} \\
& SawyerReach & 26.70\scriptsize{$\pm$0.14} & 26.78\scriptsize{$\pm$0.12} & 26.78\scriptsize{$\pm$0.10} & \textbf{27.07}\scriptsize{$\pm$0.07} & \textbf{26.89}\scriptsize{$\pm$0.06} \\
& SawyerDoor & 19.05\scriptsize{$\pm$0.66} & 20.37\scriptsize{$\pm$1.18} & 20.04\scriptsize{$\pm$0.93} & \textbf{21.69}\scriptsize{$\pm$2.36} & \textbf{21.04}\scriptsize{$\pm$1.84} \\
& FetchReach & \textbf{30.42}\scriptsize{$\pm$0.04} & \textbf{30.44}\scriptsize{$\pm$0.02} & 30.30\scriptsize{$\pm$0.03} & \textbf{30.42}\scriptsize{$\pm$0.04} & 30.27\scriptsize{$\pm$0.03} \\
& FetchPush & 5.21\scriptsize{$\pm$0.43} & 6.83\scriptsize{$\pm$0.32} & 6.02\scriptsize{$\pm$1.28} & \textbf{7.22}\scriptsize{$\pm$0.35} & \textbf{6.94}\scriptsize{$\pm$0.56} \\
& FetchPick & 3.75\scriptsize{$\pm$0.18} & \textbf{4.22}\scriptsize{$\pm$0.16} & \textbf{3.89}\scriptsize{$\pm$0.15} & 4.36\scriptsize{$\pm$0.19} & 4.01\scriptsize{$\pm$0.17} \\
& FetchSlide & 2.67\scriptsize{$\pm$0.35} & \textbf{2.98}\scriptsize{$\pm$0.21} & 2.59\scriptsize{$\pm$0.18} & \textbf{3.15}\scriptsize{$\pm$0.14} & 2.85\scriptsize{$\pm$0.15} \\
& HandReach & 14.89\scriptsize{$\pm$2.54} & 18.24\scriptsize{$\pm$2.18} & 15.37\scriptsize{$\pm$1.66} & \textbf{20.06}\scriptsize{$\pm$3.06} & \textbf{18.28}\scriptsize{$\pm$1.29} \\
\hline
\end{tabular}
\end{adjustbox}
\end{center}
\label{table:returns3}%
\end{table*}

\clearpage
\section{GCSL with Trajectory Stitching}
\label{app:traj}

We apply a modified version of the nearest-neighbor trajectory stitching operation to GCSL and report the performance in \Cref{table:ts1} and \Cref{table:ts2}, averaged over 10 seeds. The technique described in \Cref{sec:traj} applies to reverse trajectories, for GCSL we construct forward trajectories by adding the state-action pair succeeding the nearest neighbors. We observe that this technique improves performance for most tasks, demonstrating it as a general-purpose data augmentation technique for offline GCRL.

\begin{figure}[h]
\vspace{-2em}
\begin{minipage}{0.49\textwidth}
\begin{table}[H]
\caption{\small{Discounted returns.}}
\vspace{-1em}
\begin{center}
\begin{adjustbox}{width=\textwidth}
\begin{tabular}{|ll|rr|rr|}
\hline \\[-2.35ex]
\rule{0pt}{3ex}
& \textbf{Task Name} & \textbf{Merlin} & \textbf{Merlin-NP} & \textbf{GCSL} & \textbf{GCSL+TS} \\
\hline \rule{0pt}{3ex}
\multirow{10}{*}{\rotatebox[origin=c]{90}{\textbf{Expert}}}
& PointReach        & {29.26}\scriptsize{$\pm$0.04} & {29.34}\scriptsize{$\pm$0.05} & 22.85\scriptsize{$\pm$1.26} & 23.22\scriptsize{$\pm$1.71} \\
& PointRooms        & 25.38\scriptsize{$\pm$0.37} & {25.63}\scriptsize{$\pm$0.32} & 18.28\scriptsize{$\pm$2.29}  & 19.87\scriptsize{$\pm$1.55} \\
& Reacher           & 22.75\scriptsize{$\pm$0.59} & {24.97}\scriptsize{$\pm$0.54}  & 20.05\scriptsize{$\pm$1.37} & 22.12\scriptsize{$\pm$1.16} \\
& SawyerReach       &{26.89}\scriptsize{$\pm$0.07} &  {27.35}\scriptsize{$\pm$0.06} & 19.20\scriptsize{$\pm$1.79} & 20.88\scriptsize{$\pm$1.60} \\
& SawyerDoor        &{26.18}\scriptsize{$\pm$2.19} & 26.15\scriptsize{$\pm$2.08} & 20.12\scriptsize{$\pm$1.33} & 20.61\scriptsize{$\pm$1.26} \\
& FetchReach        &{30.29}\scriptsize{$\pm$0.03} & {30.42}\scriptsize{$\pm$0.04} & 23.68\scriptsize{$\pm$1.07} & 23.59\scriptsize{$\pm$1.32} \\
& FetchPush         & 19.91\scriptsize{$\pm$1.20} & 21.58\scriptsize{$\pm$1.63} & 17.58\scriptsize{$\pm$1.47} & 19.15\scriptsize{$\pm$1.29} \\
& FetchPick         & 19.66\scriptsize{$\pm$0.78} & {20.41}\scriptsize{$\pm$0.92} & 12.95\scriptsize{$\pm$1.90} & 13.85\scriptsize{$\pm$1.66} \\
& FetchSlide        & 4.19\scriptsize{$\pm$1.89} & {5.19}\scriptsize{$\pm$2.02}  & 1.67\scriptsize{$\pm$1.41} &  2.11\scriptsize{$\pm$1.46} \\
& HandReach         & {22.11}\scriptsize{$\pm$0.55} & {24.93}\scriptsize{$\pm$0.49} & 0.15\scriptsize{$\pm$0.11} & 0.16\scriptsize{$\pm$0.13} \\
\hline \rule{0pt}{3ex}
\multirow{10}{*}{\rotatebox[origin=c]{90}{\textbf{Random}}}
& PointReach        &{29.26}\scriptsize{$\pm$0.04} & {29.31}\scriptsize{$\pm$0.04}  & 17.74\scriptsize{$\pm$1.84}  & 20.01\scriptsize{$\pm$1.63} \\
& PointRooms        &{24.80}\scriptsize{$\pm$0.36} & {25.16}\scriptsize{$\pm$0.59} & 14.69\scriptsize{$\pm$2.51}  & 16.05\scriptsize{$\pm$1.97} \\
& Reacher           &21.09\scriptsize{$\pm$0.65} & 22.24\scriptsize{$\pm$0.54} & 10.62\scriptsize{$\pm$2.30} &  12.89\scriptsize{$\pm$2.34} \\
& SawyerReach       &{26.70}\scriptsize{$\pm$0.14} & {27.07}\scriptsize{$\pm$0.07} & 8.78\scriptsize{$\pm$2.59}  & 9.12\scriptsize{$\pm$2.26} \\
& SawyerDoor        & 19.05\scriptsize{$\pm$0.66} & {21.69}\scriptsize{$\pm$2.36} & 12.47\scriptsize{$\pm$3.08} & 13.64\scriptsize{$\pm$2.68} \\
& FetchReach        &{30.42}\scriptsize{$\pm$0.04} &  {30.42}\scriptsize{$\pm$0.04}  & 18.96\scriptsize{$\pm$1.77} & 19.58\scriptsize{$\pm$1.72} \\
& FetchPush         & 5.21\scriptsize{$\pm$0.43} & {7.22}\scriptsize{$\pm$0.35} & 4.22\scriptsize{$\pm$2.19} & 5.21\scriptsize{$\pm$1.98} \\
& FetchPick         & 3.75\scriptsize{$\pm$0.18} & {4.36}\scriptsize{$\pm$0.19} & 0.81\scriptsize{$\pm$0.82} & 0.95\scriptsize{$\pm$0.90} \\
& FetchSlide        & {2.67}\scriptsize{$\pm$0.35} & {3.15}\scriptsize{$\pm$0.14} & 0.24\scriptsize{$\pm$0.27} & 0.31\scriptsize{$\pm$0.36} \\
& HandReach         & {14.89}\scriptsize{$\pm$2.54} & {20.06}\scriptsize{$\pm$3.06}  & 1.41\scriptsize{$\pm$0.51} & 2.06\scriptsize{$\pm$0.76} \\
\hline
\end{tabular}
\end{adjustbox}
\end{center}
\label{table:ts1}
\end{table}
\end{minipage}
\hfill
\begin{minipage}{0.49\textwidth}
\begin{table}[H]
\caption{\small{Success rates.}}
\vspace{-1em}
\begin{center}
\begin{adjustbox}{width=0.965\textwidth}
\begin{tabular}{|ll|rr|rr|}
\hline \\[-2.35ex]
\rule{0pt}{3ex}
& \textbf{Task Name} & \textbf{Merlin} & \textbf{Merlin-NP} & \textbf{GCSL} & \textbf{GCSL+TS} \\
\hline \rule{0pt}{3ex}
\multirow{10}{*}{\rotatebox[origin=c]{90}{\textbf{Expert}}}
& PointReach        & {1.00}\scriptsize{$\pm$0.00} & {1.00}\scriptsize{$\pm$0.00} & {1.00}\scriptsize{$\pm$0.00} & {1.00}\scriptsize{$\pm$0.00}\\
& PointRooms        & 0.91\scriptsize{$\pm$0.16} & {0.94}\scriptsize{$\pm$0.01} & 0.79\scriptsize{$\pm$0.60} & {0.80}\scriptsize{$\pm$0.62} \\
& Reacher           & {1.00}\scriptsize{$\pm$0.00} & {1.00}\scriptsize{$\pm$0.00} & {1.00}\scriptsize{$\pm$0.00} & {1.00}\scriptsize{$\pm$0.00} \\
& SawyerReach       & {1.00}\scriptsize{$\pm$0.00} & {1.00}\scriptsize{$\pm$0.00} & {1.00}\scriptsize{$\pm$0.00} & {1.00}\scriptsize{$\pm$0.00} \\
& SawyerDoor        & {0.95}\scriptsize{$\pm$0.08} & {0.94}\scriptsize{$\pm$0.11} & 0.84\scriptsize{$\pm$0.16} & {0.85}\scriptsize{$\pm$0.14} \\
& FetchReach        & {1.00}\scriptsize{$\pm$0.00} & {1.00}\scriptsize{$\pm$0.00}  & 0.98\scriptsize{$\pm$0.00} & {1.00}\scriptsize{$\pm$0.00} \\
& FetchPush         & 0.92\scriptsize{$\pm$0.05} & {0.96}\scriptsize{$\pm$0.04} & 0.88\scriptsize{$\pm$0.09} & {0.89}\scriptsize{$\pm$0.10} \\
& FetchPick         & {0.92}\scriptsize{$\pm$0.03} & {0.96}\scriptsize{$\pm$0.06}  & 0.64\scriptsize{$\pm$0.09} & {0.67}\scriptsize{$\pm$0.09}\\
& FetchSlide        & 0.32\scriptsize{$\pm$0.04} & 0.45\scriptsize{$\pm$0.07} & 0.22\scriptsize{$\pm$0.14} &{0.24}\scriptsize{$\pm$0.12} \\
& HandReach         & {0.78}\scriptsize{$\pm$0.04} & {0.85}\scriptsize{$\pm$0.02} &  0.03\scriptsize{$\pm$0.05} & {0.04}\scriptsize{$\pm$0.05} \\
\hline \rule{0pt}{3ex}  
\multirow{10}{*}{\rotatebox[origin=c]{90}{\textbf{Random}}}
& PointReach        & {1.00}\scriptsize{$\pm$0.00} & {1.00}\scriptsize{$\pm$0.00} & {1.00}\scriptsize{$\pm$0.00} & {1.00}\scriptsize{$\pm$0.00} \\
& PointRooms        & {0.89}\scriptsize{$\pm$0.02} & {0.92}\scriptsize{$\pm$0.02} & 0.77\scriptsize{$\pm$0.11}  &  {0.79}\scriptsize{$\pm$0.12} \\
& Reacher           & 0.98\scriptsize{$\pm$0.02} & {1.00}\scriptsize{$\pm$0.00} & 0.80\scriptsize{$\pm$0.06} & {0.84}\scriptsize{$\pm$0.07} \\
& SawyerReach       & {1.00}\scriptsize{$\pm$0.00} & {1.00}\scriptsize{$\pm$0.00} & 0.91\scriptsize{$\pm$0.09}  & {0.93}\scriptsize{$\pm$0.11} \\
& SawyerDoor        & {0.57}\scriptsize{$\pm$0.03} & {0.59}\scriptsize{$\pm$0.05}  & 0.44\scriptsize{$\pm$0.16} & {0.45}\scriptsize{$\pm$0.14} \\
& FetchReach        & {1.00}\scriptsize{$\pm$0.00} & {1.00}\scriptsize{$\pm$0.00}  & 0.96\scriptsize{$\pm$0.05} & {0.98}\scriptsize{$\pm$0.03} \\
& FetchPush         & 0.20\scriptsize{$\pm$0.06} & {0.24}\scriptsize{$\pm$0.09} & 0.20\scriptsize{$\pm$0.11} &  {0.22}\scriptsize{$\pm$0.10}\\
& FetchPick         & {0.12}\scriptsize{$\pm$0.01} & {0.18}\scriptsize{$\pm$0.01} & 0.06\scriptsize{$\pm$0.08} & {0.07}\scriptsize{$\pm$0.06}\\
& FetchSlide        & {0.11}\scriptsize{$\pm$0.02} & {0.20}\scriptsize{$\pm$0.04} & 0.06\scriptsize{$\pm$0.08} & {0.06}\scriptsize{$\pm$0.07}\\
& HandReach         & {0.49}\scriptsize{$\pm$0.05} & {0.62}\scriptsize{$\pm$0.07} & 0.04\scriptsize{$\pm$0.04} & {0.04}\scriptsize{$\pm$0.04}\\
\hline
\end{tabular}
\end{adjustbox}
\end{center}
\label{table:ts2}
\end{table}
\end{minipage}
\end{figure}

\section{Full Experimental Results}
\label{app:exp}

\begin{table*}[h]
\captionof{table}{\small{Discounted returns for state-space input, averaged over 10 seeds.}}
\vspace{-1em}
\begin{center}
\begin{adjustbox}{width=\textwidth}
\begin{tabular}{|ll|rrr|rrrr|rrr|}
\hline \\[-2.35ex]
\multicolumn{2}{c|}{\multirow{2}{*}{\textbf{Task Name}}} & \multicolumn{3}{c|}{\textbf{Ours}} & \multicolumn{4}{c|}{\textbf{Offline GCRL}} & \multicolumn{3}{c|}{\textbf{Diffusion-based}}\\
\rule{0pt}{2ex}
& & \multicolumn{1}{c}{\textbf{Merlin}} & \multicolumn{1}{c}{\textbf{Merlin-P}} & \multicolumn{1}{c|}{\textbf{Merlin-NP}} & \multicolumn{1}{c}{\textbf{GoFAR}} & \multicolumn{1}{c}{\textbf{WGCSL}} & \multicolumn{1}{c}{\textbf{GCSL}} & \multicolumn{1}{c|}{\textbf{AM}} & \multicolumn{1}{c}{\textbf{DD}} & \multicolumn{1}{c}{\textbf{g-DQL}} & \multicolumn{1}{c|}{\textbf{BESO}} \\
\hline \rule{0pt}{3ex}
\multirow{10}{*}{\rotatebox[origin=c]{90}{\textbf{Expert}}}
& PointReach        & 29.26\scriptsize{$\pm$0.04} & \textbf{29.34}\scriptsize{$\pm$0.15} & \textbf{29.34}\scriptsize{$\pm$0.05} & 27.18\scriptsize{$\pm$0.65} & 25.91\scriptsize{$\pm$0.87} & 22.85\scriptsize{$\pm$1.26} & 26.14\scriptsize{$\pm$1.11} & 15.03\scriptsize{$\pm$0.88} & 28.65\scriptsize{$\pm$0.44} & 29.10\scriptsize{$\pm$0.28} \\
& PointRooms        & 25.38\scriptsize{$\pm$0.37} & 25.25\scriptsize{$\pm$0.27} & \textbf{25.63}\scriptsize{$\pm$0.32} & 20.40\scriptsize{$\pm$1.00}  & 19.90\scriptsize{$\pm$0.99} & 18.28\scriptsize{$\pm$2.29}  & 23.24\scriptsize{$\pm$1.58} & 10.84\scriptsize{$\pm$2.67} & \textbf{27.53}\scriptsize{$\pm$0.57} & 24.13\scriptsize{$\pm$0.46} \\
& Reacher           & 22.75\scriptsize{$\pm$0.59} & \textbf{24.25}\scriptsize{$\pm$0.47} & \textbf{24.97}\scriptsize{$\pm$0.54}  & 22.51\scriptsize{$\pm$0.82} & 23.35\scriptsize{$\pm$0.64} & 20.05\scriptsize{$\pm$1.37} & 22.36\scriptsize{$\pm$1.03} & 14.39\scriptsize{$\pm$1.08} & 22.54\scriptsize{$\pm$1.42} & 22.78\scriptsize{$\pm$1.02} \\
& SawyerReach       & 26.89\scriptsize{$\pm$0.07} & \textbf{26.92}\scriptsize{$\pm$0.09} & \textbf{27.35}\scriptsize{$\pm$0.06} & 22.82\scriptsize{$\pm$1.15}& 22.07\scriptsize{$\pm$1.46} & 19.20\scriptsize{$\pm$1.79} & 23.56\scriptsize{$\pm$0.33}& 13.39\scriptsize{$\pm$0.75} & 24.17\scriptsize{$\pm$0.01}  & 26.44\scriptsize{$\pm$0.31} \\
& SawyerDoor        &\textbf{26.18}\scriptsize{$\pm$2.19} & 25.85\scriptsize{$\pm$0.97} & 26.15\scriptsize{$\pm$2.08} & 23.62\scriptsize{$\pm$0.35} & 23.92\scriptsize{$\pm$1.10} & 20.12\scriptsize{$\pm$1.33} & \textbf{26.39}\scriptsize{$\pm$0.42} & 12.85\scriptsize{$\pm$0.77} & 24.81\scriptsize{$\pm$0.38} & 23.14\scriptsize{$\pm$0.56} \\
& FetchReach        & 30.29\scriptsize{$\pm$0.03} & \textbf{30.34}\scriptsize{$\pm$0.02} & \textbf{30.42}\scriptsize{$\pm$0.04} & 29.21\scriptsize{$\pm$0.26} & 28.17\scriptsize{$\pm$0.38} & 23.68\scriptsize{$\pm$1.07} & 29.08\scriptsize{$\pm$0.12} & 11.55\scriptsize{$\pm$0.68} & 28.71\scriptsize{$\pm$0.15} & 29.18\scriptsize{$\pm$0.25} \\
& FetchPush         & 19.91\scriptsize{$\pm$1.20} & 22.13\scriptsize{$\pm$1.41} & 21.58\scriptsize{$\pm$1.63} & \textbf{22.41}\scriptsize{$\pm$1.69} & \textbf{22.22}\scriptsize{$\pm$1.51} & 17.58\scriptsize{$\pm$1.47} & 19.86\scriptsize{$\pm$3.16}& 9.49\scriptsize{$\pm$2.85} & 17.82\scriptsize{$\pm$0.55} & 14.52\scriptsize{$\pm$0.95} \\
& FetchPick         & 19.66\scriptsize{$\pm$0.78} & \textbf{21.78}\scriptsize{$\pm$1.01} & \textbf{20.41}\scriptsize{$\pm$0.92} & 19.79\scriptsize{$\pm$1.12} & 18.32\scriptsize{$\pm$1.56} & 12.95\scriptsize{$\pm$1.90} & 17.04\scriptsize{$\pm$3.81} & 8.76\scriptsize{$\pm$0.64} & 14.45\scriptsize{$\pm$0.61} & 18.56\scriptsize{$\pm$0.82} \\
& FetchSlide        & 4.19\scriptsize{$\pm$1.89} & 4.98\scriptsize{$\pm$1.46} & \textbf{5.19}\scriptsize{$\pm$2.02}  & 3.34\scriptsize{$\pm$1.01} & \textbf{5.17}\scriptsize{$\pm$3.17} & 1.67\scriptsize{$\pm$1.41} & 3.31\scriptsize{$\pm$1.46} & 1.21\scriptsize{$\pm$0.59} & 0.98\scriptsize{$\pm$0.58} & 3.40\scriptsize{$\pm$0.80} \\
& HandReach         & 22.11\scriptsize{$\pm$0.55} & \textbf{23.44}\scriptsize{$\pm$0.62} & \textbf{24.93}\scriptsize{$\pm$0.49} & 15.39\scriptsize{$\pm$6.37} & 18.05\scriptsize{$\pm$5.12} & 0.15\scriptsize{$\pm$0.11} & 0.00\scriptsize{$\pm$0.00} & 0.00\scriptsize{$\pm$0.00} & 0.00\scriptsize{$\pm$0.00} & 15.44\scriptsize{$\pm$0.24} \\
\hline \rule{0pt}{2.5ex}%
& Average Rank & \multicolumn{1}{c}{3.5} & \multicolumn{1}{c}{\textbf{2.4}} & \multicolumn{1}{c|}{\textbf{1.7}} & \multicolumn{1}{c}{5.4} & \multicolumn{1}{c}{5.5} & \multicolumn{1}{c}{8.6} & \multicolumn{1}{c|}{6.2} & \multicolumn{1}{c}{9.7} & \multicolumn{1}{c}{6.2} & \multicolumn{1}{c|}{5.4} \\
\hline \rule{0pt}{3ex}%
\multirow{10}{*}{\rotatebox[origin=c]{90}{\textbf{Random}}}
& PointReach        & 29.26\scriptsize{$\pm$0.04} & \textbf{29.36}\scriptsize{$\pm$0.08} & \textbf{29.31}\scriptsize{$\pm$0.04} & 23.96\scriptsize{$\pm$0.93} & 25.76\scriptsize{$\pm$0.96} & 17.74\scriptsize{$\pm$1.84}  & 25.55\scriptsize{$\pm$0.57} & 11.12\scriptsize{$\pm$0.72} & 22.65\scriptsize{$\pm$1.57} & 26.12\scriptsize{$\pm$1.04} \\
& PointRooms        & 24.80\scriptsize{$\pm$0.36} & \textbf{25.17}\scriptsize{$\pm$0.19} & \textbf{25.16}\scriptsize{$\pm$0.59} & 18.09\scriptsize{$\pm$4.13} & 19.41\scriptsize{$\pm$1.01} & 14.69\scriptsize{$\pm$2.51}  & 19.10\scriptsize{$\pm$1.39} & 9.76\scriptsize{$\pm$2.99} & 20.88\scriptsize{$\pm$0.96} & 22.80\scriptsize{$\pm$1.12} \\
& Reacher           &21.09\scriptsize{$\pm$0.65} & \textbf{24.49}\scriptsize{$\pm$0.48} & 22.24\scriptsize{$\pm$0.54} & \textbf{25.20}\scriptsize{$\pm$0.48} & 22.98\scriptsize{$\pm$0.91} & 10.62\scriptsize{$\pm$2.30} & 23.70\scriptsize{$\pm$0.62} & 4.74\scriptsize{$\pm$0.36} & 6.06\scriptsize{$\pm$0.84} & 18.16\scriptsize{$\pm$1.08} \\
& SawyerReach       & 26.70\scriptsize{$\pm$0.14} & \textbf{26.78}\scriptsize{$\pm$0.12} & \textbf{27.07}\scriptsize{$\pm$0.07} & 19.48\scriptsize{$\pm$1.39} & 21.32\scriptsize{$\pm$1.40} & 8.78\scriptsize{$\pm$2.59}  & 25.29\scriptsize{$\pm$0.35} & 3.46\scriptsize{$\pm$0.86} & 2.84\scriptsize{$\pm$0.05} & 21.16\scriptsize{$\pm$0.95} \\
& SawyerDoor        & 19.05\scriptsize{$\pm$0.66} & 20.37\scriptsize{$\pm$1.18} & \textbf{21.69}\scriptsize{$\pm$2.36} & \textbf{20.69}\scriptsize{$\pm$2.14} & 19.58\scriptsize{$\pm$3.55} & 12.47\scriptsize{$\pm$3.08} & 18.82\scriptsize{$\pm$1.67} & 7.92\scriptsize{$\pm$0.86} & 14.77\scriptsize{$\pm$0.51} & 16.56\scriptsize{$\pm$0.92} \\
& FetchReach        &\textbf{30.42}\scriptsize{$\pm$0.04} & \textbf{30.44}\scriptsize{$\pm$0.02} & \textbf{30.42}\scriptsize{$\pm$0.04} & 28.34\scriptsize{$\pm$0.98} & 27.94\scriptsize{$\pm$0.30} & 18.96\scriptsize{$\pm$1.77} & 27.11\scriptsize{$\pm$0.22} & 1.71\scriptsize{$\pm$0.77} & 1.21\scriptsize{$\pm$0.46} & 23.02\scriptsize{$\pm$1.64} \\
& FetchPush         & 5.21\scriptsize{$\pm$0.43} & 6.83\scriptsize{$\pm$0.32} & \textbf{7.22}\scriptsize{$\pm$0.35} & \textbf{6.99}\scriptsize{$\pm$1.27} & 5.35\scriptsize{$\pm$3.36} & 4.22\scriptsize{$\pm$2.19} & 4.53\scriptsize{$\pm$1.94} & 4.49\scriptsize{$\pm$1.34} & 5.35\scriptsize{$\pm$0.23} & 5.10\scriptsize{$\pm$0.62}\\
& FetchPick         & 3.75\scriptsize{$\pm$0.18} & \textbf{4.22}\scriptsize{$\pm$0.16} & \textbf{4.36}\scriptsize{$\pm$0.19} & 3.81\scriptsize{$\pm$3.71} & 1.87\scriptsize{$\pm$1.59} & 0.81\scriptsize{$\pm$0.82} & 3.08\scriptsize{$\pm$1.35} & 2.16\scriptsize{$\pm$0.75} & 2.17\scriptsize{$\pm$0.18} & 3.21\scriptsize{$\pm$0.32} \\
& FetchSlide        & 2.67\scriptsize{$\pm$0.35} & \textbf{2.98}\scriptsize{$\pm$0.21} & \textbf{3.15}\scriptsize{$\pm$0.14} & 1.32\scriptsize{$\pm$1.22} & 1.04\scriptsize{$\pm$0.98} & 0.24\scriptsize{$\pm$0.27} & 1.12\scriptsize{$\pm$0.39} & 1.31\scriptsize{$\pm$0.52} & 0.00\scriptsize{$\pm$0.00} & 0.54\scriptsize{$\pm$0.12} \\
& HandReach         & 14.89\scriptsize{$\pm$2.54} & \textbf{18.24}\scriptsize{$\pm$2.18} & \textbf{20.06}\scriptsize{$\pm$3.06} & 0.08\scriptsize{$\pm$0.07} & 2.54\scriptsize{$\pm$1.42} & 1.41\scriptsize{$\pm$0.51} & 0.00\scriptsize{$\pm$0.00} & 0.00\scriptsize{$\pm$0.00} & 0.00\scriptsize{$\pm$0.00} & 8.36\scriptsize{$\pm$0.18} \\
\hline \rule{0pt}{2.5ex}%
& Average Rank & \multicolumn{1}{c}{3.8} & \multicolumn{1}{c}{\textbf{1.9}} & \multicolumn{1}{c|}{\textbf{1.7}} & \multicolumn{1}{c}{4.5} & \multicolumn{1}{c}{5.5} & \multicolumn{1}{c}{8.6} & \multicolumn{1}{c|}{6.0} & \multicolumn{1}{c}{8.8} & \multicolumn{1}{c}{7.9} & \multicolumn{1}{c|}{5.9} \\
\hline
\end{tabular}
\end{adjustbox}
\end{center}
\label{table:returns2}%
\end{table*}

\begin{table}[h]
\caption{\small{Success rates for state-space input,, averaged over 10 seeds.}}
\vspace{-1em}
\begin{center}
\begin{adjustbox}{width=0.965\textwidth}
\begin{tabular}{|ll|rrr|rrrr|rrr|}
\hline \\[-2.35ex]
\multicolumn{2}{c|}{\multirow{2}{*}{\textbf{Task Name}}} & \multicolumn{3}{c|}{\textbf{Ours}} & \multicolumn{4}{c|}{\textbf{Offline GCRL}} & \multicolumn{3}{c|}{\textbf{Diffusion-based}}\\
\rule{0pt}{2ex}
& & \multicolumn{1}{c}{\textbf{Merlin}} & \multicolumn{1}{c}{\textbf{Merlin-P}} & \multicolumn{1}{c|}{\textbf{Merlin-NP}} & \multicolumn{1}{c}{\textbf{GoFAR}} & \multicolumn{1}{c}{\textbf{WGCSL}} & \multicolumn{1}{c}{\textbf{GCSL}} & \multicolumn{1}{c|}{\textbf{AM}} & \multicolumn{1}{c}{\textbf{DD}} & \multicolumn{1}{c}{\textbf{g-DQL}} & \multicolumn{1}{c|}{\textbf{BESO}} \\
\hline \rule{0pt}{3ex}
\multirow{10}{*}{\rotatebox[origin=c]{90}{\textbf{Expert}}}
& PointReach        & \textbf{1.00}\scriptsize{$\pm$0.00} & \textbf{1.00}\scriptsize{$\pm$0.00} & \textbf{1.00}\scriptsize{$\pm$0.00} & \textbf{1.00}\scriptsize{$\pm$0.00} & \textbf{1.00}\scriptsize{$\pm$0.00} & \textbf{1.00}\scriptsize{$\pm$0.00} & \textbf{1.00}\scriptsize{$\pm$0.00} & 0.40\scriptsize{$\pm$0.00} & \textbf{1.00}\scriptsize{$\pm$0.00} & \textbf{1.00}\scriptsize{$\pm$0.00} \\
& PointRooms        & 0.91\scriptsize{$\pm$0.16} & 0.90\scriptsize{$\pm$0.04} & \textbf{0.94}\scriptsize{$\pm$0.01} & 0.82\scriptsize{$\pm$0.04} & 0.82\scriptsize{$\pm$0.04} & 0.79\scriptsize{$\pm$0.6} & 0.87\scriptsize{$\pm$0.05} & 0.27\scriptsize{$\pm$0.17} & \textbf{1.00}\scriptsize{$\pm$0.00} & 0.89\scriptsize{$\pm$0.04}\\
& Reacher           & \textbf{1.00}\scriptsize{$\pm$0.00} & \textbf{1.00}\scriptsize{$\pm$0.00} & \textbf{1.00}\scriptsize{$\pm$0.00} & \textbf{1.00}\scriptsize{$\pm$0.00} & \textbf{1.00}\scriptsize{$\pm$0.00} & \textbf{1.00}\scriptsize{$\pm$0.00} & \textbf{1.00}\scriptsize{$\pm$0.00} & 0.20\scriptsize{$\pm$0.00} & \textbf{1.00}\scriptsize{$\pm$0.00} & \textbf{1.00}\scriptsize{$\pm$0.00}\\
& SawyerReach       & \textbf{1.00}\scriptsize{$\pm$0.00} & \textbf{1.00}\scriptsize{$\pm$0.00} & \textbf{1.00}\scriptsize{$\pm$0.00} & \textbf{1.00}\scriptsize{$\pm$0.00} & \textbf{1.00}\scriptsize{$\pm$0.00} & \textbf{1.00}\scriptsize{$\pm$0.00} & \textbf{1.00}\scriptsize{$\pm$0.00} & 0.13\scriptsize{$\pm$0.05} & \textbf{1.00}\scriptsize{$\pm$0.00} & \textbf{1.00}\scriptsize{$\pm$0.00}\\
& SawyerDoor        & \textbf{0.95}\scriptsize{$\pm$0.08} & 0.92\scriptsize{$\pm$0.08} & \textbf{0.94}\scriptsize{$\pm$0.11} & 0.82\scriptsize{$\pm$0.12} & 0.86\scriptsize{$\pm$0.15} & 0.84\scriptsize{$\pm$0.16} & 0.92\scriptsize{$\pm$0.12} & 0.20\scriptsize{$\pm$0.00} & \textbf{0.94}\scriptsize{$\pm$0.12} & 0.79\scriptsize{$\pm$0.08}\\
& FetchReach        & \textbf{1.00}\scriptsize{$\pm$0.00} & \textbf{1.00}\scriptsize{$\pm$0.00} & \textbf{1.00}\scriptsize{$\pm$0.00} & \textbf{1.00}\scriptsize{$\pm$0.00} & \textbf{1.00}\scriptsize{$\pm$0.00} & 0.98\scriptsize{$\pm$0.00} & \textbf{1.00}\scriptsize{$\pm$0.00} & 0.07\scriptsize{$\pm$0.04} & \textbf{1.00}\scriptsize{$\pm$0.00} & \textbf{1.00}\scriptsize{$\pm$0.00}\\
& FetchPush         & 0.92\scriptsize{$\pm$0.05} & 0.94\scriptsize{$\pm$0.03} & \textbf{0.96}\scriptsize{$\pm$0.04} & \textbf{0.96}\scriptsize{$\pm$0.04} & 0.95\scriptsize{$\pm$0.04} & 0.88\scriptsize{$\pm$0.09} & 0.90\scriptsize{$\pm$0.08} & 0.17\scriptsize{$\pm$0.09} & 0.89\scriptsize{$\pm$0.11} & 0.85\scriptsize{$\pm$0.10}\\
& FetchPick         & \textbf{0.92}\scriptsize{$\pm$0.03} & \textbf{0.96}\scriptsize{$\pm$0.03} & \textbf{0.96}\scriptsize{$\pm$0.06} & 0.78\scriptsize{$\pm$0.04} & 0.76\scriptsize{$\pm$0.07} & 0.64\scriptsize{$\pm$0.09} & 0.69\scriptsize{$\pm$0.16} & 0.07\scriptsize{$\pm$0.07} & 0.78\scriptsize{$\pm$0.07} & 0.84\scriptsize{$\pm$0.11}\\
& FetchSlide        & 0.32\scriptsize{$\pm$0.04} & 0.40\scriptsize{$\pm$0.06} & \textbf{0.45}\scriptsize{$\pm$0.07} & 0.28\scriptsize{$\pm$0.09} & \textbf{0.42}\scriptsize{$\pm$0.14} & 0.22\scriptsize{$\pm$0.14} & 0.32\scriptsize{$\pm$0.12} & 0.10\scriptsize{$\pm$0.08} & 0.05\scriptsize{$\pm$0.04} & 0.30\scriptsize{$\pm$0.08}\\
& HandReach         & \textbf{0.78}\scriptsize{$\pm$0.04} & 0.82\scriptsize{$\pm$0.07} & \textbf{0.85}\scriptsize{$\pm$0.02} &  0.54\scriptsize{$\pm$0.23} & 0.68\scriptsize{$\pm$0.19} & 0.03\scriptsize{$\pm$0.05} & 0.00\scriptsize{$\pm$0.00} & 0.00\scriptsize{$\pm$0.00} & 0.00\scriptsize{$\pm$0.00} & 0.48\scriptsize{$\pm$0.12}\\
\hline \rule{0pt}{3ex}  
& Average Rank & 2.3 & \textbf{2.2} & \textbf{1.2} & 3.7 & 3.3 & 6.0 & 4.0 & 8.8 & 3.7 & 4.2 \\
\hline \rule{0pt}{3ex}  
\multirow{10}{*}{\rotatebox[origin=c]{90}{\textbf{Random}}}
& PointReach        & \textbf{1.00}\scriptsize{$\pm$0.00} & \textbf{1.00}\scriptsize{$\pm$0.00} & \textbf{1.00}\scriptsize{$\pm$0.00} & \textbf{1.00}\scriptsize{$\pm$0.00} & \textbf{1.00}\scriptsize{$\pm$0.00} & \textbf{1.00}\scriptsize{$\pm$0.00} & \textbf{1.00}\scriptsize{$\pm$0.00} & 0.40\scriptsize{$\pm$0.00} & 0.95\scriptsize{$\pm$0.06} & \textbf{1.00}\scriptsize{$\pm$0.00}\\
& PointRooms        & \textbf{0.89}\scriptsize{$\pm$0.02} & \textbf{0.92}\scriptsize{$\pm$0.03} & \textbf{0.92}\scriptsize{$\pm$0.02} & 0.78\scriptsize{$\pm$0.11} & 0.83\scriptsize{$\pm$0.08} & 0.77\scriptsize{$\pm$0.11}  & 0.70\scriptsize{$\pm$0.16}& 0.27\scriptsize{$\pm$0.17} & 0.82\scriptsize{$\pm$0.08} & 0.86\scriptsize{$\pm$0.08} \\
& Reacher           & 0.98\scriptsize{$\pm$0.02} & \textbf{1.00}\scriptsize{$\pm$0.00} & \textbf{1.00}\scriptsize{$\pm$0.00} & 0.98\scriptsize{$\pm$0.03}& \textbf{1.00}\scriptsize{$\pm$0.00} & 0.80\scriptsize{$\pm$0.06} & \textbf{1.00}\scriptsize{$\pm$0.00} & 0.23\scriptsize{$\pm$0.05} & 0.15\scriptsize{$\pm$0.05} & 0.92\scriptsize{$\pm$0.06}\\
& SawyerReach       & \textbf{1.00}\scriptsize{$\pm$0.00} & \textbf{1.00}\scriptsize{$\pm$0.00} & \textbf{1.00}\scriptsize{$\pm$0.00}  & 0.92\scriptsize{$\pm$0.07} & \textbf{1.00}\scriptsize{$\pm$0.00} & 0.91\scriptsize{$\pm$0.09}  & \textbf{1.00}\scriptsize{$\pm$0.00} & 0.13\scriptsize{$\pm$0.05} & 0.10\scriptsize{$\pm$0.03} & \textbf{1.00}\scriptsize{$\pm$0.00}\\
& SawyerDoor        & \textbf{0.57}\scriptsize{$\pm$0.03} & \textbf{0.59}\scriptsize{$\pm$0.08} & \textbf{0.59}\scriptsize{$\pm$0.05}  & 0.46\scriptsize{$\pm$0.19} & 0.48\scriptsize{$\pm$0.17} & 0.26\scriptsize{$\pm$0.09} & 0.44\scriptsize{$\pm$0.16} & 0.20\scriptsize{$\pm$0.00} & 0.35\scriptsize{$\pm$0.09} & 0.38\scriptsize{$\pm$0.11}\\
& FetchReach        & \textbf{1.00}\scriptsize{$\pm$0.00} & \textbf{1.00}\scriptsize{$\pm$0.00} & \textbf{1.00}\scriptsize{$\pm$0.00}  & \textbf{1.00}\scriptsize{$\pm$0.00} & \textbf{1.00}\scriptsize{$\pm$0.00}  & 0.96\scriptsize{$\pm$0.05} & \textbf{1.00}\scriptsize{$\pm$0.00} & 0.07\scriptsize{$\pm$0.04} & 0.00\scriptsize{$\pm$0.00} & 0.95\scriptsize{$\pm$0.05}\\
& FetchPush         & 0.20\scriptsize{$\pm$0.06} & \textbf{0.22}\scriptsize{$\pm$0.01} & \textbf{0.24}\scriptsize{$\pm$0.09} & \textbf{0.22}\scriptsize{$\pm$0.04} & 0.14\scriptsize{$\pm$0.10} & 0.20\scriptsize{$\pm$0.11} & 0.13\scriptsize{$\pm$0.09} & 0.13\scriptsize{$\pm$0.05} & 0.17\scriptsize{$\pm$0.04} & 0.18\scriptsize{$\pm$0.04}\\
& FetchPick         & 0.12\scriptsize{$\pm$0.01} & \textbf{0.17}\scriptsize{$\pm$0.02} & \textbf{0.18}\scriptsize{$\pm$0.01} & \textbf{0.12}\scriptsize{$\pm$0.11} & 0.08\scriptsize{$\pm$0.07} & 0.06\scriptsize{$\pm$0.08} & 0.10\scriptsize{$\pm$0.02} & 0.07\scriptsize{$\pm$0.05} & 0.09\scriptsize{$\pm$0.02} & 0.12\scriptsize{$\pm$0.02}\\
& FetchSlide        & 0.11\scriptsize{$\pm$0.02} & \textbf{0.18}\scriptsize{$\pm$0.04} & \textbf{0.20}\scriptsize{$\pm$0.04} & 0.10\scriptsize{$\pm$0.06} & 0.04\scriptsize{$\pm$0.08} & 0.06\scriptsize{$\pm$0.08} & 0.07\scriptsize{$\pm$0.04} & 0.07\scriptsize{$\pm$0.05} & 0.00\scriptsize{$\pm$0.00} & 0.08\scriptsize{$\pm$0.02}\\
& HandReach         & 0.49\scriptsize{$\pm$0.05} & \textbf{0.56}\scriptsize{$\pm$0.12} & \textbf{0.62}\scriptsize{$\pm$0.07} & 0.00\scriptsize{$\pm$0.00} & 0.12\scriptsize{$\pm$0.07} & 0.04\scriptsize{$\pm$0.04} & 0.00\scriptsize{$\pm$0.00} & 0.00\scriptsize{$\pm$0.00} & 0.00\scriptsize{$\pm$0.00} & 0.26\scriptsize{$\pm$0.06}\\
\hline \rule{0pt}{3ex}  
& Average Rank & 2.7 & \textbf{1.4} & \textbf{1.0} & 4.2 & 4.3 & 6.9 & 4.7 & 8.8 & 8.4 & 4.6 \\
\hline
\end{tabular}
\end{adjustbox}
\end{center}
\label{table:success}
\end{table}

\begin{table}[H]
\captionof{table}{\small{Discounted returns for pixel-space input, averaged over 10 seeds.}}%
\vspace{-1em}
\begin{center}
\begin{adjustbox}{width=\textwidth}
\begin{tabular}{|ll|rrr|rrrr|rrr|}
\hline \\[-2.35ex]
\multicolumn{2}{c|}{\multirow{2}{*}{\textbf{Task Name}}} & \multicolumn{3}{c|}{\textbf{Ours}} & \multicolumn{4}{c|}{\textbf{Offline GCRL}} & \multicolumn{3}{c|}{\textbf{Diffusion-based}}\\
\rule{0pt}{2ex}
& & \multicolumn{1}{c}{\textbf{Merlin}} & \multicolumn{1}{c}{\textbf{Merlin-P}} & \multicolumn{1}{c|}{\textbf{Merlin-NP}} & \multicolumn{1}{c}{\textbf{GoFAR}} & \multicolumn{1}{c}{\textbf{WGCSL}} & \multicolumn{1}{c}{\textbf{GCSL}} & \multicolumn{1}{c|}{\textbf{AM}} & \multicolumn{1}{c}{\textbf{DD}} & \multicolumn{1}{c}{\textbf{g-DQL}} & \multicolumn{1}{c|}{\textbf{BESO}} \\
\hline \rule{0pt}{3ex}
\multirow{4}{*}{\rotatebox[origin=c]{90}{\textbf{Expert}}}
& PointReach        & 27.69\scriptsize{$\pm$0.06} & \textbf{28.54}\scriptsize{$\pm$0.08} & \textbf{28.95}\scriptsize{$\pm$0.05} & 25.14\scriptsize{$\pm$0.52} & 24.25\scriptsize{$\pm$0.60} & 21.06\scriptsize{$\pm$1.06} & 25.16\scriptsize{$\pm$1.22} & 8.20\scriptsize{$\pm$0.75} & 26.48\scriptsize{$\pm$0.76} & 27.92\scriptsize{$\pm$0.55} \\
& PointRooms        & 23.76\scriptsize{$\pm$0.19} & \textbf{25.16}\scriptsize{$\pm$0.26}  & \textbf{25.28}\scriptsize{$\pm$0.22} & 20.06\scriptsize{$\pm$0.34} & 19.72\scriptsize{$\pm$0.86} & 18.15\scriptsize{$\pm$1.59} & 22.47\scriptsize{$\pm$1.25}  & 5.54\scriptsize{$\pm$1.88} & 26.28\scriptsize{$\pm$0.64} & 23.80\scriptsize{$\pm$0.62} \\
& SawyerReach       & 26.87\scriptsize{$\pm$0.04} & \textbf{26.98}\scriptsize{$\pm$0.08}  & \textbf{27.15}\scriptsize{$\pm$0.06} & 22.16\scriptsize{$\pm$0.84} & 21.59\scriptsize{$\pm$1.02} & 19.04\scriptsize{$\pm$1.14} & 23.10\scriptsize{$\pm$1.12} & 6.89\scriptsize{$\pm$0.88} & 23.92\scriptsize{$\pm$0.15} & 25.96\scriptsize{$\pm$0.44} \\
& SawyerDoor        & 25.42\scriptsize{$\pm$0.08} & 25.15\scriptsize{$\pm$0.18}  & \textbf{26.08}\scriptsize{$\pm$0.08} & 23.17\scriptsize{$\pm$0.32} & 23.24\scriptsize{$\pm$0.75} & 19.76\scriptsize{$\pm$1.36} & \textbf{25.89}\scriptsize{$\pm$0.48} & 6.06\scriptsize{$\pm$1.12} & 24.44\scriptsize{$\pm$0.85} & 22.78\scriptsize{$\pm$1.28} \\
\hline \rule{0pt}{2.5ex}%
& Average Rank & \multicolumn{1}{c}{3.75} & \multicolumn{1}{c}{\textbf{2.75}} & \multicolumn{1}{c|}{\textbf{1.25}} & \multicolumn{1}{c}{7.0} & \multicolumn{1}{c}{7.5} & \multicolumn{1}{c}{9.0} & \multicolumn{1}{c|}{5.0} & \multicolumn{1}{c}{10.0} & \multicolumn{1}{c}{4.0} & \multicolumn{1}{c|}{4.75} \\
\hline \rule{0pt}{3ex}%
\multirow{4}{*}{\rotatebox[origin=c]{90}{\textbf{Random}}}
& PointReach        & 27.52\scriptsize{$\pm$0.05} & \textbf{28.80}\scriptsize{$\pm$0.08} & \textbf{28.76}\scriptsize{$\pm$0.06} & 23.51\scriptsize{$\pm$0.68} & 25.10\scriptsize{$\pm$0.88} & 17.34\scriptsize{$\pm$1.20} & 24.89\scriptsize{$\pm$0.72} & 6.36\scriptsize{$\pm$1.04} & 22.15\scriptsize{$\pm$1.32} & 25.85\scriptsize{$\pm$0.98} \\
& PointRooms        & 22.40\scriptsize{$\pm$0.07} & \textbf{24.05}\scriptsize{$\pm$0.22}  & \textbf{24.02}\scriptsize{$\pm$0.09} & 17.82\scriptsize{$\pm$1.89} & 19.02\scriptsize{$\pm$1.20} & 14.12\scriptsize{$\pm$1.92} & 18.82\scriptsize{$\pm$1.72} & 4.67\scriptsize{$\pm$2.15} & 20.16\scriptsize{$\pm$0.98} & 22.24\scriptsize{$\pm$1.08} \\
& SawyerReach       & 26.14\scriptsize{$\pm$0.04} & \textbf{26.46}\scriptsize{$\pm$0.10}  & \textbf{26.78}\scriptsize{$\pm$0.05} & 19.22\scriptsize{$\pm$1.08} & 21.04\scriptsize{$\pm$1.18} & 8.64\scriptsize{$\pm$2.44} & 25.01\scriptsize{$\pm$0.42} &  2.02\scriptsize{$\pm$2.59} & 2.32\scriptsize{$\pm$1.01} & 20.89\scriptsize{$\pm$0.98} \\
& SawyerDoor        & 18.99\scriptsize{$\pm$0.08} & 20.10\scriptsize{$\pm$1.78}  & \textbf{21.12}\scriptsize{$\pm$0.09} & \textbf{20.43}\scriptsize{$\pm$1.89} & 19.38\scriptsize{$\pm$1.68} & 12.04\scriptsize{$\pm$2.81} & 17.72\scriptsize{$\pm$0.84} & 4.12\scriptsize{$\pm$1.32} & 14.18\scriptsize{$\pm$0.65} & 16.24\scriptsize{$\pm$1.14} \\
\hline \rule{0pt}{2.5ex}%
& Average Rank & \multicolumn{1}{c}{3.5} & \multicolumn{1}{c}{\textbf{1.75}} & \multicolumn{1}{c|}{\textbf{1.5}} & \multicolumn{1}{c}{6.0} & \multicolumn{1}{c}{5.0} & \multicolumn{1}{c}{8.75} & \multicolumn{1}{c|}{5.75} & \multicolumn{1}{c}{10.0} & \multicolumn{1}{c}{7.5} & \multicolumn{1}{c|}{5.25} \\
\hline
\end{tabular}
\end{adjustbox}
\end{center}
\label{table:returns_img2}%
\end{table}

\begin{table}[H]
\captionof{table}{\small{Success rates for pixel-space input, averaged over 10 seeds.}}%
\vspace{-1em}
\begin{center}
\begin{adjustbox}{width=0.965\textwidth}
\begin{tabular}{|ll|rrr|rrrr|rrr|}
\hline \\[-2.35ex]
\multicolumn{2}{c|}{\multirow{2}{*}{\textbf{Task Name}}} & \multicolumn{3}{c|}{\textbf{Ours}} & \multicolumn{4}{c|}{\textbf{Offline GCRL}} & \multicolumn{3}{c|}{\textbf{Diffusion-based}}\\
\rule{0pt}{2ex}
& & \multicolumn{1}{c}{\textbf{Merlin}} & \multicolumn{1}{c}{\textbf{Merlin-P}} & \multicolumn{1}{c|}{\textbf{Merlin-NP}} & \multicolumn{1}{c}{\textbf{GoFAR}} & \multicolumn{1}{c}{\textbf{WGCSL}} & \multicolumn{1}{c}{\textbf{GCSL}} & \multicolumn{1}{c|}{\textbf{AM}} & \multicolumn{1}{c}{\textbf{DD}} & \multicolumn{1}{c}{\textbf{g-DQL}} & \multicolumn{1}{c|}{\textbf{BESO}} \\
\hline \rule{0pt}{3ex}
\multirow{4}{*}{\rotatebox[origin=c]{90}{\textbf{Expert}}}
& PointReach        & \textbf{1.00}\scriptsize{$\pm$0.00} & \textbf{1.00}\scriptsize{$\pm$0.00} & \textbf{1.00}\scriptsize{$\pm$0.00} & \textbf{1.00}\scriptsize{$\pm$0.00}& \textbf{1.00}\scriptsize{$\pm$0.00} & \textbf{1.00}\scriptsize{$\pm$0.00} & \textbf{1.00}\scriptsize{$\pm$0.00} & 0.44\scriptsize{$\pm$0.12} & \textbf{1.00}\scriptsize{$\pm$0.00} & \textbf{1.00}\scriptsize{$\pm$0.00} \\
& PointRooms        & 0.86\scriptsize{$\pm$0.09} & \textbf{0.90}\scriptsize{$\pm$0.06}  & \textbf{0.90}\scriptsize{$\pm$0.08} & 0.80\scriptsize{$\pm$0.04} & 0.78\scriptsize{$\pm$0.06} & 0.78\scriptsize{$\pm$0.09} & 0.81\scriptsize{$\pm$0.05}  & 0.42\scriptsize{$\pm$0.08} & \textbf{0.92}\scriptsize{$\pm$0.07} & 0.87\scriptsize{$\pm$0.08} \\
& SawyerReach       & \textbf{1.00}\scriptsize{$\pm$0.00} & \textbf{1.00}\scriptsize{$\pm$0.00}  & \textbf{1.00}\scriptsize{$\pm$0.00} & \textbf{1.00}\scriptsize{$\pm$0.00} & \textbf{1.00}\scriptsize{$\pm$0.00} & \textbf{1.00}\scriptsize{$\pm$0.00} & \textbf{1.00}\scriptsize{$\pm$0.00} & 0.08\scriptsize{$\pm$0.02} & \textbf{1.00}\scriptsize{$\pm$0.00} & \textbf{1.00}\scriptsize{$\pm$0.00} \\
& SawyerDoor        & 0.90\scriptsize{$\pm$0.08} & 0.90\scriptsize{$\pm$0.10}  & \textbf{0.92}\scriptsize{$\pm$0.08} & 0.88\scriptsize{$\pm$0.04} & 0.88\scriptsize{$\pm$0.07} & 0.82\scriptsize{$\pm$0.06} & \textbf{0.92}\scriptsize{$\pm$0.08} & 0.16\scriptsize{$\pm$0.02} & 0.80\scriptsize{$\pm$0.05} & 0.78\scriptsize{$\pm$0.08} \\
\hline \rule{0pt}{2.5ex}%
& Average Rank & \multicolumn{1}{c}{2.5} & \multicolumn{1}{c}{\textbf{1.75}} & \multicolumn{1}{c|}{\textbf{1.25}} & \multicolumn{1}{c}{3.5} & \multicolumn{1}{c}{3.75} & \multicolumn{1}{c}{4.25} & \multicolumn{1}{c|}{2.25} & \multicolumn{1}{c}{10.0} & \multicolumn{1}{c}{2.75} & \multicolumn{1}{c|}{3.75} \\
\hline \rule{0pt}{3ex}%
\multirow{4}{*}{\rotatebox[origin=c]{90}{\textbf{Random}}}
& PointReach        & \textbf{1.00}\scriptsize{$\pm$0.00} & \textbf{1.00}\scriptsize{$\pm$0.00} & \textbf{1.00}\scriptsize{$\pm$0.00} & \textbf{1.00}\scriptsize{$\pm$0.00} & \textbf{1.00}\scriptsize{$\pm$0.00} & \textbf{1.00}\scriptsize{$\pm$0.00} & \textbf{1.00}\scriptsize{$\pm$0.00} & 0.34\scriptsize{$\pm$0.08} & 0.93\scriptsize{$\pm$0.04} & \textbf{1.00}\scriptsize{$\pm$0.00} \\
& PointRooms        & 0.78\scriptsize{$\pm$0.06} & \textbf{0.83}\scriptsize{$\pm$0.05}  & 0.82\scriptsize{$\pm$0.09} & 0.70\scriptsize{$\pm$0.09} & 0.79\scriptsize{$\pm$0.06} & 0.69\scriptsize{$\pm$0.04} & 0.72\scriptsize{$\pm$0.07} & 0.20\scriptsize{$\pm$0.05} & 0.80\scriptsize{$\pm$0.09} & \textbf{0.84}\scriptsize{$\pm$0.07} \\
& SawyerReach       & \textbf{1.00}\scriptsize{$\pm$0.00} & \textbf{1.00}\scriptsize{$\pm$0.00}  & \textbf{1.00}\scriptsize{$\pm$0.00} & 0.89\scriptsize{$\pm$0.08} & \textbf{1.00}\scriptsize{$\pm$0.00} & 0.89\scriptsize{$\pm$0.04} & \textbf{1.00}\scriptsize{$\pm$0.00} &  0.08\scriptsize{$\pm$0.02} & 0.09\scriptsize{$\pm$0.01} & \textbf{1.00}\scriptsize{$\pm$0.00} \\
& SawyerDoor        & 0.55\scriptsize{$\pm$0.08} & \textbf{0.56}\scriptsize{$\pm$0.06}  & \textbf{0.57}\scriptsize{$\pm$0.08} & \textbf{0.57}\scriptsize{$\pm$0.09} & 0.48\scriptsize{$\pm$0.06} & 0.28\scriptsize{$\pm$0.04} & 0.39\scriptsize{$\pm$0.04} & 0.14\scriptsize{$\pm$0.03} & 0.32\scriptsize{$\pm$0.08} & 0.33\scriptsize{$\pm$0.04} \\
\hline \rule{0pt}{2.5ex}%
& Average Rank & \multicolumn{1}{c}{3.0} & \multicolumn{1}{c}{\textbf{1.75}} & \multicolumn{1}{c|}{\textbf{1.5}} & \multicolumn{1}{c}{4.25} & \multicolumn{1}{c}{3.0} & \multicolumn{1}{c}{6.5} & \multicolumn{1}{c|}{3.75} & \multicolumn{1}{c}{10.0} & \multicolumn{1}{c}{7.5} & \multicolumn{1}{c|}{2.5} \\
\hline
\end{tabular}
\end{adjustbox}
\end{center}
\label{table:successes_img}%
\end{table}


\end{document}